\pdfoutput=1
\documentclass{article}
\usepackage{iclr2027_conference,times}
\usepackage{orcidlink}         %

\usepackage{amsmath,amssymb}
\usepackage{booktabs}
\usepackage{graphicx}
\usepackage{hyperref}
\usepackage{url}
\iclrfinalcopy %

\let\cite\citep

\usepackage{multirow}
\usepackage{caption}
\usepackage{subcaption}
\usepackage{xcolor}
\usepackage{etoc}
\usepackage{adjustbox}
\usepackage{etoolbox}

\newcommand{\yes}{\checkmark}
\newcommand{\no}{$\times$}
\newcommand{\prt}{$\circ$}

\newcommand{\rNseqHuPR}{235}      %
\newcommand{\rNseqRT}{100}        %
\newcommand{\rNseqRTOne}{63}      %
\newcommand{\rNseqRTTwo}{37}      %

\newcommand{\rChirpsTotal}{584 million}
\newcommand{\rChirpRenders}{234 billion}
\newcommand{\rScaleVsThreeDPS}{$63{,}000$}

\newcommand{\rCorrFitMDc}{0.993}  %
\newcommand{\rCorrFitMDs}{0.966}  %
\newcommand{\rCorrFitMDcRT}{0.973}  %
\newcommand{\rCorrFitMDsRT}{0.704}  %

\newcommand{\rCorrRA}{0.940}  %
\newcommand{\rCorrRAc}{0.666}  %
\newcommand{\rCorrRDocc}{0.829}  %
\newcommand{\rCorrRDfloorU}{0.92 / 0.92}  %
\newcommand{\rCorrRDfloorURT}{0.96}  %
\newcommand{\rCorrRDfloorUHu}{0.92}  %

\newcommand{\rChirpsSeqHuPR}{921{,}600}
\newcommand{\rChirpsSeqRT}{3{,}674{,}112}
\newcommand{\rChirpRatioHuPR}{0.25}
\newcommand{\rAngResHuPR}{14.3}
\newcommand{\rAngResRT}{1.3}
\newcommand{\rNelHuPR}{2}
\newcommand{\rNelRT}{8}
\newcommand{\rCamResHuPR}{0.2}
\newcommand{\rLabelVsRadarHuPR}{seventy}
\newcommand{\rCorrMD}{0.988}  %
\newcommand{\rCorrRART}{0.830}  %
\newcommand{\rCorrRAcRT}{0.470}  %
\newcommand{\rCorrRDoccRT}{0.695}  %
\newcommand{\rCorrMDRT}{0.966}  %

\newcommand{\rApReal}{--}
\newcommand{\rApTSTR}{--}
\newcommand{\rApSub}{--}
\newcommand{\rApSelf}{--}
\newcommand{\rRetention}{--}       %

\newcommand{\rMPJPEReal}{X.XX}     %
\newcommand{\rMPJPETSTR}{X.XX}     %
\newcommand{\rMPJPESub}{X.XX}      %
\newcommand{\rMPJPESelf}{X.XX}     %
\newcommand{\rMPJPEAug}{X.XX}      %
\newcommand{\rMPJPEMixed}{\rMPJPEAug} %
\newcommand{\rApAug}{--}           %
\newcommand{\rApMixed}{\rApAug}    %
\newcommand{\rRetentionRT}{0.76}  %

\newcommand{\rMinHuPR}{\rRtFitMinHu}        %
\newcommand{\rMinRT}{\rRtFitMinRT}          %
\newcommand{\rGPUh}{\rRtFitGPUhTotal}       %

\renewcommand{\rApReal}{65.6 (1.7)}
\renewcommand{\rApSub}{49.2 (0.7)}
\renewcommand{\rApTSTR}{38.4 (2.9)}
\renewcommand{\rApSelf}{59.6 (1.4)}
\renewcommand{\rMPJPEReal}{16.67 (0.42)}
\renewcommand{\rMPJPESub}{20.05 (1.50)}
\renewcommand{\rMPJPETSTR}{21.86 (1.33)}
\renewcommand{\rMPJPESelf}{15.71 (1.38)}

\renewcommand{\rApAug}{61.6 (0.4)}
\renewcommand{\rMPJPEAug}{16.52 (0.80)}
\renewcommand{\rRetention}{0.59}

\newcommand{\rRfgRA}{0.134}  %
\newcommand{\rRfgRAc}{0.339}  %
\newcommand{\rRfgRDocc}{0.518}  %
\newcommand{\rRfgRDfloorU}{0.05}  %
\newcommand{\rRfgMDc}{0.820}  %
\newcommand{\rRfgMDs}{0.063}  %
\newcommand{\rRfgMD}{0.671}  %
\newcommand{\rRfgRART}{0.119}  %
\newcommand{\rRfgRAcRT}{0.221}  %
\newcommand{\rRfgRDoccRT}{0.324}  %
\newcommand{\rRfgRDfloorURT}{0.09}  %
\newcommand{\rRfgMDcRT}{0.750}  %
\newcommand{\rRfgMDsRT}{0.042}  %
\newcommand{\rRfgMDRT}{0.394}  %
\newcommand{\rSuppSplitRowRA}{0.835 / 0.451}  %
\newcommand{\rSuppSplitRowRDMD}{0.695 / 0.968}  %
\newcommand{\rSuppRaRaeHuHOurs}{\textbf{0.960} & \textbf{0.699} & \textbf{0.79} & \textbf{0.28} & \textbf{34.0} & \textbf{0.942} & \textbf{0.80} & \textbf{0.33} & \textbf{33.3}}  %
\newcommand{\rSuppRdMdHuHOurs}{\textbf{0.891} & \textbf{0.71} & \textbf{0.92} & \textbf{0.88} & \textbf{1.00} & \textbf{0.98} & \textbf{0.993} & \textbf{0.87} & 0.87 & \textbf{0.91}}  %
\newcommand{\rSuppAdcHuHOurs}{\textbf{0.857} & \textbf{21.2} & \textbf{0.58} & \textbf{0.088} & \textbf{0.55} & \textbf{63}}  %
\newcommand{\rSuppCrpHuHOurs}{\textbf{0.856} & \textbf{26.0} & \textbf{0.65} & \textbf{0.052} & \textbf{0.66} & \textbf{56} & \textbf{0.75} & \textbf{48}}  %
\newcommand{\rSuppRaRaeHuHRfg}{0.139 & 0.374 & 0.51 & 1.08 & 22.5 & 0.165 & 0.48 & 1.11 & 22.7}  %
\newcommand{\rSuppRdMdHuHRfg}{0.564 & 1.51 & 0.06 & 1.30 & 0.82 & 0.08 & 0.670 & 0.28 & 4.22 & 0.34}  %
\newcommand{\rSuppAdcHuHRfg}{0.649 & 14.0 & 0.37 & 0.205 & -0.21 & 109}  %
\newcommand{\rSuppCrpHuHRfg}{0.246 & 20.0 & 0.28 & 0.102 & -0.63 & 147 & 0.52 & 67}  %
\newcommand{\rSuppRaRaeHuHUnfit}{0.901 & 0.353 & 0.68 & 0.44 & 30.3 & -- & -- & -- & --}  %
\newcommand{\rSuppRdMdHuHUnfit}{0.584 & 0.60 & 0.57 & 0.66 & 0.94 & 0.31 & 0.771 & 0.79 & \textbf{1.07} & 0.89}  %
\newcommand{\rSuppRaRaeHuVOurs}{\textbf{0.925} & \textbf{0.622} & \textbf{0.67} & \textbf{0.38} & \textbf{30.2} & \textbf{0.904} & \textbf{0.72} & \textbf{0.40} & \textbf{30.7}}  %
\newcommand{\rSuppRdMdHuVOurs}{\textbf{0.795} & 0.54 & \textbf{0.92} & \textbf{0.88} & \textbf{0.99} & \textbf{0.94} & \textbf{0.984} & \textbf{0.80} & 0.79 & \textbf{0.83}}  %
\newcommand{\rSuppAdcHuVOurs}{\textbf{0.772} & \textbf{19.0} & \textbf{0.48} & \textbf{0.114} & \textbf{0.44} & \textbf{72}}  %
\newcommand{\rSuppCrpHuVOurs}{\textbf{0.785} & \textbf{24.8} & \textbf{0.56} & \textbf{0.061} & \textbf{0.55} & \textbf{65} & \textbf{0.47} & \textbf{71}}  %
\newcommand{\rSuppRaRaeHuVRfg}{0.129 & 0.317 & 0.35 & 1.10 & 21.0 & 0.153 & 0.36 & 1.13 & 21.8}  %
\newcommand{\rSuppRdMdHuVRfg}{0.480 & \textbf{1.12} & 0.05 & 1.24 & 0.82 & 0.03 & 0.672 & 0.28 & 3.79 & 0.37}  %
\newcommand{\rSuppAdcHuVRfg}{0.629 & 14.0 & 0.34 & 0.203 & 0.11 & 90}  %
\newcommand{\rSuppCrpHuVRfg}{0.274 & 20.0 & 0.19 & 0.102 & -0.52 & 139 & 0.31 & 82}  %
\newcommand{\rSuppRaRaeHuVUnfit}{0.863 & 0.234 & 0.53 & 0.52 & 27.6 & -- & -- & -- & --}  %
\newcommand{\rSuppRdMdHuVUnfit}{0.513 & 0.37 & 0.56 & 0.60 & 0.93 & 0.25 & 0.740 & 0.77 & \textbf{0.91} & \textbf{0.83}}  %
\newcommand{\rSuppRaRaeRTOurs}{\textbf{0.830} & \textbf{0.470} & \textbf{0.78} & \textbf{0.53} & \textbf{31.8} & \textbf{0.779} & \textbf{0.81} & \textbf{0.58} & \textbf{31.9}}  %
\newcommand{\rSuppRdMdRTOurs}{\textbf{0.695} & \textbf{0.63} & \textbf{0.96} & \textbf{0.94} & \textbf{0.97} & \textbf{0.70} & \textbf{0.966} & \textbf{0.65} & \textbf{0.71} & \textbf{0.71}}  %
\newcommand{\rSuppAdcRTOurs}{\textbf{0.667} & \textbf{20.0} & \textbf{0.32} & \textbf{0.100} & \textbf{0.24} & \textbf{88}}  %
\newcommand{\rSuppCrpRTOurs}{\textbf{0.735} & \textbf{28.2} & \textbf{0.76} & \textbf{0.039} & \textbf{0.50} & \textbf{68} & \textbf{0.33} & \textbf{82}}  %
\newcommand{\rSuppRaRaeRTRfg}{0.119 & 0.221 & 0.46 & 1.13 & 25.4 & 0.117 & 0.51 & 1.11 & 26.5}  %
\newcommand{\rSuppRdMdRTRfg}{0.324 & 0.61 & 0.09 & 1.51 & 0.75 & 0.04 & 0.394 & 0.07 & 1.53 & 0.52}  %
\newcommand{\rSuppAdcRTRfg}{0.630 & 15.1 & 0.28 & 0.179 & -0.05 & 103}  %
\newcommand{\rSuppCrpRTRfg}{0.211 & 22.8 & 0.43 & 0.073 & -0.54 & 137 & 0.16 & 94}  %
\newcommand{\rSuppRaRaeRTUnfit}{0.648 & 0.181 & 0.72 & 0.90 & 27.5 & -- & -- & -- & --}  %
\newcommand{\rSuppRdMdRTUnfit}{0.291 & 0.28 & 0.58 & 0.75 & 0.80 & 0.19 & 0.515 & 0.46 & 0.60 & 0.59}  %
\newcommand{\rRfgNseq}{335}  %
\newcommand{\rRfgScatterers}{800}  %
\newcommand{\rAdcPhaseClean}{0.51}  %
\newcommand{\rAdcPhaseCluttered}{0.23}  %

\newcommand{\rAblAdcHuHNone}{\textbf{0.857} & \textbf{21.2} & \textbf{0.58} & \textbf{0.088} & 0.55 & 63}  %
\newcommand{\rAblAdcHuHNoRaeZero}{0.742 & 18.4 & 0.45 & 0.121 & \textbf{0.63} & \textbf{56}}  %
\newcommand{\rAblAdcHuHNoRaePl}{0.856 & \textbf{21.2} & \textbf{0.58} & \textbf{0.088} & 0.54 & 64}  %
\newcommand{\rAblAdcHuHNoWin}{0.856 & \textbf{21.2} & \textbf{0.58} & \textbf{0.088} & 0.55 & 64}  %
\newcommand{\rAblAdcHuVNone}{0.772 & \textbf{19.0} & \textbf{0.48} & 0.114 & 0.44 & 72}  %
\newcommand{\rAblAdcHuVNoRaeZero}{0.668 & 16.1 & 0.35 & 0.157 & \textbf{0.56} & \textbf{62}}  %
\newcommand{\rAblAdcHuVNoRaePl}{0.773 & \textbf{19.0} & \textbf{0.48} & \textbf{0.113} & 0.45 & 72}  %
\newcommand{\rAblAdcHuVNoWin}{\textbf{0.776} & \textbf{19.0} & \textbf{0.48} & 0.114 & 0.45 & 72}  %
\newcommand{\rAblAdcRTNone}{0.667 & 20.0 & 0.32 & 0.100 & 0.24 & 88}  %
\newcommand{\rAblAdcRTNoRaeZero}{0.596 & 18.2 & 0.26 & 0.123 & \textbf{0.27} & \textbf{85}}  %
\newcommand{\rAblAdcRTNoRaePl}{\textbf{0.668} & \textbf{20.2} & \textbf{0.33} & \textbf{0.098} & 0.24 & 87}  %
\newcommand{\rAblAdcRTNoWin}{0.658 & 20.0 & 0.31 & 0.101 & 0.23 & 89}  %
\newcommand{\rAblCrpHuHNone}{\textbf{0.856} & \textbf{26.0} & \textbf{0.65} & \textbf{0.052} & \textbf{0.66} & \textbf{56} & \textbf{0.75} & \textbf{48}}  %
\newcommand{\rAblCrpHuHNoRaeZero}{0.667 & 22.4 & 0.44 & 0.076 & 0.64 & 58 & 0.66 & 57}  %
\newcommand{\rAblCrpHuHNoRaePl}{0.850 & 25.9 & 0.62 & \textbf{0.052} & 0.65 & \textbf{56} & \textbf{0.75} & \textbf{48}}  %
\newcommand{\rAblCrpHuHNoWin}{0.854 & \textbf{26.0} & \textbf{0.65} & \textbf{0.052} & 0.64 & 57 & \textbf{0.75} & \textbf{48}}  %
\newcommand{\rAblCrpHuVNone}{0.785 & \textbf{24.8} & \textbf{0.56} & 0.061 & 0.55 & 65 & 0.47 & 71}  %
\newcommand{\rAblCrpHuVNoRaeZero}{0.491 & 20.6 & 0.26 & 0.095 & \textbf{0.57} & \textbf{63} & \textbf{0.49} & \textbf{69}}  %
\newcommand{\rAblCrpHuVNoRaePl}{\textbf{0.789} & \textbf{24.8} & 0.52 & 0.061 & 0.55 & 65 & 0.48 & 71}  %
\newcommand{\rAblCrpHuVNoWin}{0.788 & \textbf{24.8} & \textbf{0.56} & \textbf{0.060} & 0.55 & 65 & 0.48 & 71}  %
\newcommand{\rAblCrpRTNone}{0.735 & 28.2 & \textbf{0.76} & 0.039 & 0.50 & 68 & \textbf{0.33} & \textbf{82}}  %
\newcommand{\rAblCrpRTNoRaeZero}{0.227 & 23.7 & 0.52 & 0.066 & \textbf{0.53} & \textbf{66} & 0.27 & 85}  %
\newcommand{\rAblCrpRTNoRaePl}{\textbf{0.739} & \textbf{28.4} & 0.75 & \textbf{0.038} & 0.51 & 68 & \textbf{0.33} & \textbf{82}}  %
\newcommand{\rAblCrpRTNoWin}{0.727 & 27.9 & 0.73 & 0.041 & 0.51 & 69 & \textbf{0.33} & \textbf{82}}  %
\newcommand{\rAblRaRaeHuHNone}{\textbf{0.960} & \textbf{0.699} & \textbf{0.79} & \textbf{0.28} & 34.0 & \textbf{0.942} & \textbf{0.80} & 0.33 & \textbf{33.3}}  %
\newcommand{\rAblRaRaeHuHNoRaeZero}{0.646 & 0.456 & 0.67 & 1.60 & 19.0 & 0.628 & 0.66 & 1.87 & 18.7}  %
\newcommand{\rAblRaRaeHuHNoRaePl}{\textbf{0.960} & 0.696 & 0.77 & \textbf{0.28} & 34.0 & 0.941 & 0.78 & 0.33 & 33.2}  %
\newcommand{\rAblRaRaeHuHNoWin}{\textbf{0.960} & 0.690 & \textbf{0.79} & \textbf{0.28} & \textbf{34.1} & \textbf{0.942} & \textbf{0.80} & \textbf{0.32} & \textbf{33.3}}  %
\newcommand{\rAblRaRaeHuVNone}{0.925 & \textbf{0.622} & \textbf{0.67} & \textbf{0.38} & \textbf{30.2} & 0.904 & \textbf{0.72} & \textbf{0.40} & 30.7}  %
\newcommand{\rAblRaRaeHuVNoRaeZero}{0.337 & 0.126 & 0.56 & 10.23 & 1.6 & 0.349 & 0.56 & 11.71 & 1.7}  %
\newcommand{\rAblRaRaeHuVNoRaePl}{\textbf{0.926} & 0.618 & 0.65 & \textbf{0.38} & \textbf{30.2} & \textbf{0.905} & 0.70 & \textbf{0.40} & 30.7}  %
\newcommand{\rAblRaRaeHuVNoWin}{0.925 & 0.615 & \textbf{0.67} & \textbf{0.38} & \textbf{30.2} & 0.904 & \textbf{0.72} & \textbf{0.40} & \textbf{30.8}}  %
\newcommand{\rAblRaRaeRTNone}{0.830 & 0.470 & \textbf{0.78} & 0.53 & 31.8 & 0.779 & \textbf{0.81} & \textbf{0.58} & \textbf{31.9}}  %
\newcommand{\rAblRaRaeRTNoRaeZero}{0.141 & 0.166 & 0.65 & 7.79 & 9.9 & 0.125 & 0.65 & 7.08 & 11.7}  %
\newcommand{\rAblRaRaeRTNoRaePl}{\textbf{0.838} & \textbf{0.473} & 0.77 & \textbf{0.52} & \textbf{31.9} & \textbf{0.785} & 0.80 & \textbf{0.58} & \textbf{31.9}}  %
\newcommand{\rAblRaRaeRTNoWin}{0.828 & 0.444 & 0.77 & 0.53 & 31.7 & 0.762 & \textbf{0.81} & 0.59 & \textbf{31.9}}  %
\newcommand{\rAblRdMdHuHNone}{0.891 & \textbf{0.71} & \textbf{0.92} & 0.88 & \textbf{1.00} & \textbf{0.98} & 0.993 & 0.87 & 0.87 & \textbf{0.91}}  %
\newcommand{\rAblRdMdHuHNoRaeZero}{0.892 & 0.70 & \textbf{0.92} & 0.88 & \textbf{1.00} & \textbf{0.98} & 0.993 & 0.87 & 0.87 & \textbf{0.91}}  %
\newcommand{\rAblRdMdHuHNoRaePl}{0.878 & 0.69 & 0.79 & \textbf{0.94} & \textbf{1.00} & \textbf{0.98} & 0.993 & 0.87 & 0.87 & 0.90}  %
\newcommand{\rAblRdMdHuHNoWin}{\textbf{0.894} & \textbf{0.71} & \textbf{0.92} & 0.88 & \textbf{1.00} & \textbf{0.98} & \textbf{0.994} & \textbf{0.88} & \textbf{0.88} & \textbf{0.91}}  %
\newcommand{\rAblRdMdHuVNone}{0.795 & \textbf{0.54} & \textbf{0.92} & 0.88 & \textbf{0.99} & 0.94 & 0.984 & 0.80 & \textbf{0.79} & \textbf{0.83}}  %
\newcommand{\rAblRdMdHuVNoRaeZero}{0.796 & \textbf{0.54} & \textbf{0.92} & 0.88 & \textbf{0.99} & 0.94 & 0.984 & \textbf{0.81} & \textbf{0.79} & \textbf{0.83}}  %
\newcommand{\rAblRdMdHuVNoRaePl}{0.781 & \textbf{0.54} & 0.80 & \textbf{1.07} & \textbf{0.99} & 0.94 & 0.982 & 0.80 & 0.78 & \textbf{0.83}}  %
\newcommand{\rAblRdMdHuVNoWin}{\textbf{0.800} & \textbf{0.54} & \textbf{0.92} & 0.88 & \textbf{0.99} & \textbf{0.95} & \textbf{0.985} & \textbf{0.81} & \textbf{0.79} & \textbf{0.83}}  %
\newcommand{\rAblRdMdRTNone}{0.695 & 0.63 & \textbf{0.96} & \textbf{0.94} & 0.97 & 0.70 & 0.966 & 0.65 & 0.71 & \textbf{0.71}}  %
\newcommand{\rAblRdMdRTNoRaeZero}{\textbf{0.697} & 0.63 & \textbf{0.96} & \textbf{0.94} & 0.97 & 0.69 & 0.968 & 0.63 & 0.71 & \textbf{0.71}}  %
\newcommand{\rAblRdMdRTNoRaePl}{0.677 & \textbf{0.64} & 0.86 & 1.33 & 0.96 & 0.69 & 0.957 & \textbf{0.67} & \textbf{0.72} & 0.70}  %
\newcommand{\rAblRdMdRTNoWin}{0.695 & 0.60 & 0.95 & 0.93 & \textbf{0.98} & \textbf{0.71} & \textbf{0.970} & 0.63 & 0.71 & \textbf{0.71}}  %

\newcommand{\rAblCCrpCorrHuHNone}{0.856}  %
\newcommand{\rAblCRhoHuHNone}{0.960}  %
\newcommand{\rAblCLUHuHNone}{0.92}  %
\newcommand{\rAblCCrpCorrHuHNoRaeZero}{0.667}  %
\newcommand{\rAblWCrpCorrHuHNoRaeZero}{235}  %
\newcommand{\rAblNCrpCorrHuHNoRaeZero}{235}  %
\newcommand{\rAblCRhoHuHNoRaeZero}{0.646}  %
\newcommand{\rAblWRhoHuHNoRaeZero}{235}  %
\newcommand{\rAblNRhoHuHNoRaeZero}{235}  %
\newcommand{\rAblWCrpSsimHuHNoRaePl}{188}  %
\newcommand{\rAblWSsimHuHNoRaePl}{177}  %
\newcommand{\rAblNSsimHuHNoRaePl}{235}  %
\newcommand{\rAblWRhoMHuHNoRaePl}{91}  %
\newcommand{\rAblNRhoMHuHNoRaePl}{235}  %
\newcommand{\rAblCLUHuHNoRaePl}{0.79}  %
\newcommand{\rAblWLUHuHNoRaePl}{213}  %
\newcommand{\rAblNLUHuHNoRaePl}{235}  %
\newcommand{\rAblCRhoHuHNoWin}{0.960}  %
\newcommand{\rAblCLUHuVNone}{0.92}  %
\newcommand{\rAblCLUHuVNoRaePl}{0.80}  %
\newcommand{\rAblCCrpCorrRTNone}{0.735}  %
\newcommand{\rAblCRhoRTNone}{0.830}  %
\newcommand{\rAblCRhoCRTNone}{0.470}  %
\newcommand{\rAblCRaeRhoRTNone}{0.779}  %
\newcommand{\rAblCRhoMRTNone}{0.695}  %
\newcommand{\rAblCLURTNone}{0.96}  %
\newcommand{\rAblCSigmaURTNone}{0.94}  %
\newcommand{\rAblCMdSRTNone}{0.966}  %
\newcommand{\rAblCCrpCorrRTNoRaeZero}{0.227}  %
\newcommand{\rAblWCrpCorrRTNoRaeZero}{63}  %
\newcommand{\rAblNCrpCorrRTNoRaeZero}{63}  %
\newcommand{\rAblCRhoRTNoRaeZero}{0.141}  %
\newcommand{\rAblWRhoRTNoRaeZero}{63}  %
\newcommand{\rAblNRhoRTNoRaeZero}{63}  %
\newcommand{\rAblCRhoMRTNoRaeZero}{0.697}  %
\newcommand{\rAblCMdSRTNoRaeZero}{0.968}  %
\newcommand{\rAblCRhoRTNoRaePl}{0.838}  %
\newcommand{\rAblWRhoMRTNoRaePl}{53}  %
\newcommand{\rAblNRhoMRTNoRaePl}{63}  %
\newcommand{\rAblCLURTNoRaePl}{0.86}  %
\newcommand{\rAblCSigmaURTNoRaePl}{1.33}  %
\newcommand{\rAblCMdSRTNoRaePl}{0.957}  %
\newcommand{\rAblWMdSRTNoRaePl}{29}  %
\newcommand{\rAblNMdSRTNoRaePl}{61}  %
\newcommand{\rAblCRhoCRTNoWin}{0.444}  %
\newcommand{\rAblWRhoCRTNoWin}{51}  %
\newcommand{\rAblNRhoCRTNoWin}{61}  %
\newcommand{\rAblCRaeRhoRTNoWin}{0.762}  %
\newcommand{\rAblRTRhoNone}{0.830}  %
\newcommand{\rAblRTRhoSNone}{0.966}  %
\newcommand{\rAblRTGapNone}{2.00}  %
\newcommand{\rAblRTRhoNoRaeZero}{0.141}  %
\newcommand{\rAblRTRhoSNoRaeZero}{0.968}  %
\newcommand{\rAblRTGapNoRaeZero}{174.26}  %
\newcommand{\rAblRTRhoNoRaePl}{0.838}  %
\newcommand{\rAblRTRhoSNoRaePl}{0.957}  %
\newcommand{\rAblRTGapNoRaePl}{2.30}  %
\newcommand{\rAblRTRhoNoWin}{0.828}  %
\newcommand{\rAblRTRhoSNoWin}{0.970}  %
\newcommand{\rAblRTGapNoWin}{1.08}  %
\newcommand{\rAblHuRhoNone}{0.940}  %
\newcommand{\rAblHuRhoSNone}{0.988}  %
\newcommand{\rAblHuGapNone}{16.4}  %
\newcommand{\rAblHuRhoNoRaeZero}{0.485}  %
\newcommand{\rAblHuRhoSNoRaeZero}{0.989}  %
\newcommand{\rAblHuGapNoRaeZero}{16.5}  %
\newcommand{\rAblHuRhoNoRaePl}{0.941}  %
\newcommand{\rAblHuRhoSNoRaePl}{0.986}  %
\newcommand{\rAblHuGapNoRaePl}{16.7}  %
\newcommand{\rAblHuRhoNoWin}{0.941}  %
\newcommand{\rAblHuRhoSNoWin}{0.989}  %
\newcommand{\rAblHuGapNoWin}{17.4}  %
\newcommand{\rAblRTGapNNone}{13}  %
\newcommand{\rAblRTGapMRPENone}{0.47}  %
\newcommand{\rAblRTGapNNoRaeZero}{13}  %
\newcommand{\rAblRTGapMRPENoRaeZero}{262.43}  %
\newcommand{\rAblRTGapWNoRaeZero}{13}  %
\newcommand{\rAblRTGapBNoRaeZero}{0}  %
\newcommand{\rAblRTGapMRPENoRaePl}{1.21}  %
\newcommand{\rAblRTGapWNoRaePl}{4}  %
\newcommand{\rAblRTGapBNoRaePl}{2}  %
\newcommand{\rAblRTGapMRPENoWin}{0.88}  %
\newcommand{\rAblRTGapWNoWin}{2}  %
\newcommand{\rAblRTGapBNoWin}{2}  %
\newcommand{\rAblHuGapNNone}{21}  %
\newcommand{\rAblHuGapWNoRaeZero}{9}  %
\newcommand{\rAblHuGapBNoRaeZero}{5}  %
\newcommand{\rAblHuGapWNoRaePl}{5}  %
\newcommand{\rAblHuGapBNoRaePl}{9}  %
\newcommand{\rAblHuGapWNoWin}{12}  %
\newcommand{\rAblHuGapBNoWin}{2}  %
\newcommand{\rCorrUnfit}{0.878}  %
\newcommand{\rCorrUnfitRAc}{0.270}  %
\newcommand{\rCorrUnfitRD}{0.537}  %
\newcommand{\rCorrUnfitMD}{0.754}  %
\newcommand{\rCorrUnfitRT}{0.648}  %
\newcommand{\rCorrUnfitRAcRT}{0.181}  %
\newcommand{\rCorrUnfitRDRT}{0.291}  %
\newcommand{\rCorrUnfitMDRT}{0.515}  %
\newcommand{\rCorrUnfitMDc}{0.930}  %
\newcommand{\rCorrUnfitMDs}{0.283}  %
\newcommand{\rCorrUnfitMDcRT}{0.803}  %
\newcommand{\rCorrUnfitMDsRT}{0.185}  %
\newcommand{\rCorrRDfloorUnfitHu}{0.57}  %
\newcommand{\rCorrRDfloorUnfitRT}{0.58}  %
\newcommand{\rMPJPERealReleased}{16.7}  %

\newcommand{\rTstrHuAPfiftyRR}{96.9 (0.2)}  %
\newcommand{\rTstrHuAPsevfiveRR}{78.0 (3.1)}  %
\newcommand{\rTstrHuAPfiftyRS}{93.9 (0.8)}  %
\newcommand{\rTstrHuAPsevfiveRS}{45.7 (1.3)}  %
\newcommand{\rTstrHuAPfiftySR}{80.6 (3.2)}  %
\newcommand{\rTstrHuAPsevfiveSR}{30.1 (4.3)}  %
\newcommand{\rTstrHuAPfiftySS}{98.5 (0.1)}  %
\newcommand{\rTstrHuAPsevfiveSS}{67.9 (4.1)}  %
\newcommand{\rTstrHuAPfiftyMix}{96.4 (0.6)}  %
\newcommand{\rTstrHuAPsevfiveMix}{72.3 (0.2)}  %
\newcommand{\rTstrRTMRPERR}{13.14 (0.87)}  %
\newcommand{\rTstrRTAbsMPJPERR}{19.07 (0.52)}  %
\newcommand{\rTstrRTMRPERS}{14.32 (1.07)}  %
\newcommand{\rTstrRTAbsMPJPERS}{22.70 (1.76)}  %
\newcommand{\rTstrRTMRPESR}{17.06 (0.98)}  %
\newcommand{\rTstrRTAbsMPJPESR}{25.45 (1.69)}  %
\newcommand{\rTstrRTMRPESS}{10.08 (1.00)}  %
\newcommand{\rTstrRTAbsMPJPESS}{15.81 (1.30)}  %
\newcommand{\rTstrRTMRPEMix}{14.39 (0.93)}  %
\newcommand{\rTstrRTAbsMPJPEMix}{19.43 (0.35)}  %
\newcommand{\rMPJPERealMean}{16.7}  %
\newcommand{\rMPJPETSTRMean}{21.9}  %
\newcommand{\rKpWristLRR}{22.2}
\newcommand{\rKpWristRRR}{25.1}
\newcommand{\rKpWristLSR}{8.0}
\newcommand{\rKpWristRSR}{4.7}
\newcommand{\rKpRetTorsoLo}{0.73}
\newcommand{\rKpRetTorsoHi}{0.80}
\newcommand{\rKpRetArmsLo}{0.19}
\newcommand{\rKpRetArmsHi}{0.39}
\newcommand{\rTstrRfgRTMPJPERS}{28.46 (3.55)}  %
\newcommand{\rTstrRfgRTMRPERS}{28.56 (3.54)}  %
\newcommand{\rTstrRfgRTAbsMPJPERS}{38.15 (6.14)}  %
\newcommand{\rTstrRfgRTMPJPESR}{127.59 (48.32)}  %
\newcommand{\rTstrRfgRTMRPESR}{106.48 (45.00)}  %
\newcommand{\rTstrRfgRTAbsMPJPESR}{148.88 (24.86)}  %
\newcommand{\rTstrRfgRTMPJPESS}{12.44 (0.34)}  %
\newcommand{\rTstrRfgRTMRPESS}{8.97 (0.74)}  %
\newcommand{\rTstrRfgRTAbsMPJPESS}{12.72 (0.37)}  %
\newcommand{\rTstrRfgRTMPJPEMix}{16.82 (0.56)}  %
\newcommand{\rTstrRfgRTMRPEMix}{13.71 (0.88)}  %
\newcommand{\rTstrRfgRTAbsMPJPEMix}{19.41 (0.58)}  %
\newcommand{\rTstrRfgHuAPRS}{45.9 (3.9)}  %
\newcommand{\rTstrRfgHuAPfiftyRS}{97.4 (0.7)}  %
\newcommand{\rTstrRfgHuAPsevfiveRS}{28.5 (9.2)}  %
\newcommand{\rTstrRfgHuAPSR}{1.8 (3.5)}  %
\newcommand{\rTstrRfgHuAPfiftySR}{8.5 (16.1)}  %
\newcommand{\rTstrRfgHuAPsevfiveSR}{0.0 (0.1)}  %
\newcommand{\rTstrRfgHuAPSS}{77.0 (0.4)}  %
\newcommand{\rTstrRfgHuAPfiftySS}{98.8 (0.1)}  %
\newcommand{\rTstrRfgHuAPsevfiveSS}{93.8 (1.0)}  %
\newcommand{\rTstrRfgHuAPMix}{64.1 (0.9)}  %
\newcommand{\rTstrRfgHuAPfiftyMix}{96.9 (0.0)}  %
\newcommand{\rTstrRfgHuAPsevfiveMix}{76.5 (1.6)}  %
\newcommand{\rRfgMPJPETSTRMean}{127.6}  %
\newcommand{\rRfgApTSTRMean}{1.8}  %
\newcommand{\rRfgGain}{$4.57\times10^{8}$}  %
\newcommand{\rRfgGainLo}{$2.43\times10^{8}$}  %
\newcommand{\rRfgGainHi}{$8.55\times10^{8}$}  %
\newcommand{\rRfgGainSeqs}{44}  %
\newcommand{\rRfgGainFrames}{8}  %

\newcommand{\rRtFitMinHu}{5.7}  %
\newcommand{\rRtFitGPUhHu}{22.3}  %
\newcommand{\rRtFitNHu}{235}  %
\newcommand{\rRtPeakFitMedHu}{5.7}  %
\newcommand{\rRtPeakFitMaxHu}{6.7}  %
\newcommand{\rRtPeakGiBHu}{5.7}  %
\newcommand{\rRtStepMsHu}{19.4}  %
\newcommand{\rRtUnfitMinHu}{1.5}  %
\newcommand{\rRtUnfitGPUhHu}{5.9}  %
\newcommand{\rRtUnfitNHu}{235}  %
\newcommand{\rRtRfgMinHu}{5.3}  %
\newcommand{\rRtRfgGPUhHu}{21.0}  %
\newcommand{\rRtRfgNHu}{235}  %
\newcommand{\rRtTrainRealHu}{6.8 (0.1)}  %
\newcommand{\rRtTrainGPUhRealHu}{27.1}  %
\newcommand{\rRtTrainRenderedHu}{6.7 (0.0)}  %
\newcommand{\rRtTrainGPUhRenderedHu}{26.8}  %
\newcommand{\rRtTrainMixedHu}{6.7 (0.0)}  %
\newcommand{\rRtTrainGPUhMixedHu}{26.9}  %
\newcommand{\rRtTrainRfgRenderedHu}{6.7 (0.0)}  %
\newcommand{\rRtTrainGPUhRfgRenderedHu}{26.7}  %
\newcommand{\rRtTrainRfgMixedHu}{6.7 (0.0)}  %
\newcommand{\rRtTrainGPUhRfgMixedHu}{26.7}  %
\newcommand{\rRtTrainEpochsHu}{200}  %
\newcommand{\rRtEvalGPUhHu}{0.5}  %
\newcommand{\rRtEvalRfgGPUhHu}{0.5}  %
\newcommand{\rRtFitMinRT}{12.9}  %
\newcommand{\rRtFitGPUhRT}{21.6}  %
\newcommand{\rRtFitNRT}{100}  %
\newcommand{\rRtPeakFitMedRT}{18.7}  %
\newcommand{\rRtPeakFitMaxRT}{19.6}  %
\newcommand{\rRtPeakGiBRT}{18.7}  %
\newcommand{\rRtStepMsRT}{19.3}  %
\newcommand{\rRtUnfitMinRT}{1.9}  %
\newcommand{\rRtUnfitGPUhRT}{1.9}  %
\newcommand{\rRtUnfitNRT}{57}  %
\newcommand{\rRtRfgMinRT}{6.0}  %
\newcommand{\rRtRfgGPUhRT}{10.6}  %
\newcommand{\rRtRfgNRT}{100}  %
\newcommand{\rRtTrainRealRT}{6.5 (0.1)}  %
\newcommand{\rRtTrainGPUhRealRT}{26.2}  %
\newcommand{\rRtPeakTrainRealRT}{11.2}  %
\newcommand{\rRtTrainRenderedRT}{8.4 (0.0)}  %
\newcommand{\rRtTrainGPUhRenderedRT}{33.7}  %
\newcommand{\rRtPeakTrainRenderedRT}{11.2}  %
\newcommand{\rRtTrainMixedRT}{8.2 (0.2)}  %
\newcommand{\rRtTrainGPUhMixedRT}{32.9}  %
\newcommand{\rRtPeakTrainMixedRT}{11.2}  %
\newcommand{\rRtTrainRfgRenderedRT}{7.7 (0.1)}  %
\newcommand{\rRtTrainGPUhRfgRenderedRT}{31.0}  %
\newcommand{\rRtPeakTrainRfgRenderedRT}{11.2}  %
\newcommand{\rRtTrainRfgMixedRT}{7.9 (0.3)}  %
\newcommand{\rRtTrainGPUhRfgMixedRT}{31.5}  %
\newcommand{\rRtPeakTrainRfgMixedRT}{11.2}  %
\newcommand{\rRtTrainEpochsRT}{50}  %
\newcommand{\rRtEvalGPUhRT}{0.7}  %
\newcommand{\rRtEvalRfgGPUhRT}{0.7}  %
\newcommand{\rRtPeakRfgMaxHu}{3.3}  %
\newcommand{\rRtPeakRfgTorchHu}{1.1}  %
\newcommand{\rRtPeakRfgMaxRT}{10.8}  %
\newcommand{\rRtPeakRfgTorchRT}{1.6}  %
\newcommand{\rRtPeakTrainHu}{14.9}  %
\newcommand{\rRtPeakTrainSmiHu}{21.9}  %
\newcommand{\rRtPeakTrainArmsHu}{5}  %
\newcommand{\rRtPeakTrainStepsHu}{300}  %
\newcommand{\rRtFitGPUhTotal}{43.9}  %

\newcommand{\rReprGateLocal}{16.652}  %
\newcommand{\rReprGateNode}{16.651}  %
\newcommand{\rReprFrames}{3{,}887}  %
\newcommand{\rReprRrr}{0.320}  %
\newcommand{\rReprRxr}{0.234}  %
\newcommand{\rReprRratio}{1.13}  %
\newcommand{\rReprRci}{[0.99, 1.31]}  %
\newcommand{\rReprArr}{0.690}  %
\newcommand{\rReprAxr}{0.618}  %
\newcommand{\rReprAratio}{1.23}  %
\newcommand{\rReprAci}{[1.21, 1.26]}  %
\newcommand{\rReprRratioS}{1.01}  %
\newcommand{\rReprAratioS}{1.11}  %
\newcommand{\rReprCkaTone}{1.35}  %
\newcommand{\rReprCkaTtwoa}{1.17}  %
\newcommand{\rReprCkaMidLo}{0.98}  %
\newcommand{\rReprCkaMidHi}{1.10}  %
\newcommand{\rReprCkaTfouraRatio}{0.98}  %
\newcommand{\rReprCkaTheadRatio}{0.99}  %
\newcommand{\rReprProcLo}{0.98}  %
\newcommand{\rReprProcHi}{1.05}  %
\newcommand{\rReprSameLo}{0.88}  %
\newcommand{\rReprSameHi}{1.13}  %
\newcommand{\rReprMixLo}{0.92}  %
\newcommand{\rReprMixHi}{1.24}  %
\newcommand{\rPairRealLo}{2.87}  %
\newcommand{\rPairRealHi}{3.29}  %
\newcommand{\rPairRendLo}{2.54}  %
\newcommand{\rPairRendHi}{3.05}  %
\newcommand{\rProbeReal}{13.46}  %
\newcommand{\rProbeRend}{15.62}  %
\newcommand{\rProbeRatio}{1.16}  %
\newcommand{\rProbeCi}{[1.09, 1.28]}  %
\newcommand{\rProbeRatioFloor}{1.25}  %
\newcommand{\rProbeDrop}{16}  %
\newcommand{\rProbeHeadDrop}{31}  %
\newcommand{\rReprMixConsLo}{0.88}  %
\newcommand{\rReprMixConsHi}{0.99}  %
\newcommand{\rReprWhitenLo}{1.00}  %
\newcommand{\rReprWhitenHi}{1.02}  %
\newcommand{\rReprMixOutTap}{T2a}  %
\newcommand{\rReprMixOutRatio}{1.24}  %
\newcommand{\rReprMixOutCi}{[1.12, 1.38]}  %
\newcommand{\rReprNRr}{10}  %
\newcommand{\rReprNXrSame}{4}  %
\newcommand{\rReprNXr}{16}  %
\newcommand{\rReprNMxSame}{4}  %
\newcommand{\rReprNMx}{16}  %
\newcommand{\rReprNSs}{6}  %
\newcommand{\rReprNFloor}{5}  %

\newcommand{\rReprHuGateLocal}{0.66871}  %
\newcommand{\rReprHuGateNode}{0.66871}  %
\newcommand{\rReprHuFrames}{12{,}600}  %
\newcommand{\rReprHuKrr}{0.589}  %
\newcommand{\rReprHuKxr}{0.335}  %
\newcommand{\rReprHuArr}{0.936}  %
\newcommand{\rReprHuAxr}{0.812}  %
\newcommand{\rReprHuAratio}{2.94}  %
\newcommand{\rReprHuAci}{[2.78, 3.11]}  %
\newcommand{\rReprHuRratio}{2.20}  %
\newcommand{\rReprHuRci}{[2.07, 2.34]}  %
\newcommand{\rReprHuCkaChirprr}{0.976}  %
\newcommand{\rReprHuCkaChirpxr}{0.973}  %
\newcommand{\rReprHuCkaChirpfloor}{0.858}  %
\newcommand{\rReprHuCkaChirpratio}{1.15}  %
\newcommand{\rReprHuCkaChirpci}{[0.97, 1.35]}  %
\newcommand{\rReprHuProChirpratio}{1.08}  %
\newcommand{\rReprHuCkaEncrr}{0.287}  %
\newcommand{\rReprHuCkaEncxr}{0.229}  %
\newcommand{\rReprHuCkaEncfloor}{0.227}  %
\newcommand{\rReprHuCkaEncratio}{1.08}  %
\newcommand{\rReprHuCkaEncci}{[0.93, 1.27]}  %
\newcommand{\rReprHuProEncratio}{1.05}  %
\newcommand{\rReprHuCkaHeadrr}{0.674}  %
\newcommand{\rReprHuCkaHeadxr}{0.554}  %
\newcommand{\rReprHuCkaHeadfloor}{0.040}  %
\newcommand{\rReprHuCkaHeadratio}{1.37}  %
\newcommand{\rReprHuCkaHeadci}{[1.08, 1.70]}  %
\newcommand{\rReprHuProHeadratio}{1.17}  %
\newcommand{\rReprHuCkaVertrr}{0.164}  %
\newcommand{\rReprHuCkaVertxr}{0.084}  %
\newcommand{\rReprHuCkaVertfloor}{0.044}  %
\newcommand{\rReprHuCkaVertratio}{1.10}  %
\newcommand{\rReprHuCkaVertci}{[0.96, 1.30]}  %
\newcommand{\rReprHuProVertratio}{1.15}  %
\newcommand{\rReprHuProLo}{1.05}  %
\newcommand{\rReprHuProHi}{1.17}  %
\newcommand{\rReprHuCkaChirpVratio}{1.37}  %
\newcommand{\rReprHuProbeRealPx}{8.98}  %
\newcommand{\rReprHuProbeRendPx}{11.66}  %
\newcommand{\rReprHuProbeRendRatio}{1.30}  %
\newcommand{\rReprHuProbeRendCi}{[1.20, 1.38]}  %
\newcommand{\rReprHuProbeMixedPx}{9.64}  %
\newcommand{\rReprHuProbeFloorPx}{22.43}  %
\newcommand{\rReprHuProbeFloorRatio}{2.50}  %
\newcommand{\rReprHuPairRealChirp}{0.97}  %
\newcommand{\rReprHuPairRendChirp}{0.97}  %
\newcommand{\rReprHuProbeAgree}{0.03}  %
\newcommand{\rReprHuDeadHtwoh}{1}  %
\newcommand{\rReprHuDeadHthreeh}{3}  %
\newcommand{\rReprHuDeadHtwov}{5}  %
\newcommand{\rReprHuDeadHthreev}{3}  %
\newcommand{\rReprHuTrained}{13}  %

\newcommand{\rReprSimRowTZeroinput}{1.000 & 1.000 & 1.000 & -- & 0.000 & 0.000 & 0.000 & --}
\newcommand{\rReprSimRowTOnelayerOne}{0.713 & 0.613 & 0.417 & 1.35 [1.21, 1.49] & 0.816 & 0.857 & 1.013 & 1.05 [1.02, 1.07]}
\newcommand{\rReprSimRowTTwoa}{0.537 & 0.459 & 0.188 & 1.17 [1.01, 1.35] & 0.993 & 1.038 & 1.213 & 1.04 [1.01, 1.08]}
\newcommand{\rReprSimRowTTwob}{0.350 & 0.366 & 0.169 & 0.98 [0.89, 1.07] & 1.095 & 1.087 & 1.261 & 0.99 [0.96, 1.02]}
\newcommand{\rReprSimRowTThreea}{0.228 & 0.200 & 0.071 & 1.04 [0.94, 1.15] & 1.223 & 1.238 & 1.345 & 1.01 [0.97, 1.05]}
\newcommand{\rReprSimRowTThreeb}{0.300 & 0.270 & 0.085 & 1.04 [0.92, 1.19] & 1.182 & 1.202 & 1.338 & 1.02 [0.98, 1.06]}
\newcommand{\rReprSimRowTThreec}{0.251 & 0.194 & 0.133 & 1.08 [1.05, 1.11] & 1.236 & 1.275 & 1.340 & 1.03 [1.02, 1.04]}
\newcommand{\rReprSimRowTFoura}{0.069 & 0.085 & 0.039 & 0.98 [0.92, 1.05] & 1.357 & 1.359 & 1.397 & 1.00 [0.95, 1.07]}
\newcommand{\rReprSimRowTFourb}{0.267 & 0.247 & 0.078 & 1.03 [0.96, 1.10] & 1.200 & 1.222 & 1.353 & 1.02 [0.99, 1.05]}
\newcommand{\rReprSimRowTFourc}{0.368 & 0.343 & 0.140 & 1.04 [0.97, 1.11] & 1.240 & 1.266 & 1.343 & 1.02 [1.01, 1.04]}
\newcommand{\rReprSimRowTFourd}{0.332 & 0.266 & 0.111 & 1.10 [1.03, 1.16] & 1.254 & 1.302 & 1.355 & 1.04 [1.02, 1.05]}
\newcommand{\rReprSimRowTSixhead}{0.048 & 0.055 & 0.014 & 0.99 [0.92, 1.07] & 1.410 & 1.385 & 1.478 & 0.98 [0.90, 1.09]}

\newcommand{\rEditNpairHP}{20}  %
\newcommand{\rEditNrenderHP}{40}  %
\newcommand{\rEditSepRAHP}{0.863}  %
\newcommand{\rEditSepRAcHP}{0.382}  %
\newcommand{\rEditSepRDoccHP}{0.802}  %
\newcommand{\rEditSepLUHP}{0.92}  %
\newcommand{\rEditSepMDcHP}{0.970}  %
\newcommand{\rEditSepMDsHP}{0.976}  %
\newcommand{\rEditSepMDSHP}{0.985}  %
\newcommand{\rEditSepNHP}{80}  %
\newcommand{\rEditOwnRAHP}{0.942}  %
\newcommand{\rEditOwnRAcHP}{0.617}  %
\newcommand{\rEditOwnRDoccHP}{0.802}  %
\newcommand{\rEditOwnLUHP}{0.92}  %
\newcommand{\rEditOwnMDcHP}{0.970}  %
\newcommand{\rEditOwnMDsHP}{0.976}  %
\newcommand{\rEditOwnMDSHP}{0.985}  %
\newcommand{\rEditOwnNHP}{80}  %
\newcommand{\rEditUnfitRAHP}{0.884}  %
\newcommand{\rEditUnfitRAcHP}{0.136}  %
\newcommand{\rEditUnfitRDoccHP}{0.414}  %
\newcommand{\rEditUnfitLUHP}{0.57}  %
\newcommand{\rEditUnfitMDcHP}{0.823}  %
\newcommand{\rEditUnfitMDsHP}{0.265}  %
\newcommand{\rEditUnfitMDSHP}{0.733}  %
\newcommand{\rEditUnfitNHP}{80}  %
\newcommand{\rEditRfgRAHP}{0.131}  %
\newcommand{\rEditRfgRAcHP}{0.161}  %
\newcommand{\rEditRfgRDoccHP}{0.410}  %
\newcommand{\rEditRfgLUHP}{0.05}  %
\newcommand{\rEditRfgMDcHP}{0.689}  %
\newcommand{\rEditRfgMDsHP}{0.019}  %
\newcommand{\rEditRfgMDSHP}{0.648}  %
\newcommand{\rEditRfgNHP}{80}  %
\newcommand{\rEditDOwnRAHP}{$-$0.079}  %
\newcommand{\rEditRetRAHP}{92}  %
\newcommand{\rEditRetRAcHP}{62}  %
\newcommand{\rEditRetRDoccHP}{100}  %
\newcommand{\rEditRetLUHP}{100}  %
\newcommand{\rEditRetMDcHP}{100}  %
\newcommand{\rEditRetMDsHP}{100}  %
\newcommand{\rEditDOwnMDSHP}{0.000}  %
\newcommand{\rEditRetMDSHP}{100}  %
\newcommand{\rEditWUnfitRAHP}{48}  %
\newcommand{\rEditNpairRT}{24}  %
\newcommand{\rEditNrenderRT}{48}  %
\newcommand{\rEditSepRART}{0.756}  %
\newcommand{\rEditSepRAcRT}{0.484}  %
\newcommand{\rEditSepRDoccRT}{0.696}  %
\newcommand{\rEditSepLURT}{0.97}  %
\newcommand{\rEditSepMDcRT}{0.962}  %
\newcommand{\rEditSepMDsRT}{0.679}  %
\newcommand{\rEditSepMDSRT}{0.962}  %
\newcommand{\rEditSepNRT}{48}  %
\newcommand{\rEditOwnRART}{0.843}  %
\newcommand{\rEditOwnRAcRT}{0.540}  %
\newcommand{\rEditOwnRDoccRT}{0.697}  %
\newcommand{\rEditOwnLURT}{0.97}  %
\newcommand{\rEditOwnMDcRT}{0.962}  %
\newcommand{\rEditOwnMDsRT}{0.679}  %
\newcommand{\rEditOwnMDSRT}{0.962}  %
\newcommand{\rEditOwnNRT}{48}  %
\newcommand{\rEditUnfitRART}{0.660}  %
\newcommand{\rEditUnfitRAcRT}{0.120}  %
\newcommand{\rEditUnfitRDoccRT}{0.282}  %
\newcommand{\rEditUnfitLURT}{0.58}  %
\newcommand{\rEditUnfitMDcRT}{0.581}  %
\newcommand{\rEditUnfitMDsRT}{0.168}  %
\newcommand{\rEditUnfitMDSRT}{0.497}  %
\newcommand{\rEditUnfitNRT}{44}  %
\newcommand{\rEditRfgRART}{0.131}  %
\newcommand{\rEditRfgRAcRT}{0.221}  %
\newcommand{\rEditRfgRDoccRT}{0.335}  %
\newcommand{\rEditRfgLURT}{0.09}  %
\newcommand{\rEditRfgMDcRT}{0.711}  %
\newcommand{\rEditRfgMDsRT}{0.068}  %
\newcommand{\rEditRfgMDSRT}{0.401}  %
\newcommand{\rEditRfgNRT}{48}  %
\newcommand{\rEditDOwnRART}{$-$0.055}  %
\newcommand{\rEditRetRART}{90}  %
\newcommand{\rEditRetRAcRT}{90}  %
\newcommand{\rEditRetRDoccRT}{100}  %
\newcommand{\rEditRetLURT}{100}  %
\newcommand{\rEditRetMDcRT}{100}  %
\newcommand{\rEditRetMDsRT}{100}  %
\newcommand{\rEditDOwnMDSRT}{0.000}  %
\newcommand{\rEditRetMDSRT}{100}  %
\newcommand{\rEditSceneThr}{0.75}  %
\newcommand{\rEditCrossMax}{0.74}  %
\newcommand{\rEditPairRhoMinHP}{0.79}  %
\newcommand{\rEditPairRhoMinRT}{0.77}  %
\newcommand{\rNovelNmotion}{6}  %
\newcommand{\rNovelNscene}{5}  %
\newcommand{\rNovelNmoved}{7}  %
\newcommand{\rNovelMaxShift}{0.9}  %

\renewcommand{\rMPJPEMixed}{\rMPJPEAug}
\renewcommand{\rApMixed}{\rApAug}

\begin{document}

\etocsettocdepth.toc{none}

\title{4D Point Splatting for mmWave Radar Micro-Doppler Differentiable Rendering}

\author{%
  Adnan Armouti\,\orcidlink{0000-0002-4448-3089} \\
  Cornell Tech \\
  New York, NY, USA \\
  \texttt{aa2546@cornell.edu}
  \And
  Guanyue Qian\,\orcidlink{0009-0005-2494-7412} \\
  New York University \\
  New York, NY, USA \\
  \texttt{gq2032@nyu.edu}
  \And
  Rajalakshmi Nandakumar\,\orcidlink{0000-0002-1601-148X} \\
  Cornell Tech \\
  New York, NY, USA \\
  \texttt{rn283@cornell.edu}
}

\maketitle

\begin{abstract}
Millimeter-wave radar resolves the micrometer displacements of a moving body
as micro-Doppler, enabling gait, gesture, activity, pose and vital-sign
sensing. Each task relies on a downstream model trained on real data, yet
radar datasets lag orders of magnitude behind vision and language. Brute-force
collection will not close that gap, and existing synthesis methods
(forward-only simulators, learned generators and static-scene inverse
renderers) cannot fit a real capture of a moving body, leaving a substantial
sim2real gap. We present 4D Point Splatting (4DPS), a physics-based
differentiable renderer that fits an explicit moving scene to a real capture.
Using the capture's synchronized camera to initialize the room and the moving
body, we represent the scene as oriented points carrying material, normal and
velocity, and splat each point into range and Doppler bins. Our output is the
raw complex return a real radar records, so we can apply the same digital
signal processing (DSP) on the rendered data, and obtain any DSP
representation to run on and train any SOTA downstream model. We fit \rNseqHuPR{}
sequences on HuPR and \rNseqRT{} on RT-Pose, two datasets with different
radars. Our reconstructions set the state of the art, with median correlations
of \rCorrRA{} and \rCorrRART{} on range--azimuth, \rCorrRDocc{}
and \rCorrRDoccRT{} on range--Doppler, and \rCorrMD{} and \rCorrMDRT{} on
micro-Doppler. Our renders also set the state of the art in downstream
sim2real gap. On RT-Pose, a pose model trained on them alone reaches
\rMPJPETSTRMean{}\,cm MPJPE on real data, against \rMPJPERealMean{}\,cm for
real training and \rRfgMPJPETSTRMean{}\,cm for the RF-Genesis baseline, and we show that both the rendered- and
real-trained models learn nearly the same features. Finally, our custom CUDA
kernels fit a sequence in \rMinHuPR{} minutes on HuPR and \rMinRT{} on RT-Pose
on one H100, making dataset-scale inverse rendering of mmWave radar possible
for the first time. Together, this work can enable synthetic data generation for any
human-centered downstream task and model, and scene editing for data
augmentation.
\end{abstract}

\section{Introduction}
\label{sec:intro}

Millimeter-wave (mmWave) radar resolves radial motion to a fraction of a
millimeter, works in darkness and through occlusion, and captures no
identifiable imagery~\cite{zhao2018through, guan2020through, adib2015smart}.
This makes it a natural sensor for privacy-preserving human sensing,
including gait, gesture, activity, fall and pose
recognition~\cite{lien2016soli, singh2019radhar, zhao2018through,
lee2023hupr}. Each of these  applications requires a downstream model trained
on real micro-Doppler~\cite{chen2006micro}, the phase modulation a moving
body imposes on the return. However, each is bottlenecked by the same data
scarcity, because public radar data lags orders of magnitude behind vision
and language corpora. Brute-force collection cannot close that gap:
radar sensors are rarer than cameras, and where a camera is point-and-shoot,
every radar capture needs a rig calibrated and synchronized to reference cameras
or LiDAR.

\begin{figure}[t]
  \centering
  \includegraphics[width=\linewidth]{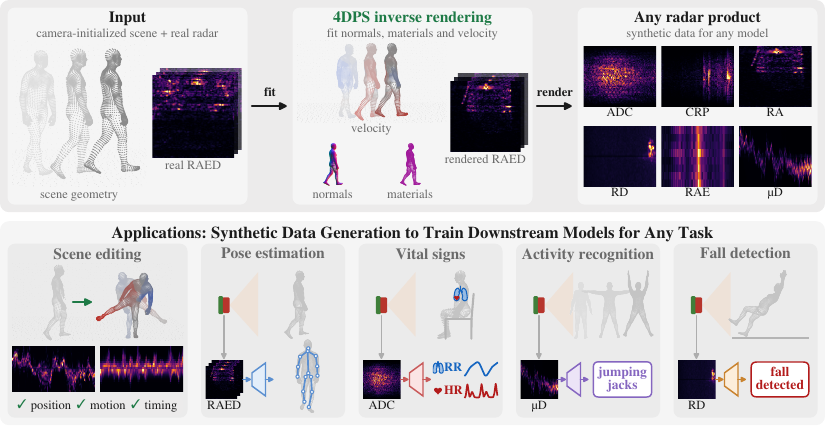}
  \caption{\textbf{4DPS fits a moving scene to a real radar capture.} Top:
  from a camera-initialized scene of oriented points, 4DPS fits materials,
  normals and velocities (color: radial velocity) until the rendered
  range--azimuth--elevation--Doppler tensor matches the recording. Bottom:
  the fitted scene can be edited (App.~\ref{sec:supp:editing}) and rendered as training data for pose,
  vital-sign, activity and fall models.}
  \label{fig:teaser}
\end{figure}

Synthesis is the next best alternative, but every existing method leaves a
sim2real gap, because none fit to a real capture of a moving body.
Motion-driven simulators~\cite{ram2008simulation,
vishwakarma2022simhumalator, chen2023rfgenesis} carry geometry, materials and motion explicitly
but run forward only, so their parameters are set by an operator and never
optimized against a real capture. Learned generators such as
RF-Diffusion~\cite{chi2024rfdiffusion} are trained on the same scarce data and keep the
scene implicit in network weights, so there is nothing to fit or edit. Radar inverse renderers do fit
real captures; mmIR~\cite{mmir}, 3DPS~\cite{3dps} and differentiable ray
tracers~\cite{sionna, hofmann2025inverse, chen2025invertwin,
chen2023diffsbr} recover explicit materials and normals and render range,
azimuth and elevation. However, all currently assume a static scene and
evaluate one phase per point per frame so they have no slow-time axis (the
chirp-to-chirp axis along which Doppler is measured) and nothing along which
velocity could be defined. What is missing is a renderer that fits an
explicit \emph{moving} scene to a real capture.

We present 4D Point Splatting (4DPS), a physics-based differentiable
renderer that fits an explicit moving scene to a real capture, with velocity
as an estimated and differentiable parameter. Like prior radar
renderers~\cite{mmir, 3dps, radarsim}, we initialize geometry from another
sensor, here the capture's synchronized camera. We represent the scene as
oriented points, each carrying an ITU-R~P.2040 material~\cite{itu_p2040}, a
surface normal and, for the body, a velocity, and we splat each point into
range and Doppler bins. The camera fixes where each surface is present but not how
strongly it reflects at millimeter wavelengths. That strength depends on the
surface's material and on the angle between its normal and the radar, and it
sets the amplitude of every return. We therefore fit the materials, normals
and velocities by gradient descent until the rendered return matches the
recording (Fig.~\ref{fig:teaser}, top). On HuPR, fitting raises the clutter-removed range--azimuth
correlation from \rCorrUnfitRAc{} for the camera-initialized render to
\rCorrRAc{}. Our output is the raw complex return that a real radar records,
for every chirp, antenna pair and range bin. We therefore apply the same
digital signal processing (DSP) to rendered and real data, and obtain any
representation, from range profiles to micro-Doppler, for any downstream
model (Fig.~\ref{fig:teaser}, bottom). We validate each representation on two datasets with different array
geometries and downstream models.

Introducing an explicit velocity into the forward model raises three
challenges. The first challenge is the cost: existing differentiable radar
renderers output one chirp per frame, so rendering Doppler would repeat that
render for every chirp, a cost that grows linearly with the chirp count and
rules out corpus-scale fitting. 4DPS instead splats each point into Doppler
bins as well as range bins (Fig.~\ref{fig:overview}). A moving point's return across
a frame's chirps is a single tone, so its Doppler point spread function has a
closed form. Each point therefore deposits into seven Doppler bins instead of
being rendered 64 times. Second, ambiguity: velocity enters the forward model only through
a phase term that wraps every $2\pi$, so velocities one Doppler ambiguity
apart (9\,m/s on HuPR) render identically, and gradient descent from an
arbitrary start lands in the nearest minimum rather than the true minimum. To address this, we initialize the joint velocities from the video's kinematics and
learn only a residual, bounded by each joint's own speed. Third,
supervision: one loss must train both the static scene and the person, who
holds a small fraction of a frame's energy. A loss over the whole frame lets
a large relative error on the person hide behind the scene's return, and a
loss on the person alone gives the scene no gradient. We therefore compare
the range--azimuth--elevation magnitude at two slow-time views: each frame's
first chirp, which carries the static scene, and its Doppler planes, which
carry the motion. Each view is compared over the whole frame and over a
window on the person (Eq.~\ref{eq:objective}). Solving these three problems lets us fit entire corpora for the
first time, with no manual input. We fit every usable sequence of both datasets,
\rNseqHuPR{} on HuPR and \rNseqRT{} on RT-Pose, at about \rMinHuPR{} and
\rMinRT{} minutes a sequence on one H100.

We set the new SOTA in 4D range--azimuth--elevation--Doppler (RAED) tensor rendering and reconstruction, with median Pearson
correlations of \rCorrRA{} on range--azimuth and \rCorrMD{} on micro-Doppler
on HuPR, and \rCorrRART{} and \rCorrMDRT{} on RT-Pose
(Table~\ref{tab:fidelity}). To test whether a model can learn from these
renders, we train each dataset's official pose model on real or on rendered
data, four seeds each, and test it on both (Tables~\ref{tab:tstr} and~\ref{tab:supp_tstr_full}). Trained on
our renders alone, the models keep \rRetention{} of their real-trained
accuracy on HuPR and \rRetentionRT{} on RT-Pose. By contrast, models trained
on RF-Genesis renders of the same motion fail on real data, at
\rRfgApTSTRMean{}\,AP on HuPR and \rRfgMPJPETSTRMean{}\,cm MPJPE on RT-Pose.
Because our rendered-trained models score far higher on renders than on
recordings, the gap that remains is realism rather than missing information.
A comparison of the two models' internal representations shows that they
compute the same features from the backbone onward (Sec.~\ref{sec:exp:repr}).

\textbf{Contributions.} In summary, we make the following contributions: 1)~4DPS, to our knowledge
the first differentiable renderer of human micro-Doppler: oriented points
carrying material, normal \emph{and} a velocity estimated per joint for the first
time, splatted in closed form into range and Doppler bins, rendering the raw
complex return so any representation follows; 2)~a range--Doppler point
splatting framework, with a closed-form Doppler point spread function and
Doppler-resolved supervision, that makes corpus-scale inverse rendering
tractable across multiple public corpora and different array geometries for
the first time; 3)~the first sim2real analysis
of radar inverse rendering on downstream models: the real/rendered
train--test matrix for radar human pose on two datasets with the official
models, and an interpretability analysis showing that rendered- and
real-trained models learn the same features; and 4)~release of the
renderer, the fitting chain, every fitted scene and rendered cube, the
mesh-derived RT-Pose labels, the retrained models and the evaluation suite,
on acceptance.

\section{Related Work}
\label{sec:related}

\begin{table}[t]
\centering
\caption{\textbf{Comparison of radar rendering and synthesis methods.} Signal:
CIR or raw ADC. Mat.: explicit materials; Vel.: velocity estimated; Doppler:
from subject motion; Real: validated on real captures; Edit: re-renderable
scene; Code: released. Radar ports of neural fields and Gaussian splatting
are omitted, because they supervise post-FFT power of a static scene.}
\label{tab:related}
\small
\setlength{\tabcolsep}{4pt}
\renewcommand{\arraystretch}{1.1}
\resizebox{\linewidth}{!}{
\begin{tabular}{lccccccccc}
\toprule
\textbf{Method} & \textbf{Repr.} & \textbf{Signal} & \textbf{Complex} &
\textbf{Mat.} & \textbf{Vel.} & \textbf{Doppler} &
\textbf{Real} & \textbf{Edit} & \textbf{Code} \\
\midrule
\multicolumn{10}{l}{\emph{Differentiable radar inverse rendering}} \\
mmIR~\cite{mmir}                            & Mesh   & ADC & \checkmark & \checkmark & -- & --         & \checkmark & \checkmark & \checkmark \\
3DPS~\cite{3dps}                            & Points & ADC & \checkmark & \checkmark & -- & --         & \checkmark & \checkmark & \checkmark \\
\midrule
\multicolumn{10}{l}{\emph{Human micro-Doppler simulation and synthesis}} \\
RF-Genesis~\cite{chen2023rfgenesis}         & Mesh     & ADC & \checkmark & --         & -- & \checkmark & \checkmark & --         & \checkmark \\
HybridSim~\cite{xiong2026hybridsim}         & Mesh     & ADC & \checkmark & \checkmark & -- & \checkmark & \checkmark & \checkmark & -- \\
\midrule
\textbf{4DPS (Ours)}                        & Points & ADC & \checkmark & \checkmark & \checkmark & \checkmark & \checkmark & \checkmark & \checkmark \\
\bottomrule
\end{tabular}
}
\end{table}

\textbf{Differentiable radar rendering.}
Radar-domain renderers fit explicit scene parameters to
captures. Differentiable ray
tracers~\cite{sionna, hofmann2025inverse, chen2025invertwin,
chen2023diffsbr} do so on synthetic or known-geometry scenes.
mmIR~\cite{mmir} fits the raw ADC of real captures and recovers materials,
normals and antenna patterns, and 3DPS~\cite{3dps} replaces mmIR's Monte
Carlo sampling with closed-form point splatting. Every one of these
renderers reconstructs a static scene. Its forward model evaluates one phase
per primitive per frame, so it has no slow time, the axis along which Doppler
and velocity are defined. 4DPS renders that axis in closed form and
estimates velocity, so it fits a moving scene (Table~\ref{tab:related}).

\textbf{Radar data synthesis.}
Radar ports of neural fields and Gaussian splatting~\cite{huang2024dart,
10.1145/3641519.3657510, kung2025radarsplat,
zhang2025rf4d, rafidashti2025neuradar} supervise on post-FFT
power and discard phase, so they cannot render the phase that encodes
micro-Doppler. DART~\cite{huang2024dart} is the only one with a Doppler axis,
and its Doppler comes from ego-motion, not from an articulated subject. Motion-driven
simulators~\cite{ram2008simulation,
vishwakarma2022simhumalator, chen2023rfgenesis} produce human micro-Doppler but run forward
only, so their residual against a real capture is either never quantified or never minimized.
mmGPE~\cite{xue2023mmgpe} narrows that residual with a learned refiner on the
simulator's output, without fitting the scene. Learned
generators~\cite{chi2024rfdiffusion} leave the scene implicit in network
weights, where nothing can be fitted or edited. Concurrent with this work,
HybridSim~\cite{xiong2026hybridsim} ray traces a physics-learning hybrid over
eight sequences from one single-chip radar, trained end to end with the
downstream task it feeds. Solving both at once is a weaker requirement than ours, because a renderer
supervised by the task's loss need only produce features that task finds
useful. We instead fit the radar return itself and score it against the
recording, so our renders must stand on their own and then transfer to a downstream model that never saw them during fitting. We do not
carry it as a baseline, because it releases no code and its captures,
collected for mmMesh~\cite{xue2021mmmesh}, are unreleased. 4DPS fits an explicit scene,
velocity included, to a real capture.

\section{Method}
\label{sec:methods}

\begin{figure}[t]
  \centering
  \includegraphics[width=\linewidth]{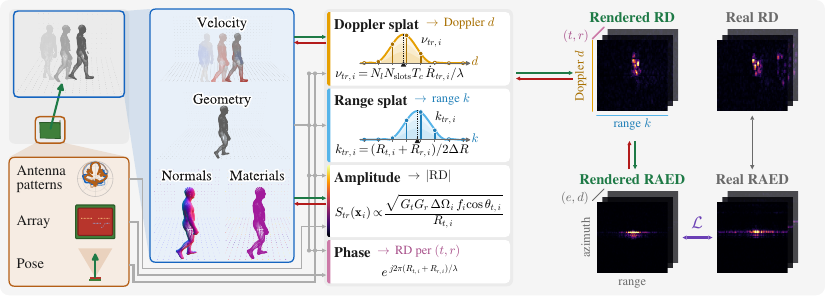}
  \caption{\textbf{4DPS overview.} Camera-initialized oriented point sets
  for a static scene and a moving person carry normals, materials and, for
  the person, velocities. For every antenna pair, 4DPS splats each point into
  its Doppler and range bins with an amplitude and a carrier phase, forming
  the RAED tensor that $\mathcal{L}$ compares with the recording. Green:
  forward pass; red: gradients; gray: fixed inputs.}
  \label{fig:overview}
\end{figure}

4DPS starts from two camera-initialized point sets, a static scene and a moving
person (Fig.~\ref{fig:overview}; RT-Pose's scene also uses its LiDAR). Each scene point carries its own
ITU-R~P.2040 material and normal. Each person point carries its own normal,
the whole person shares one material, and its points move with 14 joint
velocities skinned onto them. 4DPS renders the complex return of a TDM MIMO
FMCW radar, for every chirp, transmitter--receiver pair and range bin, by
splatting both point sets in custom CUDA kernels. Static points are splatted
along range only, and moving points along both range and Doppler.
Sec.~\ref{sec:methods:fwd} states the differentiable radar rendering model,
from which Sec.~\ref{sec:methods:splat} derives the splatting and Sec.~\ref{sec:methods:objective} the objective and its learned
parameters.

\subsection{Radar Forward Model}
\label{sec:methods:fwd}

Consider an FMCW radar with $N_{\text{TX}}$ time-division-multiplexed
transmitters and $N_{\text{RX}}$ receivers, carrier wavelength $\lambda$,
range bin width $\Delta R$, $K = 256$ range bins and $N_l = 64$ loops per
frame. In each loop every transmitter fires one chirp in its own slot, so with
$N_{\text{slots}}$ slots and a chirp interval $T_c$ the loop period is
$N_{\text{slots}}T_c$ (216\,$\mu$s on HuPR, 780\,$\mu$s on the cascade).

\textbf{The rendering equation.} Following prior radar inverse
renderers~\cite{mmir, 3dps}, we take the bistatic radar equation~\cite{skolnik2001introduction}, replace
the cross-section by a spatially varying BSDF $f$ and integrate over the
receiver-centered hemisphere $\Omega_{rx}$, giving the received power as a
rendering equation,
\begin{equation}
  P_r \;=\; \frac{P_t\,\lambda^2}{(4\pi)^2}
  \int_{\Omega_{rx}}
  \frac{G_t(\boldsymbol{\omega}_t)\,G_r(\boldsymbol{\omega}_r)\,
        f(\boldsymbol{\omega}_t,\mathbf{x},\boldsymbol{\omega}_r)\,
        \cos\theta_t\,V(\mathbf{x})}{R_t^2(\mathbf{x})}\,\mathrm{d}\omega_r,
  \label{eq:rendering_eqn}
\end{equation}
with $G_t, G_r$ the antenna gains, $R_t$ the path length to the transmitter,
$\cos\theta_t = |\mathbf{n}\cdot\boldsymbol{\omega}_t|$ and $V\in\{0,1\}$
visibility. We discretize the hemisphere on an oriented point set
$\{(\mathbf{x}_i,\mathbf{n}_i,\Delta\Omega_i,\boldsymbol{\theta}_i,
\mathbf{v}_i)\}_{i=1}^{N}$, each point subtending solid angle
$\Delta\Omega_i$ at the receiver and carrying a material, with a velocity
$\mathbf{v}_i$ on person points only. The Riemann sum cancels the surface
Jacobian against $\Delta\Omega_i$ and leaves a sum of per-point amplitudes,
carrier phase attached,
\begin{equation}
  S_{tr}(\mathbf{x}_i) \;=\; C\,\frac{\sqrt{G_t(\boldsymbol{\omega}_{t,i})}}{R_{t,i}}\,
  \sqrt{G_r(\boldsymbol{\omega}_{r,i})\,\Delta\Omega_i}\,
  \sqrt{f_i\cos\theta_{t,i}}\;e^{\,j2\pi(R_{t,i}+R_{r,i})/\lambda},
  \label{eq:amplitude}
\end{equation}
with $C$ a constant that the learned gain absorbs
(App.~\ref{sec:supp:derivation}). We set $\Delta\Omega_i = 1$ for every point,
because a cosine solid-angle weight darkened the limbs' oblique sides, and the
learned gains and materials carry the level. Nothing is sampled. Each point contributes one path per
transmitter--receiver pair.

\subsection{Splatting Static and Moving Points}
\label{sec:methods:splat}

\textbf{Static scene: range splat.} A renderer of one chirp can push the
range FFT inside the point sum, so that each point becomes the Hann window's
known response at its fractional range bin and the range profile is splatted
instead of simulated~\cite{3dps}, and we use it for the static scene. A
static point returns the same sample in every chirp, and the real front end
removes that slow-time mean before its Doppler transform, so the static return
lies in no Doppler plane and is supervised through each frame's first chirp
(Sec.~\ref{sec:methods:objective}).

\textbf{Moving person: range--Doppler splat.} A per-chirp renderer never
forms the slow-time axis, the $N_l$ loops of a frame along which Doppler is
measured. For the person we therefore splat along slow time as well
(Fig.~\ref{fig:overview}), because the DFT over loops is linear in a path's
slow-time samples just as the range FFT is linear in its fast-time samples.
Transmitter $t$ samples a given path once a loop, so the path's slow-time
samples are a single tone. A point whose bistatic path length changes at the
rate $\dot{R}_{tr,i} = \mathbf{v}_i\cdot(\mathbf{u}_{t,i}+\mathbf{u}_{r,i})$,
with $\mathbf{u}_{t,i},\mathbf{u}_{r,i}$ the unit vectors from the antennas
to the point, therefore sits at the fractional Doppler bin $\nu_{tr,i} =
N_l\,N_{\text{slots}}\,T_c\,\dot{R}_{tr,i}/\lambda$, because consecutive
loops of one transmitter are one loop period apart. The rendered
range--Doppler product of one antenna pair is a two-axis splat,
\begin{equation}
  \mathrm{RD}_{tr}[d,k] \;=\; g \sum_{i=1}^{N}
  b_i\,V_i\,S_{tr}(\mathbf{x}_i)\,e^{\,j\phi_{t,i}}\,
  \Phi_{K}(k-k_{tr,i})\,\Phi_{N_l}(d-\nu_{tr,i}),
  \label{eq:doppler_splat}
\end{equation}
with $g$ a learned receiver gain per radar, $b_i$ the person's learned
per-joint scattering gains skinned to point $i$ (1 for scene points), $V_i$
the point's visibility $V$ in that frame, $k_{tr,i} = (R_{t,i}+R_{r,i})/(2\Delta R)$
the fractional range bin, $\phi_{t,i}$ the phase the transmitter's own slot
contributes, and $\Phi_{K}, \Phi_{N_l}$ the Hann window's exact response on
the two axes. $\Phi_{N_l}$ includes the slow-time mean removal, so static points deposit nothing here. We
truncate $\Phi_{K}$ to $L = 15$ range taps and $\Phi_{N_l}$ to 7 Doppler taps,
so a person path costs 105 accumulations instead of a $64\times256$ cube per
antenna pair, and its backward pass costs the same.
App.~\ref{sec:supp:doppler_closed_form} derives this closed form from the
per-chirp signal it replaces.

\textbf{Material model.} The BSDF $f_i$ is a closed-form Jones
BSDF~\cite{wei2024csvbsdf, 3dps}, evaluated from a six-parameter ITU-R~P.2040 material
vector~\cite{itu_p2040}: a coherent specular lobe and a small-perturbation
lobe, blended by the ITU roughness gates and resolved in a local $s/p$
polarization basis (App.~\ref{sec:supp:bsdf}). The scene's materials are per
point, while the whole person shares one six-vector, because per-point
materials let the fit darken the limbs
(App.~\ref{sec:supp:training}).

\textbf{Visibility and occlusion.} The binary $V$ of
Eq.~\eqref{eq:rendering_eqn} is a per-frame ray cast that runs in both directions. A body point is
switched off when its segment to the radar is blocked by the posed mesh or by
the scene. A scene point is switched off
for the frames in which the body stands between it and the radar
(App.~\ref{sec:supp:preproc}). This matters more here than for a static renderer, because the shadow moves
with the person. An unmasked scene point renders a return that stays constant
over the frame's loops while the recording interrupts it, and the
zero-Doppler plane is most sensitive to that error.

\textbf{Noise model.} A clutter-removed range--Doppler map is mostly
empty, and its empty cells sit at the receiver's noise floor instead of zero. A renderer that models no noise predicts exactly zero there, so
an $L_2$ on linear magnitude would charge the fit for every empty cell and
the fit would answer by manufacturing power. This is a Doppler problem,
because a range--azimuth image is filled by the static scene
while most of a resolved range--Doppler product is floor. We therefore make
the floor $\mathcal{N}(d,k)$ part of the forward model, a log scale and a
range shape over a prior taken from the recording's own off-person power,
added to the rendered power wherever magnitudes are compared. It enters the
loss only through the two plane terms, which fit its level, and the
unoccupied range--Doppler level of
Table~\ref{tab:fidelity} ($L_U = \rCorrRDfloorU{}$, \rCorrRDfloorURT{} on the
cascade) tests whether it is right.

\textbf{Benefits of 4D Point Splatting.} \emph{Render any DSP representation:}
Eq.~\eqref{eq:doppler_splat} renders the raw complex return a real radar
records, so the range, azimuth, elevation and Doppler FFTs of the real data's DSP turn one fitted scene into any representation a downstream model reads
(App.~\ref{sec:supp:derivation}).
 \emph{Differentiable velocity:} $\nu_{tr,i}$ depends on the point's velocity, so the
Doppler axis carries gradient through the skinning to the joint velocities, and the radar, not the camera, settles
what the limbs did. \emph{Corpus-scale:} that per-path cost is independent of $N_l$, so a whole corpus is affordable
(Sec.~\ref{sec:methods:objective}).

\subsection{Optimization and Implementation}
\label{sec:methods:objective}

Our objective function evaluates different projections of the linear RAED
magnitude tensor:
\begin{equation}
  \mathcal{L} \;=\; \underbrace{\mathcal{L}_{\mathrm{RAE},0}
  + \mathcal{L}_{\mathrm{RAE},\mathrm{pl}}}_{\text{global}}
  \;+\; \underbrace{\mathcal{L}^{\mathrm{win}}_{\mathrm{RAE},0}
  + \mathcal{L}^{\mathrm{win}}_{\mathrm{RAE},\mathrm{pl}}}_{\text{windowed on the person}},
  \label{eq:objective}
\end{equation}
a normalized $L_2$ on $|\mathrm{RAE}|$ (App.~\ref{sec:supp:training}),
taken at two slow-time views and two scopes.
The first-chirp term $\mathcal{L}_{\mathrm{RAE},0}$ keeps the static clutter
return, so it is the only term that reaches the whole static scene. Because it is a magnitude, it supervises ADC amplitude and
inter-element phase without fitting absolute phase noise. The plane term
$\mathcal{L}_{\mathrm{RAE},\mathrm{pl}}$ supervises the 63 non-DC Doppler
planes, which only the moving person reaches, so a velocity error that leaves
the range profile intact still moves energy between planes. The windowed pair restricts both to the person's spatial
window and normalizes over it, because a frame-wide divisor lets a large
relative error on the person hide behind the scene's energy. Those cells
hold the occupied range--Doppler region and the micro-Doppler signature. All four terms apply FFTs sized by the radar's virtual
aperture and chirp count, irrespective of the downstream model's beamformer,
unlike HybridSim~\cite{xiong2026hybridsim}, which trains its renderer
end-to-end with that model. 

\phantomsection
\textbf{What we optimize.}
\label{sec:methods:learned}
For the static scene we optimize per-point materials and
normals. For the person we optimize per-point normals, one material
six-vector and 14 joint velocities, each the video's kinematic velocity times
a learned per-joint scale plus a bounded, band-limited residual, skinned to
the points so that a joint the video calls still stays still. For the radar
we optimize its gains and the noise floor (Table~\ref{tab:learned}).

\phantomsection
\textbf{Scale and runtime.}
\label{sec:methods:runtime}
A chirp is rendered once per physical transmitter--receiver pair, so fitting
HuPR is $\rNseqHuPR{} \times 600 \times 64 \times 24$ chirps and RT-Pose
$\rNseqRT{} \times 299 \times 64 \times 192$, for a total of \rChirpsTotal{}
chirps. Each is fitted for 400 epochs, so the fit renders \rChirpRenders{}
chirps forward and as many again backward, without ever forming a slow-time
cube. That is \rScaleVsThreeDPS{} times the data 3DPS~\cite{3dps} fits, at
\rGPUh{} GPU-hours on H100s (App.~\ref{sec:supp:runtime}).

\section{Evaluation}
\label{sec:experiments}
\label{sec:exp:setup}
\label{sec:exp:dataset}
\label{sec:exp:preproc}
\label{sec:exp:protocol}

Sec.~\ref{sec:exp:fidelity} fits \rNseqHuPR{} sequences from
HuPR~\cite{lee2023hupr} and \rNseqRT{} from RT-Pose~\cite{rtpose}, two
public human-sensing corpora that retain slow time and a colocated camera.
HuPR records raw ADC from two orthogonal single-chip radars, and RT-Pose
from one cascade (App.~\ref{sec:supp:preproc}). Sec.~\ref{sec:exp:tstr}
measures sim2real with the pose model each dataset released,
MSCSA-PRGCN~\cite{lee2023hupr} and HRRadarPose~\cite{rtpose}, retrained from
scratch for four seeds at equal optimizer steps. Each model keeps its own
metric, because the labels differ. HuPR's 2D image keypoints admit OKS AP but
no metric distance, and RT-Pose's 3D joints admit MPJPE in centimeters. RT-Pose's
joints come from the mesh that drives the render
(App.~\ref{sec:supp:downstream}).
No public simulator takes a real capture as input, so
RF-Genesis~\cite{chen2023rfgenesis} is driven forward from the same
registered motion (App.~\ref{sec:supp:rfgenesis}), and our unfit render is
the floor. Sec.~\ref{sec:exp:repr} compares real- and rendered-trained
models.

\begin{table}[t]
\centering
\caption{\textbf{Rendering fidelity.} Medians over sequences (HuPR pools
both radars) of the Pearson correlation with the recording: range--angle
($\rho$; $\rho_c$ clutter-removed), occupied range--Doppler cells
($\rho_M$), and micro-Doppler centroid, spread and spectrogram ($\rho_\mu$,
$\rho_\sigma$, $\rho_S$); $L_U$: rendered-to-real noise-floor level. Bold:
best per section. App.~\ref{sec:supp:diagnostics} gives the other DSP
representations.}
\label{tab:fidelity}
\small
\setlength{\tabcolsep}{4pt}
\begin{tabular}{lcccccccc}
\toprule
& \multicolumn{2}{c}{\textbf{Range--angle}} & \multicolumn{2}{c}{\textbf{Range--Doppler}} & \multicolumn{3}{c}{\textbf{Micro-Doppler}} \\
\cmidrule(lr){2-3}\cmidrule(lr){4-5}\cmidrule(lr){6-8}
\textbf{Method} & $\rho\uparrow$ & $\rho_c\uparrow$ & $\rho_M\uparrow$ & $L_U\to 1$ & $\rho_\mu\uparrow$ & $\rho_\sigma\uparrow$ & $\rho_S\uparrow$ \\
\midrule
\multicolumn{8}{l}{\emph{HuPR}, two single-chip radars} \\
RF-Genesis~\cite{chen2023rfgenesis}  & \rRfgRA{} & \rRfgRAc{} & \rRfgRDocc{} & \rRfgRDfloorU{} & \rRfgMDc{} & \rRfgMDs{} & \rRfgMD{} \\
4DPS (unfit)         & \rCorrUnfit{} & \rCorrUnfitRAc{} & \rCorrUnfitRD{} & \rCorrRDfloorUnfitHu{} & \rCorrUnfitMDc{} & \rCorrUnfitMDs{} & \rCorrUnfitMD{} \\
\textbf{4DPS (Ours)} & \textbf{\rCorrRA{}} & \textbf{\rCorrRAc{}} & \textbf{\rCorrRDocc{}} & \textbf{\rCorrRDfloorUHu{}} & \textbf{\rCorrFitMDc{}} & \textbf{\rCorrFitMDs{}} & \textbf{\rCorrMD{}} \\
\midrule
\multicolumn{8}{l}{\emph{RT-Pose}, cascade, one-person sequences} \\
RF-Genesis           & \rRfgRART{} & \rRfgRAcRT{} & \rRfgRDoccRT{} & \rRfgRDfloorURT{} & \rRfgMDcRT{} & \rRfgMDsRT{} & \rRfgMDRT{} \\
4DPS (unfit)         & \rCorrUnfitRT{} & \rCorrUnfitRAcRT{} & \rCorrUnfitRDRT{} & \rCorrRDfloorUnfitRT{} & \rCorrUnfitMDcRT{} & \rCorrUnfitMDsRT{} & \rCorrUnfitMDRT{} \\
\textbf{4DPS (Ours)} & \textbf{\rCorrRART{}} & \textbf{\rCorrRAcRT{}} & \textbf{\rCorrRDoccRT{}} & \textbf{\rCorrRDfloorURT{}} & \textbf{\rCorrFitMDcRT{}} & \textbf{\rCorrFitMDsRT{}} & \textbf{\rCorrMDRT{}} \\
\bottomrule
\end{tabular}
\end{table}

\begin{figure}[t]
  \centering
  \includegraphics[width=\linewidth]{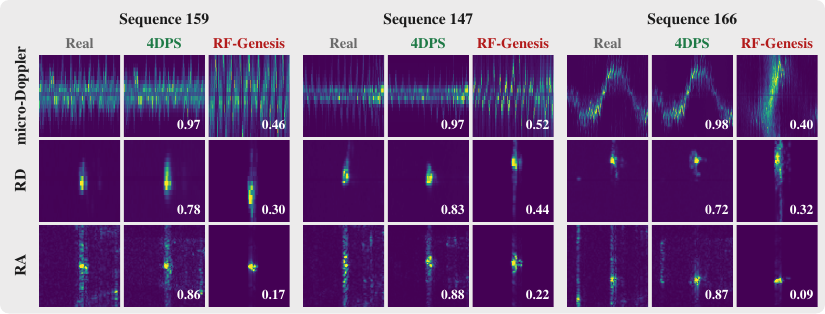}
  \caption{\textbf{Real, 4DPS and RF-Genesis on three RT-Pose test
  sequences.} Rows: micro-Doppler for the full sequence; typical
  range--Doppler frame; and typical range--azimuth frame; the number is the
  render's sequence-level correlation with the recording. Activities: 159
  stand and lift a leg, 147 stand and wave, 166 walk and wave. See
  App.~\ref{sec:supp:qualitative} for more details.}
  \label{fig:qualitative}
\end{figure}

\subsection{Rendering Fidelity}
\label{sec:exp:fidelity}

\emph{Does 4DPS reproduce the real capture, on every DSP representation?}
The fitted render reaches \rCorrRA{} range--azimuth and \rCorrMD{} micro-Doppler
correlation on HuPR, and \rCorrRART{} and \rCorrMDRT{} on RT-Pose
(Table~\ref{tab:fidelity}). Fig.~\ref{fig:qualitative} shows three RT-Pose
sequences beside the recording and RF-Genesis, and
Tables~\ref{tab:supp_fidelity_adc}--\ref{tab:supp_fidelity_rd_md} give every
column. RF-Genesis, driven by the same motion, reaches only \rRfgRA{} and
\rRfgMD{} on HuPR and \rRfgRART{} and \rRfgMDRT{} on RT-Pose. It renders the
person alone, with no room, so it misses the static scene that fills a real
range--azimuth map, and it keeps the body's bulk motion (micro-Doppler centroid \rRfgMDc{} on HuPR) but loses the limbs' spread around it (\rRfgMDs{}
against \rCorrFitMDs{} fitted), below even our unfit render, because its
\rRfgScatterers{} body scatterers are reassigned between ray casts and
nothing fits their motion to the recording. Our control is the unfit render, the same chain with the velocity
field trained for zero epochs, which reaches \rCorrUnfit{}, \rCorrUnfitMD{},
\rCorrUnfitRT{} and \rCorrUnfitMDRT{}. In a paired, one-sided comparison, the control scores lower on every
sequence, for both representations and datasets. The one exception is
range--azimuth on one HuPR sequence's vertical board. Camera motion alone already tracks the body's bulk
radial movement, so the control keeps a Doppler-centroid correlation of
\rCorrUnfitMDc{} on HuPR and \rCorrUnfitMDcRT{} on RT-Pose. What it loses is
the spread around that centroid, where the limbs' micro-motion sits. Its
spread correlation falls to \rCorrUnfitMDs{} on HuPR and \rCorrUnfitMDsRT{}
on RT-Pose, against \rCorrFitMDs{} and \rCorrFitMDsRT{} for the fitted render. The radar fit therefore recovers the motion the video
gets wrong. The cascade is the harder case, because it resolves elevation, the
axis the scaffold reconstructs least well. The still
recordings score lowest.

\begin{table}[t]
\centering
\caption{\textbf{Pose estimation downstream model training: real vs
rendered data.} On RT-Pose we train HRRadarPose~\cite{rtpose}, and on HuPR
MSCSA-PRGCN~\cite{lee2023hupr}. Every row is tested on real data, so the
rendered-trained rows measure sim-to-real transfer. Mean (std) over four
seeds trained from scratch. Table~\ref{tab:supp_tstr_full} gives the full
train--test matrix.}
\label{tab:tstr}
\small
\setlength{\tabcolsep}{2.8pt}
\begin{adjustbox}{max width=\linewidth}
\begin{tabular}{llcccccc}
\toprule
& & \multicolumn{3}{c}{\textbf{RT-Pose} (cm $\downarrow$)} & \multicolumn{3}{c}{\textbf{HuPR} (AP $\uparrow$)} \\
\cmidrule(lr){3-5}\cmidrule(lr){6-8}
\textbf{Train} & \textbf{Test} & MPJPE & MRPE & Abs-MPJPE & AP & AP$_{50}$ & AP$_{75}$ \\
\midrule
Real & Real & \rMPJPEReal{} & \rTstrRTMRPERR{} & \rTstrRTAbsMPJPERR{} & \rApReal{} & \rTstrHuAPfiftyRR{} & \rTstrHuAPsevfiveRR{} \\
\midrule
RF-Genesis & Real & \rTstrRfgRTMPJPESR{} & \rTstrRfgRTMRPESR{} & \rTstrRfgRTAbsMPJPESR{} & \rTstrRfgHuAPSR{} & \rTstrRfgHuAPfiftySR{} & \rTstrRfgHuAPsevfiveSR{} \\
4DPS & Real & \rMPJPETSTR{} & \rTstrRTMRPESR{} & \rTstrRTAbsMPJPESR{} & \rApTSTR{} & \rTstrHuAPfiftySR{} & \rTstrHuAPsevfiveSR{} \\
\bottomrule
\end{tabular}
\end{adjustbox}
\end{table}

\begin{figure}[t]
  \centering
  \includegraphics[width=\linewidth]{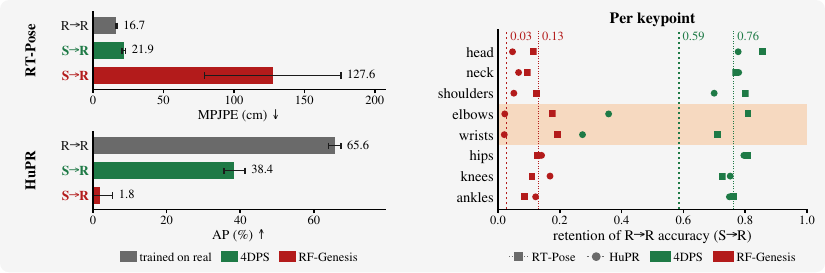}
  \caption{\textbf{The sim-to-real gap.} Left: real-data MPJPE on RT-Pose
  ($\downarrow$) and AP on HuPR ($\uparrow$) after training on real data,
  4DPS or RF-Genesis (four seeds, $\pm$1 std; bars on their true scale).
  Right: retention of real-trained accuracy per keypoint group
  (S$\to$R over R$\to$R, the inverse for MPJPE); on HuPR the 4DPS deficit
  sits in the arms (shaded). Table~\ref{tab:tstr} gives every metric.}
  \label{fig:downstream}
\end{figure}

\subsection{Training Pose Models on Rendered Data}
\label{sec:exp:tstr}

\emph{How much of real-trained accuracy survives training on renders?}
Every row of Table~\ref{tab:tstr} is tested on real data. Trained on 4DPS
renders alone, each model keeps \rRetention{} of its real-trained accuracy on
HuPR (\rApTSTR{} against \rApReal{} AP) and \rRetentionRT{} on RT-Pose
(\rMPJPETSTR{} against \rMPJPEReal{}\,cm). A model trained on RF-Genesis renders of the same motion transfers almost
nothing. It reaches \rTstrRfgHuAPSR{} AP on HuPR and
\rTstrRfgRTMPJPESR{}\,cm MPJPE on RT-Pose (Fig.~\ref{fig:downstream}, left).
Both renderers see the same motion, so the difference lies in what each
makes of it. An RF-Genesis render holds the person alone and carries little
of the limbs' micro-motion (Sec.~\ref{sec:exp:fidelity}), so a model trained
on it never learns to separate a moving body from the room around it. The
4DPS renders do carry the pose information a model needs. Tested on held-out
renders, a 4DPS-trained model scores \rApSelf{} AP and \rMPJPESelf{}\,cm,
close to the real-trained model on recordings
(Table~\ref{tab:supp_tstr_full}). Its drop
on recordings therefore comes from the remaining rendered--real
differences, which the model first meets at test time. Per
keypoint, the drop concentrates in the arms (Fig.~\ref{fig:downstream},
right). On HuPR the wrists fall from \rKpWristLRR{} and \rKpWristRRR{} AP to
\rKpWristLSR{} and \rKpWristRSR{}, while hips, knees and the head keep
\rKpRetTorsoLo{} to \rKpRetTorsoHi{} of their real-trained AP. The arms are
the body's smallest and fastest-moving parts, and their rendered motion
comes from the video's kinematics, corrected only within a bounded residual
(Sec.~\ref{sec:methods:learned}). HuPR's coarse azimuth resolution also gives
the fit little signal to correct them, so they depart most from the
recording.

HuPR is the harder downstream task of the two, although its renders score
higher on fidelity (Table~\ref{tab:fidelity}), for three reasons. i) One sequence gives the
renderer \rChirpsSeqHuPR{} chirps against RT-Pose's \rChirpsSeqRT{}, a factor
of \rChirpRatioHuPR{}. ii) Its radars also resolve the lateral axes far more
coarsely. Eight azimuth elements on a single chip put the broadside resolution
near \rAngResHuPR{}$^\circ$, against \rAngResRT{}$^\circ$ for the cascade's 86
slots. Each HuPR board resolves one axis only, azimuth on the horizontal and
elevation on the vertical, and carries \rNelHuPR{} elevation bins where the
cascade carries \rNelRT{}. The labels, meanwhile, are keypoints in a camera
image whose pixels subtend about \rCamResHuPR{}$^\circ$. The model must place
joints on a grid some \rLabelVsRadarHuPR{} times finer than its own sensor
resolves laterally. iii) Coarse angles make the camera-to-radar extrinsic
harder to estimate as well, and our fit is sensitive to that registration, as
prior radar inverse renderers are~\cite{mmir, 3dps}
(App.~\ref{sec:supp:limitations}).

\subsection{Comparing Real- and Rendered-Trained Models}
\label{sec:exp:repr}

\emph{Does a model trained on renders learn the same features as one trained
on real data?} We compare the two models' activations layer by layer with two standard similarity measures, orthogonal
Procrustes distance~\cite{williams2021shape} and debiased linear
CKA~\cite{kornblith2019cka, nguyen2021wide, murphy2024cka}, because no single
measure wins across settings~\cite{klabunde2025resi}. Each distance (Procrustes, $1-$CKA) is divided by that between two of five real-trained
models, which differ only in seed,
so 1.0 means as similar as two real-trained models, and we fixed $1.2$ as the
margin before running the analysis (Table~\ref{tab:repr}). The Procrustes
ratio stays between \rReprProcLo{} and \rReprProcHi{} at every layer on
RT-Pose and between \rReprHuProLo{} and \rReprHuProHi{} on HuPR, and CKA
clears the margin with confidence only at RT-Pose's first layer
(\rReprCkaTone{}), where the rendered and real inputs themselves differ most.
Training on renders therefore teaches the same features as training on real
data, consistent with Sec.~\ref{sec:exp:tstr}, where the rendered-trained
model's drop comes from rendered--real differences in its input. Features that look alike can still serve pose estimation differently, so
we also freeze each model's feature extractor and fit a new linear pose
readout on real data. On HuPR, features learned from renders give
\rReprHuProbeRendRatio{} times the keypoint error of real-trained features,
against \rReprHuProbeFloorRatio{} times for an untrained network. Therefore, they
carry most but not all of the pose information. At the
deepest layers even an untrained network scores about as close
($^\dagger$), so similarity there is no evidence either way.
App.~\ref{sec:supp:repr} gives every layer, the error-consistency analysis
and the protocol.

\begin{table}[t]
\centering
\caption{\textbf{Do rendered- and real-trained models learn the same
features?} Each cell is the real--rendered distance over the real--real
distance between seeds (1.0: as close as two real-trained models), with its
95\% interval; the readout columns are error ratios. Bold: the interval lies
above the $1.2\times$ margin fixed before the analysis. $^\dagger$:
uninformative, as an untrained network is about as close. HuPR's middle
layers are undefined (dead temporal merges); App.~\ref{sec:supp:repr} gives
every layer and the error-consistency analysis.}
\label{tab:repr}
\footnotesize
\setlength{\tabcolsep}{3pt}
\newcommand{\rcell}[2]{\shortstack{#1\\[1pt]{\tiny #2}}}
\begin{tabular}{l ccccc cc}
\toprule
& \multicolumn{5}{c}{\emph{Feature similarity}}
& \multicolumn{2}{c}{\emph{Linear readout}} \\
\cmidrule(lr){2-6}\cmidrule(lr){7-8}
& \shortstack{CKA\\first} & \shortstack{CKA\\middle} & \shortstack{CKA\\backbone}
& \shortstack{CKA\\logits} & \shortstack{Procrustes\\every layer}
& \shortstack{Rendered-\\trained} & \shortstack{Untrained\\backbone} \\
\midrule
RT-Pose
 & \textbf{\rReprCkaTone{}}, \rReprCkaTtwoa{}
 & \rReprCkaMidLo{}--\rReprCkaMidHi{}
 & \rReprCkaTfouraRatio{}$^\dagger$
 & \rReprCkaTheadRatio{}$^\dagger$
 & \rReprProcLo{}--\rReprProcHi{}
 & \rcell{\rProbeRatio{}}{\rProbeCi{}}
 & \rProbeRatioFloor{} \\
HuPR
 & \shortstack{\rReprHuCkaChirpratio{}\\\rReprHuCkaChirpVratio{}}
 & undefined
 & \shortstack{\rReprHuCkaEncratio{}$^\dagger$\\\rReprHuCkaVertratio{}$^\dagger$}
 & \rcell{\rReprHuCkaHeadratio{}}{\rReprHuCkaHeadci{}}
 & \rReprHuProLo{}--\rReprHuProHi{}
 & \rcell{\rReprHuProbeRendRatio{}}{\rReprHuProbeRendCi{}}
 & \rReprHuProbeFloorRatio{} \\
\bottomrule
\end{tabular}
\end{table}

\section{Conclusion}
\label{sec:conclusion}

We presented 4DPS, a physics-based differentiable renderer that fits an
explicit moving scene to a real radar capture. Camera-initialized oriented
points carry an ITU-R~P.2040 material, a normal and a velocity, and are
splatted in closed form into range and Doppler bins on the TDM timing of a
MIMO array. The closed-form Doppler splat makes the fit cheap enough to run
unattended over all \rNseqHuPR{} HuPR and \rNseqRT{} RT-Pose sequences, and
because 4DPS renders the raw complex return, the real data's DSP turns it into
any representation. 4DPS sets the state of the art on range--azimuth,
range--Doppler and micro-Doppler. On HuPR it reaches median correlations of
\rCorrRAc{} on clutter-removed range--azimuth and \rCorrMD{} on micro-Doppler,
against \rRfgRAc{} and \rRfgMD{} for RF-Genesis driven by the same motion.
Pose models trained on our renders alone keep \rRetention{} of their
real-trained accuracy on HuPR and \rRetentionRT{} on RT-Pose, while models
trained on RF-Genesis renders do not transfer. The rendered- and real-trained
models also compute the same features, so the remaining gap comes from
rendered--real differences in their input.
\textbf{Limitations and future work.} Velocity stays within a bounded
residual of the video kinematics. The background is static and rendered along
single-bounce paths, and we model one person per sequence. RT-Pose's labels
come from the body mesh that drives the render, and no public simulator takes
a capture as input, so our baselines run forward
(App.~\ref{sec:supp:limitations}). Scoring edited scenes against real
recordings is the next step (App.~\ref{sec:supp:editing}).

\section*{Reproducibility Statement}
Code, configurations, the fitted checkpoints of every sequence and the
scripts behind every table will be released. Both datasets are public
(HuPR~\cite{lee2023hupr}, RT-Pose~\cite{rtpose}). The forward model is
stated in Sec.~\ref{sec:methods} with every learned quantity in
Table~\ref{tab:learned}; the preprocessing chain, the runtime design and the
downstream protocol are in the supplement
(Secs.~\ref{sec:supp:preproc}, \ref{sec:supp:runtime_gates},
\ref{sec:supp:training}, \ref{sec:supp:repr}).

\section*{Use of AI Statement}
We used large language models to implement and edit code, clean and reformat
the datasets, assist in writing, and give feedback on the research
methodology and experiments. The authors reviewed all AI-assisted work and
take full responsibility for the content of this paper.

\bibliography{main}

\begin{thebibliography}{66}
\providecommand{\natexlab}[1]{#1}
\providecommand{\url}[1]{\texttt{#1}}
\expandafter\ifx\csname urlstyle\endcsname\relax
  \providecommand{\doi}[1]{doi: #1}\else
  \providecommand{\doi}{doi: \begingroup \urlstyle{rm}\Url}\fi

\bibitem[Adib et~al.(2015)Adib, Mao, Kabelac, Katabi, and
  Miller]{adib2015smart}
Fadel Adib, Hongzi Mao, Zachary Kabelac, Dina Katabi, and Robert~C. Miller.
\newblock Smart Homes That Monitor Breathing and Heart Rate.
\newblock In \emph{Proceedings of the 33rd Annual ACM Conference on Human
  Factors in Computing Systems}, CHI '15, pp.\  837–846, New York, NY, USA,
  2015. Association for Computing Machinery.
\newblock ISBN 9781450331456.
\newblock \doi{10.1145/2702123.2702200}.
\newblock URL \url{https://doi.org/10.1145/2702123.2702200}.

\bibitem[An \& Ogras(2021)An and Ogras]{mars}
Sizhe An and Umit~Y. Ogras.
\newblock MARS: mmWave-Based Assistive Rehabilitation System for Smart Healthcare.
\newblock \emph{ACM Trans. Embed. Comput. Syst.}, 20\penalty0 (5s), September
  2021.
\newblock ISSN 1539-9087.
\newblock \doi{10.1145/3477003}.
\newblock URL \url{https://doi.org/10.1145/3477003}.

\bibitem[An et~al.(2022)An, Li, and Ogras]{mri}
Sizhe An, Yin Li, and Umit Ogras.
\newblock mRI: Multi-Modal 3D Human Pose Estimation Dataset Using mmWave, RGB-D, and Inertial Sensors.
\newblock In S.~Koyejo, S.~Mohamed, A.~Agarwal, D.~Belgrave, K.~Cho, and A.~Oh
  (eds.), \emph{Advances in Neural Information Processing Systems}, volume~35,
  pp.\  27414--27426. Curran Associates, Inc., 2022.
\newblock \doi{10.52202/068431-1988}.
\newblock URL
  \url{https://proceedings.neurips.cc/paper_files/paper/2022/file/af9c9c6d2da701da5a0acf91ec217815-Paper-Datasets_and_Benchmarks.pdf}.

\bibitem[Armouti et~al.(2026{\natexlab{a}})Armouti, Gao, and Nandakumar]{3dps}
Adnan Armouti, Yixuan Gao, and Rajalakshmi Nandakumar.
\newblock 3D Point Splatting for mmWave Radar Novel View Synthesis, 2026{\natexlab{a}}.
\newblock URL \url{https://arxiv.org/abs/2609.11894}.

\bibitem[Armouti et~al.(2026{\natexlab{b}})Armouti, Gao, and Nandakumar]{mmir}
Adnan Armouti, Yixuan Gao, and Rajalakshmi Nandakumar.
\newblock mmIR: Frequency-Space Inverse Rendering for 3D Millimeter-Wave Radar ADC Synthesis.
\newblock In Paolo Favaro, Zuzana Kukelova, Atsuto Maki, Anna Rohrbach, Konrad
  Schindler, and Federico Tombari (eds.), \emph{Computer Vision -- ECCV 2026},
  pp.\  148--166, Cham, 2026{\natexlab{b}}. Springer Nature Switzerland.
\newblock ISBN 978-3-032-37359-5.

\bibitem[Borts et~al.(2024)Borts, Liang, Broedermann, Ramazzina, Walz,
  Palladin, Sun, Brueggemann, Sakaridis, Van~Gool, Bijelic, and
  Heide]{10.1145/3641519.3657510}
David Borts, Erich Liang, Tim Broedermann, Andrea Ramazzina, Stefanie Walz,
  Edoardo Palladin, Jipeng Sun, David Brueggemann, Christos Sakaridis, Luc
  Van~Gool, Mario Bijelic, and Felix Heide.
\newblock Radar Fields: Frequency-Space Neural Scene Representations for FMCW Radar.
\newblock In \emph{ACM SIGGRAPH 2024 Conference Papers}, SIGGRAPH '24, New
  York, NY, USA, 2024. Association for Computing Machinery.
\newblock ISBN 9798400705250.
\newblock \doi{10.1145/3641519.3657510}.
\newblock URL \url{https://doi.org/10.1145/3641519.3657510}.

\bibitem[Chen et~al.(2022)Chen, Wang, Zhu, Li, Chen, and Ye]{mmbody}
Anjun Chen, Xiangyu Wang, Shaohao Zhu, Yanxu Li, Jiming Chen, and Qi~Ye.
\newblock mmBody Benchmark: 3D Body Reconstruction Dataset and Analysis for Millimeter Wave Radar.
\newblock In \emph{Proceedings of the 30th ACM International Conference on
  Multimedia}, MM '22, pp.\  3501–3510, New York, NY, USA, 2022. Association
  for Computing Machinery.
\newblock ISBN 9781450392037.
\newblock \doi{10.1145/3503161.3548262}.
\newblock URL \url{https://doi.org/10.1145/3503161.3548262}.

\bibitem[Chen et~al.(2026)Chen, Huang, Prabhakara, Mummadi, Cong, Rowe,
  O'Toole, and Ramanan]{radarsim}
Chuhan Chen, Tianshu Huang, Akarsh Prabhakara, Chaithanya~Kumar Mummadi,
  Zhongxiao Cong, Anthony Rowe, Matthew O'Toole, and Deva Ramanan.
\newblock RadarSim: Simulating Single-Chip Radar via Multimodal Neural Fields.
\newblock In \emph{2026 International Conference on 3D Vision (3DV)}, pp.\
  1258--1268, 2026.
\newblock \doi{10.1109/3DV69130.2026.00124}.

\bibitem[Chen et~al.(2006)Chen, Li, Ho, and Wechsler]{chen2006micro}
V.C. Chen, F.~Li, S.-S. Ho, and H.~Wechsler.
\newblock Micro-Doppler Effect in Radar: Phenomenon, Model, and Simulation Study.
\newblock \emph{IEEE Transactions on Aerospace and Electronic Systems},
  42\penalty0 (1):\penalty0 2--21, 2006.
\newblock \doi{10.1109/TAES.2006.1603402}.

\bibitem[Chen \& Zhang(2024)Chen and Zhang]{chen2023rfgenesis}
Xingyu Chen and Xinyu Zhang.
\newblock RF Genesis: Zero-Shot Generalization of mmWave Sensing through Simulation-Based Data Synthesis and Generative Diffusion Models.
\newblock In \emph{Proceedings of the 21st ACM Conference on Embedded Networked
  Sensor Systems}, SenSys '23, pp.\  28–42, New York, NY, USA, 2024.
  Association for Computing Machinery.
\newblock ISBN 9798400704147.
\newblock \doi{10.1145/3625687.3625798}.
\newblock URL \url{https://doi.org/10.1145/3625687.3625798}.

\bibitem[Chen et~al.(2023)Chen, Zhang, Xia, Fang, Lu, and Li]{chen2023diffsbr}
Xingyu Chen, Xinyu Zhang, Qiyue Xia, Xinmin Fang, Chris~Xiaoxuan Lu, and
  Zhengxiong Li.
\newblock Differentiable Radio Frequency Ray Tracing for Millimeter-Wave Sensing, 2023.
\newblock URL \url{https://arxiv.org/abs/2311.13182}.

\bibitem[Chen et~al.(2025)Chen, Ding, Zheng, Fang, Zhang, Lu, and
  Li]{chen2025invertwin}
Xingyu Chen, Jianrong Ding, Kai Zheng, Xinmin Fang, Xinyu Zhang, Chris~Xiaoxuan
  Lu, and Zhengxiong Li.
\newblock InverTwin: Solving Inverse Problems via Differentiable Radio Frequency Digital Twin, 2025.
\newblock URL \url{https://arxiv.org/abs/2508.14204}.

\bibitem[Chi et~al.(2024)Chi, Yang, Wu, Xu, Gao, Liu, and Han]{chi2024rfdiffusion}
Guoxuan Chi, Zheng Yang, Chenshu Wu, Jingao Xu, Yuchong Gao, Yunhao Liu, and
  Tony~Xiao Han.
\newblock RF-Diffusion: Radio Signal Generation via Time-Frequency Diffusion.
\newblock In \emph{Proceedings of the 30th Annual International Conference on
  Mobile Computing and Networking}, ACM MobiCom '24, pp.\  77–92, New York, NY,
  USA, 2024. Association for Computing Machinery.
\newblock ISBN 9798400704895.
\newblock \doi{10.1145/3636534.3649348}.
\newblock URL \url{https://doi.org/10.1145/3636534.3649348}.

\bibitem[Cui et~al.(2023)Cui, Zhong, Wu, Shen, Dahnoun, and Zhao]{milipoint}
Han Cui, Shu Zhong, Jiacheng Wu, Zichao Shen, Naim Dahnoun, and Yiren Zhao.
\newblock MiliPoint: A Point Cloud Dataset for mmWave Radar.
\newblock In A.~Oh, T.~Naumann, A.~Globerson, K.~Saenko, M.~Hardt, and
  S.~Levine (eds.), \emph{Advances in Neural Information Processing Systems},
  volume~36, pp.\  62713--62726. Curran Associates, Inc., 2023.
\newblock \doi{10.52202/075280-2739}.
\newblock URL
  \url{https://proceedings.neurips.cc/paper_files/paper/2023/file/c60468eca9cd0b0083f0ff9d0aeb171a-Paper-Datasets_and_Benchmarks.pdf}.

\bibitem[Ding et~al.(2021)Ding, Denain, and Steinhardt]{ding2021grounding}
Frances Ding, Jean-Stanislas Denain, and Jacob Steinhardt.
\newblock Grounding Representation Similarity through Statistical Testing.
\newblock In M.~Ranzato, A.~Beygelzimer, Y.~Dauphin, P.S. Liang, and J.~Wortman
  Vaughan (eds.), \emph{Advances in Neural Information Processing Systems},
  volume~34, pp.\  1556--1568. Curran Associates, Inc., 2021.
\newblock URL
  \url{https://proceedings.neurips.cc/paper_files/paper/2021/file/0c0bf917c7942b5a08df71f9da626f97-Paper.pdf}.

\bibitem[Fan et~al.(2026)Fan, Zhou, Yang, Cui, Zhang, Xie, Yang, Lu, and
  Ding]{m4human}
Junqiao Fan, Yunjiao Zhou, Yizhuo Yang, Xinyuan Cui, Jiarui Zhang, Lihua Xie,
  Jianfei Yang, Chris~Xiaoxuan Lu, and Fangqiang Ding.
\newblock M4Human: A Large-Scale Multimodal mmWave Radar Benchmark for Human Mesh Reconstruction.
\newblock In \emph{Proceedings of the IEEE/CVF Conference on Computer Vision
  and Pattern Recognition (CVPR)}, pp.\  42836--42846, June 2026.

\bibitem[Gao et~al.(2021)Gao, Xing, Roy, and Liu]{rampcnn}
Xiangyu Gao, Guanbin Xing, Sumit Roy, and Hui Liu.
\newblock RAMP-CNN: A Novel Neural Network for Enhanced Automotive Radar Object Recognition.
\newblock \emph{IEEE Sensors Journal}, 21\penalty0 (4):\penalty0 5119--5132,
  2021.
\newblock \doi{10.1109/JSEN.2020.3036047}.

\bibitem[Geirhos et~al.(2020)Geirhos, Meding, and Wichmann]{geirhos2020error}
Robert Geirhos, Kristof Meding, and Felix~A. Wichmann.
\newblock Beyond Accuracy: Quantifying Trial-by-Trial Behaviour of CNNs and Humans by Measuring Error Consistency.
\newblock In H.~Larochelle, M.~Ranzato, R.~Hadsell, M.F. Balcan, and H.~Lin
  (eds.), \emph{Advances in Neural Information Processing Systems}, volume~33,
  pp.\  13890--13902. Curran Associates, Inc., 2020.
\newblock URL
  \url{https://proceedings.neurips.cc/paper_files/paper/2020/file/9f6992966d4c363ea0162a056cb45fe5-Paper.pdf}.

\bibitem[Guan et~al.(2020)Guan, Madani, Jog, Gupta, and
  Hassanieh]{guan2020through}
Junfeng Guan, Sohrab Madani, Suraj Jog, Saurabh Gupta, and Haitham Hassanieh.
\newblock Through Fog High-Resolution Imaging Using Millimeter Wave Radar.
\newblock In \emph{2020 IEEE/CVF Conference on Computer Vision and Pattern
  Recognition (CVPR)}, pp.\  11461--11470, 2020.
\newblock \doi{10.1109/CVPR42600.2020.01148}.

\bibitem[Guo et~al.(2022)Guo, Zou, Zuo, Wang, Ji, Li, and
  Cheng]{guo2022humanml3d}
Chuan Guo, Shihao Zou, Xinxin Zuo, Sen Wang, Wei Ji, Xingyu Li, and Li~Cheng.
\newblock Generating Diverse and Natural 3D Human Motions from Text.
\newblock In \emph{2022 IEEE/CVF Conference on Computer Vision and Pattern
  Recognition (CVPR)}, pp.\  5142--5151, 2022.
\newblock \doi{10.1109/CVPR52688.2022.00509}.

\bibitem[Guo \& Chen(2026)Guo and Chen]{dghmesh}
Rongxiao Guo and Qingchao Chen.
\newblock DGHMesh: A Large-Scale Dual-Radar mmWave Dataset and Generalization-Focused Benchmark for Human Mesh Reconstruction, 2026.
\newblock URL \url{https://arxiv.org/abs/2604.22827}.

\bibitem[Ho et~al.(2025)Ho, Cheng, Kuan, Jiang, Chai, Huang, Lin, and
  Hwang]{rtpose}
Yuan-Hao Ho, Jen-Hao Cheng, Sheng~Yao Kuan, Zhongyu Jiang, Wenhao Chai,
  Hsiang-Wei Huang, Chih-Lung Lin, and Jenq-Neng Hwang.
\newblock RT-Pose: A 4D Radar Tensor-Based 3D Human Pose Estimation and Localization Benchmark.
\newblock In Ale{\v{s}} Leonardis, Elisa Ricci, Stefan Roth, Olga Russakovsky,
  Torsten Sattler, and G{\"u}l Varol (eds.), \emph{Computer Vision -- ECCV
  2024}, pp.\  107--125, Cham, 2025. Springer Nature Switzerland.
\newblock ISBN 978-3-031-73036-8.

\bibitem[Hofmann et~al.(2025)Hofmann, Wirth, Bräunig, Ullmann, Vossiek,
  Weyrich, and Stamminger]{hofmann2025inverse}
Nikolai Hofmann, Vanessa Wirth, Johanna Bräunig, Ingrid Ullmann, Martin
  Vossiek, Tim Weyrich, and Marc Stamminger.
\newblock Inverse Rendering of Near-Field mmWave MIMO Radar for Material Reconstruction.
\newblock \emph{IEEE Journal of Microwaves}, 5\penalty0 (2):\penalty0 356--372,
  2025.
\newblock \doi{10.1109/JMW.2025.3535077}.

\bibitem[Hoydis et~al.(2023)Hoydis, Aoudia, Cammerer, Nimier-David, Binder,
  Marcus, and Keller]{sionna}
Jakob Hoydis, Faycal~Ait Aoudia, Sebastian Cammerer, Merlin Nimier-David,
  Nikolaus Binder, Guillermo Marcus, and Alexander Keller.
\newblock Sionna RT: Differentiable Ray Tracing for Radio Propagation Modeling.
\newblock In \emph{2023 IEEE Globecom Workshops (GC Wkshps)}, pp.\  317--321,
  2023.
\newblock \doi{10.1109/GCWkshps58843.2023.10465179}.

\bibitem[Huang et~al.(2024)Huang, Miller, Prabhakara, Jin, Laroia, Kolter, and
  Rowe]{huang2024dart}
Tianshu Huang, John Miller, Akarsh Prabhakara, Tao Jin, Tarana Laroia, Zico
  Kolter, and Anthony Rowe.
\newblock DART: Implicit Doppler Tomography for Radar Novel View Synthesis.
\newblock In \emph{2024 IEEE/CVF Conference on Computer Vision and Pattern
  Recognition (CVPR)}, pp.\  24118--24129. IEEE, 2024.

\bibitem[Huang et~al.(2025)Huang, Prabhakara, Chen, Karhade, Ramanan, O'Toole,
  and Rowe]{iq1m}
Tianshu Huang, Akarsh Prabhakara, Chuhan Chen, Jay Karhade, Deva Ramanan,
  Matthew O'Toole, and Anthony Rowe.
\newblock Towards Foundational Models for Single-Chip Radar.
\newblock In \emph{2025 IEEE/CVF International Conference on Computer Vision
  (ICCV)}, pp.\  24655--24665, 2025.
\newblock \doi{10.1109/ICCV51701.2025.02286}.

\bibitem[{ITU-R}(2023)]{itu_p2040}
{ITU-R}.
\newblock Effects of Building Materials and Structures on Radiowave Propagation above about 100 {MHz}.
\newblock Recommendation P.2040-3, International Telecommunication Union, aug
  2023.
\newblock URL \url{https://www.itu.int/rec/R-REC-P.2040/en}.

\bibitem[Klabunde et~al.(2025)Klabunde, Wald, Schumacher, Maier-Hein,
  Strohmaier, and Lemmerich]{klabunde2025resi}
Max Klabunde, Tassilo Wald, Tobias Schumacher, Klaus Maier-Hein, Markus
  Strohmaier, and Florian Lemmerich.
\newblock ReSi: A Comprehensive Benchmark for Representational Similarity Measures.
\newblock In Y.~Yue, A.~Garg, N.~Peng, F.~Sha, and R.~Yu (eds.),
  \emph{International Conference on Learning Representations}, volume 2025,
  pp.\  18701--18740, 2025.
\newblock URL
  \url{https://proceedings.iclr.cc/paper_files/paper/2025/file/2eef1f75516b0cfd3e944345e5f88c08-Paper-Conference.pdf}.

\bibitem[Kornblith et~al.(2019)Kornblith, Norouzi, Lee, and
  Hinton]{kornblith2019cka}
Simon Kornblith, Mohammad Norouzi, Honglak Lee, and Geoffrey Hinton.
\newblock Similarity of Neural Network Representations Revisited.
\newblock In Kamalika Chaudhuri and Ruslan Salakhutdinov (eds.),
  \emph{Proceedings of the 36th International Conference on Machine Learning},
  volume~97 of \emph{Proceedings of Machine Learning Research}, pp.\
  3519--3529. PMLR, 09--15 Jun 2019.
\newblock URL \url{https://proceedings.mlr.press/v97/kornblith19a.html}.

\bibitem[Kramer et~al.(2022)Kramer, Harlow, Williams, and
  Heckman]{kramer2022coloradar}
Andrew Kramer, Kyle Harlow, Christopher Williams, and Christoffer Heckman.
\newblock ColoRadar: The Direct 3D Millimeter Wave Radar Dataset.
\newblock \emph{Int. J. Rob. Res.}, 41\penalty0 (4):\penalty0 351–360, April
  2022.
\newblock ISSN 0278-3649.
\newblock \doi{10.1177/02783649211068535}.
\newblock URL \url{https://doi.org/10.1177/02783649211068535}.

\bibitem[Kung et~al.(2025)Kung, Harisha, Vasudevan, Eid, and
  Skinner]{kung2025radarsplat}
Pou-Chun Kung, Skanda Harisha, Ram Vasudevan, Aline Eid, and Katherine~A.
  Skinner.
\newblock RadarSplat: Radar Gaussian Splatting for High-Fidelity Data Synthesis and 3D Reconstruction of Autonomous Driving Scenes.
\newblock In \emph{2025 IEEE/CVF International Conference on Computer Vision
  (ICCV)}, pp.\  27596--27606, 2025.
\newblock \doi{10.1109/ICCV51701.2025.02562}.

\bibitem[Lee et~al.(2023)Lee, Kini, Peng, Ma, and Hwang]{lee2023hupr}
Shih-Po Lee, Niraj~Prakash Kini, Wen-Hsiao Peng, Ching-Wen Ma, and Jenq-Neng
  Hwang.
\newblock HuPR: A Benchmark for Human Pose Estimation Using Millimeter Wave Radar.
\newblock In \emph{2023 IEEE/CVF Winter Conference on Applications of Computer
  Vision (WACV)}, pp.\  5704--5713, 2023.
\newblock \doi{10.1109/WACV56688.2023.00567}.

\bibitem[Lien et~al.(2016)Lien, Gillian, Karagozler, Amihood, Schwesig, Olson,
  Raja, and Poupyrev]{lien2016soli}
Jaime Lien, Nicholas Gillian, M.~Emre Karagozler, Patrick Amihood, Carsten
  Schwesig, Erik Olson, Hakim Raja, and Ivan Poupyrev.
\newblock Soli: Ubiquitous Gesture Sensing with Millimeter Wave Radar.
\newblock \emph{ACM Trans. Graph.}, 35\penalty0 (4), July 2016.
\newblock ISSN 0730-0301.
\newblock \doi{10.1145/2897824.2925953}.
\newblock URL \url{https://doi.org/10.1145/2897824.2925953}.

\bibitem[Lim et~al.(2021)Lim, Markowitz, and Do]{radical}
Teck-Yian Lim, Spencer~A. Markowitz, and Minh~N. Do.
\newblock RaDICaL: A Synchronized FMCW Radar, Depth, IMU and RGB Camera Data Dataset with Low-Level FMCW Radar Signals.
\newblock \emph{IEEE Journal of Selected Topics in Signal Processing},
  15\penalty0 (4):\penalty0 941--953, 2021.
\newblock \doi{10.1109/JSTSP.2021.3061270}.

\bibitem[Loper et~al.(2015)Loper, Mahmood, Romero, Pons-Moll, and Black]{smpl}
Matthew Loper, Naureen Mahmood, Javier Romero, Gerard Pons-Moll, and Michael~J.
  Black.
\newblock SMPL: A Skinned Multi-Person Linear Model.
\newblock \emph{ACM Trans. Graph.}, 34\penalty0 (6), November 2015.
\newblock ISSN 0730-0301.
\newblock \doi{10.1145/2816795.2818013}.
\newblock URL \url{https://doi.org/10.1145/2816795.2818013}.

\bibitem[Mahmood et~al.(2019)Mahmood, Ghorbani, Troje, Pons-Moll, and
  Black]{mahmood2019amass}
Naureen Mahmood, Nima Ghorbani, Nikolaus~F. Troje, Gerard Pons-Moll, and
  Michael~J. Black.
\newblock AMASS: Archive of Motion Capture as Surface Shapes.
\newblock In \emph{Proceedings of the IEEE/CVF International Conference on
  Computer Vision (ICCV)}, October 2019.

\bibitem[Murphy et~al.(2024)Murphy, Zylberberg, and Fyshe]{murphy2024cka}
Alex Murphy, Joel Zylberberg, and Alona Fyshe.
\newblock Correcting Biased Centered Kernel Alignment Measures in Biological and Artificial Neural Networks, 2024.
\newblock URL \url{https://arxiv.org/abs/2405.01012}.

\bibitem[Nguyen et~al.(2021)Nguyen, Raghu, and Kornblith]{nguyen2021wide}
Thao Nguyen, Maithra Raghu, and Simon Kornblith.
\newblock Do Wide and Deep Networks Learn the Same Things? Uncovering How Neural Network Representations Vary with Width and Depth.
\newblock In \emph{International Conference on Learning Representations}, 2021.
\newblock URL \url{https://openreview.net/forum?id=KJNcAkY8tY4}.

\bibitem[Rafidashti et~al.(2025)Rafidashti, Lan, Fatemi, Fu, Hammarstrand, and
  Svensson]{rafidashti2025neuradar}
Mahan Rafidashti, Ji~Lan, Maryam Fatemi, Junsheng Fu, Lars Hammarstrand, and
  Lennart Svensson.
\newblock NeuRadar: Neural Radiance Fields for Automotive Radar Point Clouds.
\newblock In \emph{2025 IEEE/CVF Conference on Computer Vision and Pattern
  Recognition Workshops (CVPRW)}, pp.\  2479--2489, 2025.
\newblock \doi{10.1109/CVPRW67362.2025.00234}.

\bibitem[Rahman et~al.(2025)Rahman, Yataka, Kato, Wang, Li, Cardace, and
  Boufounos]{mmvr}
M.~Mahbubur Rahman, Ryoma Yataka, Sorachi Kato, Pu~Wang, Peizhao Li, Adriano
  Cardace, and Petros Boufounos.
\newblock MMVR: Millimeter-Wave Multi-View Radar Dataset and Benchmark for Indoor Perception.
\newblock In Ale{\v{s}} Leonardis, Elisa Ricci, Stefan Roth, Olga Russakovsky,
  Torsten Sattler, and G{\"u}l Varol (eds.), \emph{Computer Vision -- ECCV
  2024}, pp.\  306--322, Cham, 2025. Springer Nature Switzerland.
\newblock ISBN 978-3-031-72986-7.

\bibitem[Rebut et~al.(2022)Rebut, Ouaknine, Malik, and Pérez]{radial}
Julien Rebut, Arthur Ouaknine, Waqas Malik, and Patrick Pérez.
\newblock Raw High-Definition Radar for Multi-Task Learning.
\newblock In \emph{2022 IEEE/CVF Conference on Computer Vision and Pattern
  Recognition (CVPR)}, pp.\  17000--17009, 2022.
\newblock \doi{10.1109/CVPR52688.2022.01651}.

\bibitem[Roldan et~al.(2024)Roldan, Palffy, Kooij, Gavrila, Fioranelli, and
  Yarovoy]{radelft}
Ignacio Roldan, Andras Palffy, Julian F.~P. Kooij, Dariu~M. Gavrila, Francesco
  Fioranelli, and Alexander Yarovoy.
\newblock A Deep Automotive Radar Detector Using the RaDelft Dataset.
\newblock \emph{IEEE Transactions on Radar Systems}, 2:\penalty0 1062--1075,
  2024.
\newblock \doi{10.1109/TRS.2024.3485578}.

\bibitem[Sen et~al.(2024)Sen, Das, Pradhan, and Chakraborty]{mmdoppler}
Argha Sen, Anirban Das, Swadhin Pradhan, and Sandip Chakraborty.
\newblock A Dataset for Multi-Intensity Continuous Human Activity Recognition through Passive Sensing, 2024.
\newblock URL \url{https://arxiv.org/abs/2407.21125}.

\bibitem[Shen et~al.(2024)Shen, Pi, Xia, Cen, Peng, Hu, Bao, Hu, and
  Zhou]{gvhmr}
Zehong Shen, Huaijin Pi, Yan Xia, Zhi Cen, Sida Peng, Zechen Hu, Hujun Bao,
  Ruizhen Hu, and Xiaowei Zhou.
\newblock World-Grounded Human Motion Recovery via Gravity-View Coordinates.
\newblock In \emph{SIGGRAPH Asia 2024 Conference Papers}, SA '24, New York, NY,
  USA, 2024. Association for Computing Machinery.
\newblock ISBN 9798400711312.
\newblock \doi{10.1145/3680528.3687565}.
\newblock URL \url{https://doi.org/10.1145/3680528.3687565}.

\bibitem[Singh et~al.(2019)Singh, Sandha, Garcia, and
  Srivastava]{singh2019radhar}
Akash~Deep Singh, Sandeep~Singh Sandha, Luis Garcia, and Mani Srivastava.
\newblock RadHAR: Human Activity Recognition from Point Clouds Generated through a Millimeter-Wave Radar.
\newblock In \emph{Proceedings of the 3rd ACM Workshop on Millimeter-Wave
  Networks and Sensing Systems}, mmNets '19, pp.\  51–56, New York, NY, USA,
  2019. Association for Computing Machinery.
\newblock ISBN 9781450369329.
\newblock \doi{10.1145/3349624.3356768}.
\newblock URL \url{https://doi.org/10.1145/3349624.3356768}.

\bibitem[Skolnik et~al.(1980)]{skolnik2001introduction}
Merrill~Ivan Skolnik et~al.
\newblock \emph{Introduction to Radar Systems}, volume~3.
\newblock McGraw-hill New York, 1980.

\bibitem[Song et~al.(2025)Song, Ye, Guo, He, and Yang]{radarrgbd}
Tieshuai Song, Jiandong Ye, Ao~Guo, Guidong He, and Bin Yang.
\newblock RadarRGBD A Multi-Sensor Fusion Dataset for Perception with RGB-D and mmWave Radar, 2025.
\newblock URL \url{https://arxiv.org/abs/2505.15860}.

\bibitem[Sundar~Ram \& Ling(2008)Sundar~Ram and Ling]{ram2008simulation}
Shobha Sundar~Ram and Hao Ling.
\newblock Simulation of Human MicroDopplers Using Computer Animation Data.
\newblock In \emph{2008 IEEE Radar Conference}, pp.\  1--6, 2008.
\newblock \doi{10.1109/RADAR.2008.4720816}.

\bibitem[Tevet et~al.(2023)Tevet, Raab, Gordon, Shafir, Cohen-Or, and
  Bermano]{tevet2023mdm}
Guy Tevet, Sigal Raab, Brian Gordon, Yoni Shafir, Daniel Cohen-Or, and
  Amit~Haim Bermano.
\newblock Human Motion Diffusion Model.
\newblock In \emph{The Eleventh International Conference on Learning
  Representations}, 2023.
\newblock URL \url{https://openreview.net/forum?id=SJ1kSyO2jwu}.

\bibitem[Vilesov et~al.(2022)Vilesov, Chari, Armouti, Harish, Kulkarni,
  Deoghare, Jalilian, and Kadambi]{equipleth}
Alexander Vilesov, Pradyumna Chari, Adnan Armouti, Anirudh~Bindiganavale
  Harish, Kimaya Kulkarni, Ananya Deoghare, Laleh Jalilian, and Achuta Kadambi.
\newblock Blending Camera and 77 GHz Radar Sensing for Equitable, Robust Plethysmography.
\newblock \emph{ACM Trans. Graph.}, 41\penalty0 (4), July 2022.
\newblock ISSN 0730-0301.
\newblock \doi{10.1145/3528223.3530161}.
\newblock URL \url{https://doi.org/10.1145/3528223.3530161}.

\bibitem[Vishwakarma et~al.(2022)Vishwakarma, Li, Tang, Woodbridge, Adve, and
  Chetty]{vishwakarma2022simhumalator}
Shelly Vishwakarma, Wenda Li, Chong Tang, Karl Woodbridge, Raviraj Adve, and
  Kevin Chetty.
\newblock SimHumalator: An Open-Source End-to-End Radar Simulator for Human Activity Recognition.
\newblock \emph{IEEE Aerospace and Electronic Systems Magazine}, 37\penalty0
  (3):\penalty0 6--22, 2022.
\newblock \doi{10.1109/MAES.2021.3138948}.

\bibitem[Walter et~al.(2007)Walter, Marschner, Li, and
  Torrance]{walter2007microfacet}
Bruce Walter, Stephen~R. Marschner, Hongsong Li, and Kenneth~E. Torrance.
\newblock Microfacet Models for Refraction through Rough Surfaces.
\newblock In \emph{Proceedings of the 18th Eurographics Conference on Rendering
  Techniques}, EGSR'07, pp.\  195–206, Goslar, DEU, 2007. Eurographics
  Association.
\newblock ISBN 9783905673524.

\bibitem[Wang et~al.(2024)Wang, Lv, Zhu, Ding, and Han]{xrf55}
Fei Wang, Yizhe Lv, Mengdie Zhu, Han Ding, and Jinsong Han.
\newblock XRF55: A Radio Frequency Dataset for Human Indoor Action Analysis.
\newblock \emph{Proc. ACM Interact. Mob. Wearable Ubiquitous Technol.},
  8\penalty0 (1), March 2024.
\newblock \doi{10.1145/3643543}.
\newblock URL \url{https://doi.org/10.1145/3643543}.

\bibitem[Wei et~al.(2024)Wei, Luomei, Zhang, and Xu]{wei2024csvbsdf}
Jiangtao Wei, Yixiang Luomei, Xu~Zhang, and Feng Xu.
\newblock Learning Surface Scattering Parameters from SAR Images Using Differentiable Ray Tracing.
\newblock \emph{IEEE Transactions on Geoscience and Remote Sensing},
  62:\penalty0 1--15, 2024.
\newblock \doi{10.1109/TGRS.2024.3459620}.

\bibitem[Williams et~al.(2021)Williams, Kunz, Kornblith, and
  Linderman]{williams2021shape}
Alex~H Williams, Erin Kunz, Simon Kornblith, and Scott Linderman.
\newblock Generalized Shape Metrics on Neural Representations.
\newblock In M.~Ranzato, A.~Beygelzimer, Y.~Dauphin, P.S. Liang, and J.~Wortman
  Vaughan (eds.), \emph{Advances in Neural Information Processing Systems},
  volume~34, pp.\  4738--4750. Curran Associates, Inc., 2021.
\newblock URL
  \url{https://proceedings.neurips.cc/paper_files/paper/2021/file/252a3dbaeb32e7690242ad3b556e626b-Paper.pdf}.

\bibitem[Wu et~al.(2023)Wu, Zhang, Xie, Yu, Chen, Hu, and Chen]{hiber}
Zhi Wu, Dongheng Zhang, Chunyang Xie, Cong Yu, Jinbo Chen, Yang Hu, and Yan
  Chen.
\newblock RFMask: A Simple Baseline for Human Silhouette Segmentation with Radio Signals.
\newblock \emph{IEEE Transactions on Multimedia}, 25:\penalty0 4730--4741,
  2023.
\newblock \doi{10.1109/TMM.2022.3181455}.

\bibitem[Xiong et~al.(2026)Xiong, Liu, Li, Chua, Khim, Wang, and
  Xue]{xiong2026hybridsim}
Weitao Xiong, Tianyu Liu, Peng Li, Kok~Chung Chua, Toa~Chean Khim, Pu~Wang, and
  Hongfei Xue.
\newblock HybridSim: A Physics-Learning Hybrid Digital Twin for mmWave Human Sensing, 2026.
\newblock URL \url{https://arxiv.org/abs/2607.15806}.

\bibitem[Xue et~al.(2021)Xue, Ju, Miao, Wang, Wang, Zhang, and
  Su]{xue2021mmmesh}
Hongfei Xue, Yan Ju, Chenglin Miao, Yijiang Wang, Shiyang Wang, Aidong Zhang,
  and Lu~Su.
\newblock mmMesh: Towards 3D Real-Time Dynamic Human Mesh Construction Using Millimeter-Wave.
\newblock In \emph{Proceedings of the 19th Annual International Conference on
  Mobile Systems, Applications, and Services}, MobiSys '21, pp.\  269–282,
  New York, NY, USA, 2021. Association for Computing Machinery.
\newblock ISBN 9781450384438.
\newblock \doi{10.1145/3458864.3467679}.
\newblock URL \url{https://doi.org/10.1145/3458864.3467679}.

\bibitem[Xue et~al.(2023)Xue, Cao, Miao, Ju, Hu, Zhang, and Su]{xue2023mmgpe}
Hongfei Xue, Qiming Cao, Chenglin Miao, Yan Ju, Haochen Hu, Aidong Zhang, and
  Lu~Su.
\newblock Towards Generalized mmWave-Based Human Pose Estimation through Signal Augmentation.
\newblock In \emph{Proceedings of the 29th Annual International Conference on
  Mobile Computing and Networking}, ACM MobiCom '23, New York, NY, USA, 2023.
  Association for Computing Machinery.
\newblock ISBN 9781450399906.
\newblock \doi{10.1145/3570361.3613302}.
\newblock URL \url{https://doi.org/10.1145/3570361.3613302}.

\bibitem[Yang et~al.(2023)Yang, Huang, Zhou, Chen, Xu, Yuan, Zou, Lu, and
  Xie]{mmfi}
Jianfei Yang, He~Huang, Yunjiao Zhou, Xinyan Chen, Yuecong Xu, Shenghai Yuan,
  Han Zou, Chris~Xiaoxuan Lu, and Lihua Xie.
\newblock MM-Fi: Multi-Modal Non-Intrusive 4D Human Dataset for Versatile Wireless Sensing.
\newblock In A.~Oh, T.~Naumann, A.~Globerson, K.~Saenko, M.~Hardt, and
  S.~Levine (eds.), \emph{Advances in Neural Information Processing Systems},
  volume~36, pp.\  18756--18768. Curran Associates, Inc., 2023.
\newblock \doi{10.52202/075280-0822}.
\newblock URL
  \url{https://proceedings.neurips.cc/paper_files/paper/2023/file/3baf7a39d07e9f4f1e258a412df94521-Paper-Datasets_and_Benchmarks.pdf}.

\bibitem[Yang et~al.(2024)Yang, Kang, Huang, Zhao, Xu, Feng, and
  Zhao]{yang2024depthanything}
Lihe Yang, Bingyi Kang, Zilong Huang, Zhen Zhao, Xiaogang Xu, Jiashi Feng, and
  Hengshuang Zhao.
\newblock Depth Anything V2.
\newblock In A.~Globerson, L.~Mackey, D.~Belgrave, A.~Fan, U.~Paquet,
  J.~Tomczak, and C.~Zhang (eds.), \emph{Advances in Neural Information
  Processing Systems}, volume~37, pp.\  21875--21911. Curran Associates, Inc.,
  2024.
\newblock \doi{10.52202/079017-0688}.
\newblock URL
  \url{https://proceedings.neurips.cc/paper_files/paper/2024/file/26cfdcd8fe6fd75cc53e92963a656c58-Paper-Conference.pdf}.

\bibitem[Yuan et~al.(2025)Yuan, Zhang, Hou, Lyu, Yu, Yu, Li, and Wu]{octonet}
Dongsheng Yuan, Xie Zhang, Weiying Hou, Sheng Lyu, Yuemin Yu, Luca Jiang-Tao
  Yu, Chengxiao Li, and Chenshu Wu.
\newblock OctoNet: A Large-Scale Multi-Modal Dataset for Human Activity Understanding Grounded in Motion-Captured 3D Pose Labels.
\newblock In D.~Belgrave, C.~Zhang, H.~Lin, R.~Pascanu, P.~Koniusz,
  M.~Ghassemi, and N.~Chen (eds.), \emph{Advances in Neural Information
  Processing Systems}, volume 38, Main Conference. Curran Associates, Inc.,
  2025.
\newblock \doi{10.52202/085713-0469}.
\newblock URL
  \url{https://proceedings.neurips.cc/paper_files/paper/2025/file/14950adea25005f89545688fe97fe5ea-Paper-Datasets_and_Benchmarks_Track.pdf}.

\bibitem[Zhang et~al.(2024)Zhang, Zhou, Jiang, and Zheng]{super}
Bo~Zhang, Zimeng Zhou, Boyu Jiang, and Rong Zheng.
\newblock SUPER: Seated Upper Body Pose Estimation Using mmWave Radars.
\newblock In \emph{2024 IEEE/ACM Ninth International Conference on
  Internet-of-Things Design and Implementation (IoTDI)}, pp.\  181--191, 2024.
\newblock \doi{10.1109/IoTDI61053.2024.00020}.

\bibitem[Zhang et~al.(2026)Zhang, Li, Wang, and Wen]{zhang2025rf4d}
Jiarui Zhang, Zhihao Li, Chong Wang, and Bihan Wen.
\newblock RF4D: Neural Radar Fields for Novel View Synthesis in Outdoor Dynamic Scenes.
\newblock In \emph{Proceedings of the IEEE/CVF Conference on Computer Vision
  and Pattern Recognition (CVPR)}, pp.\  15387--15397, June 2026.

\bibitem[Zhao et~al.(2018)Zhao, Li, Abu~Alsheikh, Tian, Zhao, Torralba, and
  Katabi]{zhao2018through}
Mingmin Zhao, Tianhong Li, Mohammad Abu~Alsheikh, Yonglong Tian, Hang Zhao,
  Antonio Torralba, and Dina Katabi.
\newblock Through-Wall Human Pose Estimation Using Radio Signals.
\newblock In \emph{Proceedings of the IEEE Conference on Computer Vision and
  Pattern Recognition (CVPR)}, June 2018.

\bibitem[Zhao et~al.(2023)Zhao, Lu, Wang, Trigoni, and Markham]{cubelearn}
Peijun Zhao, Chris~Xiaoxuan Lu, Bing Wang, Niki Trigoni, and Andrew Markham.
\newblock CubeLearn: End-to-End Learning for Human Motion Recognition from Raw mmWave Radar Signals.
\newblock \emph{IEEE Internet of Things Journal}, 10\penalty0 (12):\penalty0
  10236--10249, 2023.
\newblock \doi{10.1109/JIOT.2023.3237494}.

\end{thebibliography}
\bibliographystyle{iclr2027_conference}

\appendix
\etocsettocdepth.toc{subsection}
\clearpage

\etocsettocstyle{\section*{Table of Contents}\vspace{-0.5em}}{}
\begingroup
\hypersetup{linkcolor=black}
\small
\tableofcontents
\endgroup
\vspace{1em}

\clearpage
\clearpage
\section{Radar and Rendering Theory}
\label{sec:supp:theory}

\subsection{From the Radar Equation to the Rendering Equation}
\label{sec:supp:derivation}

\textbf{Radar equation.} The power received over the single-bounce path
TX$\,\to\mathbf{x}\to\,$RX is the bistatic radar
equation~\cite{skolnik2001introduction},
\begin{equation}
  P_r \;=\; \frac{P_t\,G_t(\boldsymbol{\omega}_t)\,G_r(\boldsymbol{\omega}_r)\,\lambda^2\,\sigma}
                 {(4\pi)^3\,R_t^2\,R_r^2},
  \label{eq:radar_bistatic}
\end{equation}
with $P_t$ the transmit power, $G_t,G_r$ the antenna gain patterns and
$\sigma$ the bistatic radar cross-section. As in prior radar inverse
rendering~\cite{mmir, 3dps}, we replace $\sigma$ by a spatially varying BSDF
$f(\boldsymbol{\omega}_t,\mathbf{x},\boldsymbol{\omega}_r)$ and integrate
over the receiver-centered hemisphere $\Omega_{rx}$, converting the surface
element to receiver solid angle through
$\mathrm{d}\omega_r = |\mathbf{n}\cdot\boldsymbol{\omega}_r|\,\mathrm{d}A/R_r^2$,
which gives Eq.~\eqref{eq:rendering_eqn} of the main text. Transmitter $t$ sits at
$\mathbf{p}_t$ and receiver $r$ at $\mathbf{p}_r$; for a surface point
$\mathbf{x}$ with unit normal $\mathbf{n}$, $R_t(\mathbf{x}) =
\|\mathbf{x}-\mathbf{p}_t\|$, $R_r(\mathbf{x}) = \|\mathbf{x}-\mathbf{p}_r\|$,
and $\boldsymbol{\omega}_t,\boldsymbol{\omega}_r$ are the unit directions from
$\mathbf{x}$ toward the transmitter and the receiver.

\textbf{Surface to solid angle.} A distributed surface replaces the point
target's cross-section by a BSDF $f$ through $\sigma = 4\pi\int f\,
\cos\theta_t\cos\theta_r\,\mathrm{d}A$, with $\cos\theta_r = |\mathbf{n}\cdot
\boldsymbol{\omega}_r|$, so that the power received from the surface element
$\mathrm{d}A$ at $\mathbf{x}$ is $V\,P_t\,G_t G_r\lambda^2 f\cos\theta_t
\cos\theta_r\,\mathrm{d}A / ((4\pi)^2 R_t^2 R_r^2)$. That element subtends
$\mathrm{d}\omega_r = \cos\theta_r\,\mathrm{d}A/R_r^2$ at the receiver, which
absorbs both $\cos\theta_r$ and $R_r^2$ and leaves
Eq.~\eqref{eq:rendering_eqn}, the form mmIR integrates by Monte Carlo and
3DPS discretizes deterministically.

\textbf{Riemann sum on the oriented point set.} Each point $i$ of the set would be assigned the solid angle its surface patch subtends at the receiver. We set it to 1 for every point. For the scene, the per-point materials absorb the level. For the person, the SMPL mesh's sampling density is not the view's, and a cosine weight darkens every oblique surface, the limbs' sides among
them, so the level is left to the area
scale and the learned gains. The sum of per-point powers is then
$P_r = \sum_i P_t\lambda^2 G_t G_r f_i \cos\theta_{t,i}\,\Delta\Omega_i\,V_i /
((4\pi)^2 R_{t,i}^2)$, whose amplitude form, the quantity
Eq.~\eqref{eq:doppler_splat} of the main text calls
$S_{tr}(\mathbf{x}_i)$, is
\begin{equation}
\begin{aligned}
  S_{tr}(\mathbf{x}_i) \;=\;&
  C\,\frac{\sqrt{G_t(\boldsymbol{\omega}_{t,i})}}{R_{t,i}}\,
  \sqrt{G_r(\boldsymbol{\omega}_{r,i})\,\Delta\Omega_i}\,
  \sqrt{f_i(\boldsymbol{\omega}_{t,i},\boldsymbol{\omega}_{r,i})\cos\theta_{t,i}} \\
  &\times\; e^{\,j\,2\pi\,(R_{t,i}+R_{r,i})/\lambda},
\end{aligned}
  \label{eq:point_amplitude}
\end{equation}
with the carrier phase attached and $C$ a constant that is immaterial
because the absolute level is learned. The solid angle $\Delta\Omega_i$ is unit for every point, because the scene's per-point materials and the person's area scale $\alpha$ and learned joint gains carry the level instead.
The renderer evaluates exactly this product: the per-(point, TX) factor
$\sqrt{G_t}/R_t$, the per-(point, RX) factor $\sqrt{G_r\,\Delta\Omega_i}$,
the per-path factor $\sqrt{f_i\cos\theta_{t,i}}$ and the constant $C =
\sqrt{P_t\,L\,\lambda^2}/(4\pi)$ times fixed receiver and ADC scale factors, with
$L$ the fixed antenna loss ($P_t = 20$\,mW and $L = 0.40$, both immaterial under the learned gain $g$). Back-facing
paths ($\cos\theta_t$ or $\cos\theta_r$ at the numerical floor) and paths
switched off by the visibility masks are given exactly zero weight; the
renderer is double-sided, so a normal's sign flips rather than culls.

\textbf{The splat.} The Hann window's exact response, the $\Phi$ of
Eq.~\eqref{eq:doppler_splat}, is
\begin{equation}
  \Phi_{K}(\delta) \;=\; \tfrac12\,D_{K}(\delta) - \tfrac14\,D_{K}(\delta-1) - \tfrac14\,D_{K}(\delta+1),
  \label{eq:hann_psf}
\end{equation}
with $D_{K}(\delta) = \sum_{n=0}^{K-1} e^{-j2\pi\delta n/K}$ the $K$-point
Dirichlet kernel. Pushing the range FFT inside the point sum~\cite{3dps}
makes the complex range profile of one physical pair the coherent sum over
the visible points
\begin{equation}
  S_{tr}[k] \;=\; g \sum_{i=1}^{N} b_i V_i\,S_{tr}(\mathbf{x}_i)\,
  \Phi_{K}(k - k_{tr,i}) + C_{tr}[k],
  \label{eq:range_splat}
\end{equation}
where $C_{tr}[k]$ is the antenna-coupling profile estimated from the
recording, non-zero on the first 15 bins only, and $b_i$ is 1 for scene points and $\sqrt{\alpha\gamma_i}$ for person points,
with $\alpha$ the body's area scale and $\gamma_i$ a per-point scattering
gain skinned from 14 learned joint gains. The rendered profile has the shape
of the recording's demultiplexed ADC after the range FFT,
$(N_{\text{TX}},N_{\text{RX}},K)$ with $K = 256$, and no channel or bin is
dropped. The range FFT of a Hann-windowed chirp is, per point, the
window's response at the point's fractional bin, Eq.~\eqref{eq:hann_psf} with
$K = 256$, so the ADC-then-FFT pipeline and the direct scatter-add of
Eq.~\eqref{eq:range_splat} are the same computation up to the truncation to
$L=15$ taps. The bin index $k_{tr,i} = (R_{t,i}+R_{r,i})/(2\Delta R)$ with
$\Delta R = c f_s/(2 S K)$ ($4.29$\,cm on HuPR, $4.51$\,cm on the cascade) is
rounded only to choose the 15 receiving bins; the carrier phase and the PSF's
fractional argument use the unrounded path length, and the PSF is read from a
1024-entry table over the fractional bin by linear interpolation. Check: one
point target through the real and the rendered chain agrees to
$1.5\times10^{-7}$ (float32 rounding) on range profiles, loop range--azimuth,
the Doppler cube and the Doppler planes; the 15-tap truncation is the one
approximation left, $7.3\times10^{-4}$ of the peak, the same on both datasets.

\textbf{Products.} Every product is a Fourier transform of the rendered
return, taken with the shared windows of App.~\ref{sec:supp:preproc} on the
real and the rendered side alike. The range--Doppler map of one antenna pair
is the windowed $N_l$-point DFT over loops after removing the slow-time mean,
\begin{equation}
  \mathrm{RD}_{tr}[d,k] \;=\; \sum_{l=0}^{N_l-1} w_D[l]\,
  \bigl(S[l,t,r,k] - \bar{S}_{tr}[k]\bigr)\,e^{-j2\pi d\,l/N_l},
  \label{eq:range_doppler}
\end{equation}
with $\bar{S}_{tr}[k]$ the mean over $l$ and $w_D$ the Doppler window,
shifted so that $d = N_l/2$ is DC; Eq.~\eqref{eq:doppler_splat} evaluates
this directly, without forming $S[l,t,r,k]$. Range--azimuth is the angle FFT
over the azimuth virtual array formed from the same cube,
$\mathrm{RA}[l,k,a] = \mathcal{F}_{v}\{w_A[v]\,
S[l,t(v),r(v),k]\}$; range--azimuth--elevation adds an elevation FFT over the
array's elevation rows, two on HuPR and seven on the cascade, so every
transmitter contributes; and the range--azimuth--Doppler cube
$\mathrm{RAD}[d,k,a]$ is Eq.~\eqref{eq:range_doppler} applied along $l$ of
$\mathrm{RA}$. The micro-Doppler spectrogram column of a frame is a person-range-gated
power sum over channels,
\begin{equation}
  \mathrm{MD}[d] \;=\; \sum_{t,r}\sum_{|m|\le 10} w[m]\,\bigl|\mathrm{RD}_{tr}[d,k_p+m]\bigr|^2,
  \label{eq:microdoppler}
\end{equation}
with $w$ a 21-bin Hann window and $k_p$ the person's range bin (the 25th
percentile of its point cloud's range); correlations are taken on
$\sqrt{\mathrm{MD}}$. The trainer's angle FFTs have 16 azimuth bins on HuPR's 8-slot row and 86 on
the cascade (2 and 8 elevation bins); the products and metrics use 64 and 128.
The angle axes are untapered inside the trainer and the downstream
models, because the downstream models beamform untapered, and a wall's untapered
response is far stronger than its tapered view shows.

\subsection{The BSDF as Evaluated}
\label{sec:supp:bsdf}

The per-path BSDF is the closed-form BSDF of \citet{3dps}, unchanged, in the
coherent spatially varying BSDF (CSVBSDF) formulation of
\citet{wei2024csvbsdf}. With
$\boldsymbol{\theta}_i = (\varepsilon'_r,\varepsilon''_r,\sigma_h,\ell_c,\tau,d)$
mapped to their physical ranges by sigmoid ($\tau$) and shifted softplus (the
rest), the value the kernel returns for the path $(i,t,r)$ is
\begin{equation}
  f_i\cos\theta_{t,i} \;=\;
  \Bigl[\,\tau_{\text{eff}}\,|\Gamma_{\mathbf{h}}|^2\,K_{\text{KA}}
  + (1-\tau_{\text{eff}})\,|\Gamma_{\mathbf{n}}|^2\,K_{\text{SPM}}\Bigr]\cos\theta_{t,i},
  \label{eq:bsdf_evaluated}
\end{equation}
where $K_{\text{KA}} = D\,G/(4\cos\theta_t\cos\theta_r)$ is a GGX microfacet
lobe~\cite{walter2007microfacet} with Smith masking and roughness
$\alpha_{\text{GGX}} = \sqrt{4\pi\sigma_h/\lambda}$ (clamped to $[0.05, 0.95]$);
$K_{\text{SPM}} = \frac{\kappa}{4\pi\sinh\kappa}\,e^{\kappa(\cos\Delta - 1)}\,\eta_\varepsilon$
is an exponential lobe about the mirror direction, with the von Mises--Fisher prefactor and concentration
$\kappa = \sqrt{\ell_c/\lambda}/(1+3\sigma_h/\ell_c)$ clamped to $[0.5, 10]$ and a
permittivity-contrast factor $\eta_\varepsilon = \mathrm{clamp}((|\varepsilon'_r-1| +
\varepsilon''_r)/5,\,0.2,\,1)$; $\Gamma_{\mathbf{h}}$ and $\Gamma_{\mathbf{n}}$
are the multilayer-slab Fresnel coefficients~\cite{itu_p2040} of thickness $d$
and permittivity $\varepsilon_r = \varepsilon'_r - j\varepsilon''_r$, evaluated at
the half-vector and at the macro normal respectively, each projected from the
transmit polarization through the local $s/p$ Jones basis onto the receive
polarization and squared; and $\tau_{\text{eff}}$ is the learned blend $\tau$
gated by the ITU validity sigmoids in $2\pi\ell_c/\lambda$, $2\pi\sigma_h/\lambda$
and the incidence angle. The scene's normals are the third column of a
per-point unit quaternion; the person's are a free unit vector per visible
point per frame.

\subsection{Closed-Form Doppler-Plane Amplitudes}
\label{sec:supp:doppler_closed_form}

\textbf{The slow-time cube the trainer never forms.} Under TDM, transmitter
$t$ fires in slot $s_t$ of every loop, so loop $l$ of transmitter $t$ is
acquired at
\begin{equation}
  t_{l,t} \;=\; \bigl(l\,N_{\text{slots}} + s_t\bigr)\,T_c,
  \label{eq:tdm_times}
\end{equation}
with $T_c$ the chirp interval and $N_{\text{slots}}$ the slots per loop
(HuPR: $T_c = 72\,\mu$s, 3 slots, a 216\,$\mu$s loop; the cascade:
65\,$\mu$s, 12 slots, 780\,$\mu$s),
and a person point moving with velocity $\mathbf{v}_i$ changes its bistatic
path length to $(t,r)$ at the rate
\begin{equation}
  \dot{R}_{tr,i} \;=\; \frac{\mathrm{d}}{\mathrm{d}t}\bigl(R_{t,i}+R_{r,i}\bigr)
  \;=\; \mathbf{v}_i\cdot\mathbf{u}_{t,i} + \mathbf{v}_i\cdot\mathbf{u}_{r,i},
  \label{eq:path_rate}
\end{equation}
with $\mathbf{u}_{t,i} = (\mathbf{x}_i-\mathbf{p}_t)/R_{t,i}$ and
$\mathbf{u}_{r,i} = (\mathbf{x}_i-\mathbf{p}_r)/R_{r,i}$, positive when the
point recedes, so its contribution to loop $l$ carries the phase ramp
$e^{\,j2\pi\dot{R}_{tr,i}t_{l,t}/\lambda}$. The rendered product of a frame
is therefore the complex slow-time cube
\begin{equation}
\begin{aligned}
  S[l,t,r,k] \;=\;& g\!\!\sum_{i\in\text{scene}}\!\! b_i V_i S_{tr}(\mathbf{x}_i)\,\Phi_{K}(k-k_{tr,i})
  \;+\; C_{tr}[k] \\
  &+\; g\!\!\sum_{i\in\text{person}}\!\! b_i V_i S_{tr}(\mathbf{x}_i)\,\Phi_{K}(k-k_{tr,i})\,
  e^{\,j2\pi\dot{R}_{tr,i}\,t_{l,t}/\lambda},
\end{aligned}
  \label{eq:slow_time_cube}
\end{equation}
of shape $(N_l, N_{\text{TX}}, N_{\text{RX}}, K)$, exactly the shape of the
recording's demultiplexed, range-transformed ADC, for every physical TX--RX
pair. Because Eq.~\eqref{eq:tdm_times} places each transmitter at its own
intra-loop offset, a moving point accumulates the inter-TX phase progression
that a real TDM capture must compensate before angle estimation, and Doppler
is coupled to whichever lateral axes the array resolves. Range migration and
the change of the BSDF within the frame's slow-time span (13.8\,ms on HuPR,
49.9\,ms on the cascade) are neglected; the carrier phase uses every physical
transmitter time.

The trainer never forms the cube of Eq.~\eqref{eq:slow_time_cube}.
For a person path the slow-time samples are one tone: with
Eq.~\eqref{eq:tdm_times}, $t_{l,t} = t_{0,t} + l\,T_{\text{loop}}$, so
\begin{equation}
  e^{\,j2\pi\dot{R}_{tr,i}t_{l,t}/\lambda} \;=\; e^{\,j\phi_{0}}\,e^{\,j\omega l},
  \quad \phi_0 = \tfrac{2\pi}{\lambda}\dot{R}_{tr,i}\,t_{0,t},
  \quad \omega = \tfrac{2\pi}{\lambda}\dot{R}_{tr,i}\,T_{\text{loop}},
\end{equation}
and, with $\nu = N_l\,\omega/2\pi$ the path's fractional Doppler bin,
the windowed, mean-removed DFT of Eq.~\eqref{eq:range_doppler} over
$l$ evaluates at unshifted bin $d'$ to
\begin{equation}
\begin{aligned}
  A_{d'}(\nu) \;=\; e^{\,j\phi_0}\Bigl[&\tfrac12 D_{N_l}(d'-\nu) - \tfrac14 D_{N_l}(d'-1-\nu) \\
  &- \tfrac14 D_{N_l}(d'+1-\nu) - m_{d'}\,D_{N_l}(-\nu)\Bigr],
\end{aligned}
  \label{eq:doppler_closed_form}
\end{equation}
the Hann Doppler PSF (Eq.~\eqref{eq:hann_psf} with $K = N_l$) plus the
mean removal folded into the row, $m_{d'} = \tfrac12$ at $d' = 0$,
$-\tfrac14$ at $d' = \pm1$ and $0$ elsewhere, which is what makes the
rendered planes agree with a real chain that removes the mean \emph{before}
it windows. The loop-0 anchor plane the range--azimuth term reads is
$A_{\text{anchor}} = e^{\,j\phi_0}$. The rate gradient $\partial A_{d'}/\partial\dot{R}$ is
the same expression with the Dirichlet kernel's derivative, and both are
evaluated in float32 with $\nu$ and $\phi_0$ formed in float64 and reduced to
an integer bin and a fraction first. Check, on both
datasets, windowed and not: against a float64 reference that is itself
$4\times10^{-14}$ from the 64-loop sum and whose gradient is $8\times10^{-9}$
from finite differences, the closed-form kernels are at $7.3\times10^{-7}$
(amplitudes) and $5.4\times10^{-7}$ (rate gradient).

\textbf{Truncation.} With the window the Doppler response is compact (the
first sidelobe is $2.8\times10^{-2}$ of the peak in amplitude at 2.5 bins and
the envelope falls by a factor of eight per octave, in bin units whatever $N_l$ is), so each
path deposits into the $L_d$ planes around its own fractional bin, as the
range PSF is cut at 15. Measured on the point-target check, the truncation of
the planes is $7.3\times10^{-4}$ at 15 taps, $3.5\times10^{-3}$ at 9,
$7.7\times10^{-3}$ at 7 and $2.3\times10^{-2}$ at 5; $L_d = 7$ is the default,
and at 400 epochs on sequence 183 the 7-tap and exact arms agree
(spectrogram $+0.941$ / $+0.933$, energy $+0.884$ / $+0.861$, range--Doppler
occupied $+0.399$ / $+0.394$, the same learned floor). A path whose 7 planes
fall outside a view's rendered support deposits nothing there.

\clearpage
\section{Implementation Details}
\label{sec:supp:impl}

\subsection{Preprocessing}
\label{sec:supp:preproc}

\textbf{HuPR scaffold.} Frames 0, 150, 300 and 450 of the sequence are read
from the released RGB, passed in one batch through Depth-Anything-V2
(Large, metric)~\cite{yang2024depthanything}, the annotated person's bounding
box (padded by 12\%) is cut out of each depth map, and the maps are
back-projected under a $52^\circ$ horizontal field of view, estimated from the radar's range and the released 2D keypoints, at a
pixel stride of 2, voxel-downsampled at 4\,cm per view and 5\,cm after
merging, and given PCA normals (0.18\,m radius, 50 neighbors) oriented
toward the radar. Two person-free real maps are formed per board: the initial
material fit trains on the pixelwise median of the four sampled frames'
loop-mean $|\mathrm{RA}|$, and the range alignment scores against the
per-cell 25th percentile of loop-mean $|\mathrm{RA}|$ over every fourth frame
of the sequence (the subject is the only mover and occupies any one cell in
a minority of frames, so a low quantile is the static scene with the person
gone). The range translation is searched on a 2\,cm grid over $\pm2.4$\,m and
refined at 2\,mm, every candidate scored at once on the GPU (0.3\,s): the
score is the min--max-normalized Pearson correlation of the shifted cloud's
(range, angle) occupancy against the person-free map, mean over both boards,
restricted to the cloud's own angle bins (the camera sees $\pm26^\circ$ of
the radar's $\pm75^\circ$) and to ranges beyond 0.5\,m. Lateral offset and
height are not fitted because the map constrains them poorly: positions
within 0.03 of the best score span a median of 1.3\,m laterally and 1.0\,m in
height, while the range optimum has no rival within 0.03 on 233 of 235
sequences: sequence 256: $-0.55$\,m ($r$ 0.23 to 0.49);
38: $-1.11$\,m (0.10 to 0.48); 102: $-0.77$\,m (0.13 to 0.57). The point-cloud
initialization of \citet{3dps} then reduces the aligned cloud to $N = 20{,}000$ points by a
field-of-view restriction, cosine-weighted resampling toward boresight and
farthest-point sampling, and fits per-point materials and normals to the
person-free static $|\mathrm{RA}|$ of both boards for 75 iterations
(2--3\,s), the trainer refining both for its 400 epochs.

\textbf{RT-Pose scaffold.} Eight annotated radar frames spread over the
sequence supply Livox LiDAR frames, each densified by the left camera's
Depth-Anything depth rescaled to that frame's LiDAR by the median LiDAR/predicted
depth ratio at the projected LiDAR pixels; the person is cut from each frame by the
registered body mesh of that frame (points within 0.25\,m of a vertex, the
floor kept), so that what the person touches, such as a chair, survives. The clouds are
taken to the radar frame by the rig extrinsic, merged at a 3\,cm voxel and
given PCA normals over a 0.3\,m radius (0.5\,m is needed for the LiDAR alone,
whose scan lines are 0.4\,m apart on the far wall). The extrinsic is the
identity: the radar sits at the LiDAR to within 0.03\,m laterally by two
independent person measurements, and a $\pm0.5$\,m translation search
against the real static map cannot identify a lateral shift (correlation
0.189 to 0.235 while walking the height to the edge of its range), so it is
not refitted; the annotation frame's origin is 0.2\,m to the left of the
LiDAR and the board is rotated $180^\circ$ about boresight, both verified from
the data. The cascade frames are decoded with TI's phase and amplitude
calibration.

\textbf{Antenna coupling.} Per radar, the coherent mean over every third
frame and over loops of the range-windowed profiles at the first 15 range
bins, per azimuth virtual channel. Nothing else lives within those bins
(0.68\,m on the cascade), so the coherent mean is the board's own static
near-field return: it explains 99.98\% of the near-bin power on HuPR
sequence 256 and 99.95\% on cascade sequence 183 (the fraction of the near-bin
energy the mean explains). It is added to the rendered profiles in training
and rendering, and the absolute-gain initialization leaves those bins out.

\textbf{Mesh recovery.} GVHMR~\cite{gvhmr} runs in static-camera mode in its
own environment, started with the scene stage so that its ${\sim}30$\,s
overlap: the frames held in memory, YOLO tracking of the largest persistent
track, ViTPose and the HMR2 features in half precision on half-resolution
crops, giving the $6{,}890$ SMPL vertices~\cite{smpl} per camera frame. RT-Pose's
true focal length (25.65\,mm full-frame equivalent) is passed; HuPR's is
GVHMR's default. The mesh is a deterministic function of the frames
(the same video twice is bitwise; a lossy re-encode moves it 3--5\,mm
median, a tenth of a range bin).

\textbf{Camera-to-radar placement.} RT-Pose: camera to LiDAR by the shipped
stereo calibration, LiDAR to radar by the extrinsic; the camera and LiDAR file
belonging to radar frame $f$ has stem $3f-4$ on 85 sequences and $2f-2$ on
17, a rig clock fitted on the annotation's pairs and exact on every annotated
frame. HuPR: the camera at the rig,
co-boresighted, camera (right, down, forward) to radar $(x, -z, y)$, with
the same $52^\circ$ intrinsics at 512\,px for the joints' image projection.

\textbf{Per-frame radar registration.} For every frame, the clutter-removed
range--azimuth cube (slow-time mean removed) is averaged over loops and
squared-summed over angle into a range profile of moving power; its peak
between bins 10 and 200, refined by a parabola over the two neighbors, is
the radar's range to the person, and the fraction of the profile's energy
within 5 bins of the peak is the detection confidence (a rig of two boards
contributes the confidence-weighted mean). The per-frame ratio $r_f$ =
radar range / GVHMR centroid range is interpolated across frames whose
confidence is below 0.35, median-filtered over $\pm7$ frames, clamped to
$[0.5, 2]$, flagged where it leaves the confident frames' interquartile
range widened three times (GVHMR's track is wrong there, and the radar's
range is followed), and every unflagged run is polished by a quadratic
Savitzky--Golay filter 51 frames wide, because the median filter's steps would otherwise add spurious radial motion
to the cloud. $r_f$ moves the body along its centroid's camera ray to $r_f$ times its
range; the body keeps GVHMR's size, since GVHMR's error is its depth, not the body's size. The range search is confined to the camera's angular sector, on both
mirrors of the beamformer's angle axis, and to a range window bounded by the
scale's own range. Azimuth: one registration per sequence maps the mesh's
azimuth onto the radar's moving return, $\sin\theta_{\text{radar}} = a +
b\sin\theta_{\text{mesh}}$, the scale $b$ fitted only where the mesh spans
azimuth (sine std $\ge 0.10$, $r \ge 0.90$) and a yaw alone otherwise; the
offsets are 3 to 5 degrees on the RT-Pose walking sequences, and the labels
inherit the session's lateral scale. The
median scale is 0.707 (IQR 0.700--0.717) on HuPR sequence 256, i.e.\ GVHMR
41\% long; RT-Pose's GVHMR runs about 10\% short. HuPR's 8-channel boards do
not resolve this frame by frame, so HuPR applies one depth scale per sequence
(the median $r_f$ over confident frames) and one rig yaw of $3.05^\circ$, and
places the body's ray from the released 2D keypoints. 

\textbf{Joints, velocities, motion frames.} The 14 joints in HuPR's order
come from the SMPL joint regressor; their camera-derived velocities are a
first-derivative Savitzky--Golay filter, 11 frames (1.1\,s at 10\,Hz),
quadratic. Motion frames for evaluation are those whose pose speed, the
90th percentile over the 14 smoothed joint speeds, is at least 0.15\,m/s
(1.5\,cm per frame), the same threshold on every sequence, pre-registered.
The person's material is initialized as one shared six-parameter vector fitted
for one epoch on all frames against the loop-0 $|\mathrm{RA}|$ with the scene
model fixed, together with the body's area scale; the trainer refines the shared vector (per-vertex materials let the fit darken the limbs), and the level calibration of App.~\ref{sec:supp:training} sets the area scale after training.

\textbf{Visibility.} For each frame the body's $13{,}776$ SMPL triangles are
binned by their (azimuth, elevation) box as seen from the radar on a
$128\times128$ grid, padded one cell either side, so a ray asks only the
triangles of its cell; the hit test is unchanged, so the result is the
brute-force one bit for bit. Person mask: field of view (a boresight cosine
above 0.1761, range in $(0.5, 9.4)$\,m), then a M\"oller--Trumbore test of the
open segment from the vertex (nudged 1\,mm toward the radar, past its own
faces) to the radar center against the body's triangles, and against the
scaffold triangles on the radar side of the body. Scene shadow mask: every
scaffold point inside the body's angular silhouette box and beyond its
nearest vertex is tested against the body's triangles; the rest are visible
without a test. Both masks are computed for the whole sequence on the device
and enter the renderer as a per-point gain of 0 or 1, never as a change of the
point set. With the shadow mask the wall behind the subject dims on all six
products as it does in the real capture.

\textbf{Windowing.} Range and Doppler take the periodic Hann
$w[n] = \tfrac12 - \tfrac12\cos(2\pi n/N)$, whose DFT is exactly three bins
and whose response is Eq.~\eqref{eq:hann_psf}; the symmetric
$\mathrm{hanning}(N)$ is not this window and would leave a $1/N$ mismatch
between the two sides. The array axis takes $\mathrm{hanning}(N+2)[1{:}-1]$,
because an aperture of 8 or 86 slots cannot afford to zero an element. The cascade's 144 azimuth virtual
channels (9 TX $\times$ 16 RX) land on 86 half-wavelength grid slots, 58 of
them holding two channels which are averaged, then windowed and zero-padded
to a 128-bin angle FFT; HuPR's 8 fill 8 slots, padded to 64. Preprocessing
runs in 10--18\,s per sequence: the ray casting, voxel downsampling, PCA
normals, person ranges and masks batched on the GPU, the ADC decoded once
into memory and shared by the stages.

\subsection{Training}
\label{sec:supp:training}

\begin{table}[htbp]
\centering
\caption{\textbf{Learned and fixed quantities of the forward model.} Materials
and normals are parameterized as in \citet{3dps}.
Learning rates: velocity coefficients $10^{-2}$, person normals $10^{-3}$,
scene materials $10^{-2}$, scene normals $5\cdot10^{-3}$, person material
$10^{-3}$, gains $10^{-2}$, noise floor $3\cdot10^{-2}$ (Adam).}
\label{tab:learned}
\small
\begin{tabular}{@{}p{0.34\textwidth}p{0.22\textwidth}p{0.37\textwidth}@{}}
\toprule
\textbf{Quantity} & \textbf{Shape} & \textbf{Note} \\
\midrule
\multicolumn{3}{@{}l}{\emph{Learned}} \\
joint velocity scales $s_j$, residual $c_{jk}$ & (14), (bands, 14, 3) & each residual component bounded by $0.75\|\mathbf{v}^{\text{kin}}_j\|$; 3.5\,Hz cap \\
person normals $\mathbf{n}_i$ & (visible points, 3) per frame & one fused parameter over the training frames \\
person material $\boldsymbol{\theta}$ & (6) & one six-vector for the body, initialized by the person stage \\
scene materials $\boldsymbol{\theta}_i$, normals & (N,6), (N,4) & normals as quaternions; $N = 20{,}000$ \\
receiver gain $g$, joint gains $\gamma_j$ & (1) per radar, (14) & log-parameterized; $\prod_j\gamma_j = 1$ \\
noise floor $\mathcal{N}(d,k)$ & log scale $+$ 8 range-DCT coeff.\ per radar & on the recording's own off-person prior \\
\midrule
\multicolumn{3}{@{}l}{\emph{Fixed}} \\
\multicolumn{3}{@{}p{0.98\textwidth}}{positions (registered per frame), skinning weights $W$, the area scale $\alpha$ (level calibration, after training), antenna patterns, array geometry, TDM schedule, windows, PSF widths ($L=15$ range, 7 Doppler), visibility masks $V_i$, coupling $C_{tr}$} \\
\bottomrule
\end{tabular}
\end{table}

\textbf{Level calibration.} The area scale $\alpha$ is not learned by the
objective. It is set once, after training, so that the rendered person has
the recording's level in the transform the downstream pose model of
Sec.~\ref{sec:exp:tstr} reads: on every fourth training frame the
render's Doppler-summed magnitude is formed in that model's own beamformer,
and $\alpha$ is the value at which the median ratio of rendered to recorded
peak inside the pelvis's window ($\pm0.5$\,m, $\pm8^\circ$) is one. This uses
no labels and no downstream model weights, it is fitted per sequence on
training frames only, and it is the operation a radar engineer calls gain
calibration.

\textbf{The velocity field.} The 14 joint velocities $\mathbf{v}_j(f)$ of
frame $f$ are the video-derived kinematic velocities
$\mathbf{v}^{\text{kin}}_j(f)$ (App.~\ref{sec:supp:preproc}) times a learned
per-joint scale, plus a bounded residual,
\begin{equation}
  \mathbf{v}_j(f) \;=\; s_j\,\mathbf{v}^{\text{kin}}_j(f)
  + \beta_j(f)\tanh\!\bigl(\mathbf{r}_j(f)/\beta_j(f)\bigr),
  \qquad \beta_j(f) = 0.75\,\|\mathbf{v}^{\text{kin}}_j(f)\|,
  \label{eq:velocity_field}
\end{equation}
with $\tanh$ taken per component and the residual zero where $\beta_j(f)=0$. The residual $\mathbf{r}_j(f) = \sum_k c_{jk} B_k(f)$ lives in an orthonormal
temporal DCT basis truncated at 420 of the sequence's 600 bands (209 of 299
on RT-Pose), a 3.5\,Hz cap at the 10\,Hz frame rate. Each point's velocity
$\mathbf{v}_i = \sum_j W_{ij}\mathbf{v}_j$ is skinned from the joints by soft
weights $W$ (normalized Gaussian weights, $\sigma = 0.25$\,m, over the three
nearest joints).

\textbf{The normalized $L_2$ form.} Every term of
Eq.~\eqref{eq:objective} has the same shape, the normalized $L_2$ that prior
radar inverse renderers apply to $|\mathrm{RA}|$~\cite{mmir, 3dps},
\begin{equation}
  \mathcal{L}_{\bullet} \;=\;
  \frac{\operatorname{mean}_{c}\bigl(\,\hat{y}_{\bullet}[c] - y_{\bullet}[c]\,\bigr)^2}
       {\operatorname{mean}_{c}\, y_{\bullet}[c]^2},
  \label{eq:normalised_l2}
\end{equation}
the divisor a detached constant per frame and radar. It never contains the
prediction, so the terms stay plain $L_2$ and the absolute level remains
supervised, and all terms exclude the near-field range bins below 0.5\,m.
The targets are $y_{\mathrm{RAE},0} = |\mathrm{RAE}[0,k,a,e]|$, the first
loop with the static clutter kept, and $y_{\mathrm{RAE},\mathrm{pl}} =
|\mathrm{RAED}[d,k,a,e]|$ for the 63 planes $d \neq N_l/2$, each with its own
divisor and with the learned floor entering the rendered side under the
square root, $\hat{y} = \sqrt{|\widehat{\mathrm{RAED}}|^2 + \mathcal{N}(d,k)}$,
so the plane terms are what fit the floor. Both are formed from every
transmitter. The windowed pair are the same two terms restricted to the
registered person's box, the pelvis $\pm0.6$\,m in range and $\pm1.2$\,m across
the line of sight at every elevation bin, and normalized over that box, because a frame-wide divisor lets a large relative error on the person hide
behind the scene's energy.  A frame renders only the Doppler
bins whose real power at the person's range rows exceeds the frame's median
($\kappa = 1$) plus the motion band ($-1.13$ to $+0.99$\,m/s), and every rendered plane
is supervised, including the empty bins whose zero targets keep the rendered
person compact in Doppler.

Both datasets use the same settings: 400 epochs on all 600 (HuPR) or 299 (RT-Pose) frames, a constant
learning rate for the first half and a cosine decay to $10^{-2}$ of it over
the second, the plane targets in float16, the Doppler support at $\kappa = 1$,
every term at unit weight, Adam with the rates of Table~\ref{tab:learned}, the trainer's angle axes untapered, the level calibration after training in the downstream model's view, the velocity residual bounded at 0.75 of the joint's kinematic speed, and the objective of Eq.~\eqref{eq:objective}.
What differs is set by the recording, not tuned: HuPR renders 24 frames a
step (two boards of $3\times4$ channels, 600 training frames, 420 of
600 DCT bands), the cascade 4 frames a step ($12\times16 = 192$ channels, the
count set for a 24\,GB card, 299 frames,
209 of 299 bands, the same 3.5\,Hz cap); the motion band is the same
$-1.13$ to $+0.99$\,m/s on both, 13 bins on HuPR's 216\,$\mu$s loop (0.1408\,m/s per
bin, $\pm4.5$\,m/s unambiguous) and 53 on the cascade's 780\,$\mu$s loop
(0.0390\,m/s per bin, $\pm1.25$\,m/s). The plane targets are 63 $\times$ 244
$\times$ $N_a$ per frame and radar (3.9\,MB in float32 on HuPR), so they are stored in float16. Every fourth frame is scored every 10 epochs and the
best-scoring snapshot is reported. Peak allocated device memory is
\rRtPeakGiBRT{}\,GiB on the cascade and \rRtPeakGiBHu{}\,GiB on HuPR (median
over sequences), and the fit's time per sequence is in
App.~\ref{sec:supp:runtime}.

\clearpage
\section{Datasets and Baseline}
\label{sec:supp:datasets_baseline}

\subsection{Dataset survey}
\label{sec:supp:survey}

Table~\ref{tab:dataset_survey} records the mmWave human-sensing datasets we
considered and why each was included or excluded, against six eligibility
criteria that its caption defines.

Two failure modes dominate. The first is the on-chip CFAR and beamforming
pipeline discarding slow time before release, which removes every dataset in
the table's second section. The second is withholding camera imagery for
privacy, a practice that is tightening rather than relaxing. Only RT-Pose~\cite{rtpose} and HuPR~\cite{lee2023hupr}
satisfy all six criteria.

One terminology caveat carries into the rest of the paper. ``Raw'' is used
inconsistently in this literature, and HIBER~\cite{hiber}, MMVR~\cite{mmvr},
and M4Human~\cite{m4human} all describe beamformed tensors as raw radar data.
Only ADC or I/Q samples preserve the per-chirp phase history that micro-Doppler
is computed from.

\begin{table*}[!h]
\centering
\caption{\textbf{Dataset survey.} \textbf{ST}: slow time retained, at least
two chirps per frame. \textbf{ADC}: raw ADC or I/Q samples, not heatmaps or
point clouds. \textbf{Cam}: colocated camera imagery released, not merely used
to generate labels. \textbf{Pose}: human pose labels (keypoints or a body
mesh). \textbf{Pub}: openly downloadable. \textbf{Code}: reference downstream
model code released. \yes~satisfied, \no~not satisfied, \prt~partially
satisfied (access by request or agreement, a withheld or blurred modality, an
expired mirror, chirps per frame undocumented, a released cube already
transformed along range, angle and Doppler, or model code without training
scripts or only from later work). Only RT-Pose and HuPR
satisfy all six. Pretrained weights are not a criterion: RT-Pose publishes
none, so we train its reference model ourselves (App.~\ref{sec:supp:downstream}).}
\label{tab:dataset_survey}
\footnotesize
\setlength{\tabcolsep}{6pt}
\begin{tabular}{@{}lcccccc@{}}
\toprule
\textbf{Dataset} & \textbf{ST} & \textbf{ADC} & \textbf{Cam} & \textbf{Pose} & \textbf{Pub} & \textbf{Code} \\
\midrule
\multicolumn{7}{@{}l}{\emph{Selected}} \\
\textbf{RT-Pose}~\cite{rtpose}      & \yes & \yes & \yes & \yes & \yes & \yes \\
\textbf{HuPR}~\cite{lee2023hupr}    & \yes & \yes & \yes & \yes & \yes & \yes \\
\midrule
\multicolumn{7}{@{}l}{\emph{No slow time (human sensing)}} \\
MMVR~\cite{mmvr}                    & \no  & \no  & \no  & \yes & \yes & \prt \\
HIBER~\cite{hiber}                  & \no  & \no  & \yes & \yes & \prt & \prt \\
mmBody~\cite{mmbody}                & \no  & \no  & \yes & \yes & \prt & \yes \\
M4Human~\cite{m4human}              & \no  & \no  & \yes & \yes & \no  & \yes \\
mm-Fi~\cite{mmfi}                   & \no  & \no  & \prt & \yes & \yes & \yes \\
mRI~\cite{mri}                      & \no  & \no  & \prt & \yes & \yes & \yes \\
OctoNet~\cite{octonet}              & \no  & \no  & \yes & \yes & \yes & \yes \\
MiliPoint~\cite{milipoint}          & \no  & \no  & \no  & \yes & \yes & \yes \\
MARS~\cite{mars}                    & \no  & \no  & \no  & \yes & \yes & \yes \\
RadHAR~\cite{singh2019radhar}       & \no  & \no  & \no  & \no  & \yes & \yes \\
mmDoppler~\cite{mmdoppler}          & \no  & \no  & \no  & \no  & \yes & \yes \\
\midrule
\multicolumn{7}{@{}l}{\emph{No camera released}} \\
CubeLearn~\cite{cubelearn}          & \yes & \yes & \no  & \no  & \yes & \yes \\
ColoRadar~\cite{kramer2022coloradar}& \yes & \yes & \no  & \no  & \yes & \no  \\
\midrule
\multicolumn{7}{@{}l}{\emph{Not (fully) public}} \\
DGHMesh~\cite{dghmesh}              & \yes & \yes & \yes & \yes & \no  & \no  \\
XRF55~\cite{xrf55}                  & \prt & \prt & \yes & \no  & \prt & \yes \\
mmMesh~\cite{xue2021mmmesh}         & \yes & \yes & \yes & \yes & \no  & \yes \\
SUPER~\cite{super}                  & \yes & \yes & \no  & \yes & \no  & \no  \\
EquiPleth~\cite{equipleth}          & \prt & \yes & \yes & \no  & \prt & \yes \\
\midrule
\multicolumn{7}{@{}l}{\emph{Not human sensing}} \\
RaDICaL~\cite{radical}              & \yes & \yes & \yes & \no  & \yes & \no  \\
IQ-1M~\cite{iq1m}                   & \yes & \yes & \yes & \no  & \yes & \yes \\
RAMP-CNN~\cite{rampcnn}             & \yes & \yes & \yes & \no  & \yes & \yes \\
RadarRGBD~\cite{radarrgbd}          & \yes & \prt & \yes & \no  & \yes & \no  \\
RADIal~\cite{radial}                & \yes & \yes & \yes & \no  & \yes & \yes \\
RaDelft~\cite{radelft}              & \yes & \yes & \yes & \no  & \prt & \yes \\
\bottomrule
\end{tabular}
\end{table*}

\subsection{Baseline survey}
\label{sec:supp:baselines}

A row of Table~\ref{tab:fidelity} needs a raw complex signal for our array
layout and chirp, driven by the recorded motion rather than by a class label or
text, from code that runs. Of the published radar simulators, only
RF-Genesis~\cite{chen2023rfgenesis} meets all of these requirements.
mmGPE~\cite{xue2023mmgpe} and HybridSim~\cite{xiong2026hybridsim} release no
code, and SimHumalator~\cite{vishwakarma2022simhumalator} renders one receive
channel from a fixed library of eleven motion-capture activity classes, so it
cannot take our motions or our arrays.

\subsection{RF-Genesis as a baseline}
\label{sec:supp:rfgenesis}

\textbf{What was run.} We ran RF-Genesis's released code (the public repository
as of 3 August 2025) on all \rRfgNseq{} sequences that 4DPS renders, on both
datasets, and no sequence failed. Each sequence is driven by the registered
SMPL mesh that drives the 4DPS render, in the same radar frame, and its output
is scored by the same code on the same pose-speed frames
(App.~\ref{sec:supp:diagnostics}), so the two rows of Table~\ref{tab:fidelity}
differ only in the renderer.

\textbf{Changes to the released code.} Each change is needed for the output to
describe our radar and the recorded motion, and none adds a fitted parameter:
(i)~the camera and the spot light that stands for the transmitter sit at the
radar board, along its boresight; (ii)~the registered mesh's vertices replace
the text-generated motion; (iii)~the ray intersection is evaluated before the
render call, which changes no value but lets the pipeline link on our driver;
(iv)~each chirp is timed at its slot in the recording's TDM schedule and fires
its own transmitter, where upstream spread the chirps over the whole frame and
so scaled every Doppler frequency by $3.6\times$; and (v)~the interpolator
reads the depth channel where it read intensity as depth.

\textbf{Adapter choices.} RF-Genesis interpolates motion at 30\,fps, so we
resample the registered mesh from the radar's 10\,Hz frames by linear
interpolation of its vertices. We give it no environment, because its only
environment model is a text-prompted image that is not registered with the
body.

\textbf{Downstream training.} The RF-Genesis arms of Table~\ref{tab:tstr} use
the models, protocol, splits and labels of the 4DPS arms. On RT-Pose one
difference is needed: RF-Genesis's units are arbitrary and HRRadarPose
normalizes its input with fixed constants, so without a gain its inputs reach
the network as zeros. We therefore scale them by one gain, $G=$ \rRfgGain{},
fixed before any training as the median, over \rRfgGainFrames{} frames of each
of the \rRfgGainSeqs{} training sequences, of the recording's moving-return
level (the 99.9th percentile of the mean-removed cube) over RF-Genesis's
(interquartile range \rRfgGainLo{} to \rRfgGainHi{}). HuPR needs no gain,
because its preprocessing normalizes each map from the data.

\textbf{Limitations of the comparison.} First, RF-Genesis renders no static
scene, so its range--angle products hold the person alone while the recording
holds the room. The range--azimuth $\rho$ therefore also measures the missing
room, and the clutter-removed columns (range--azimuth $\rho_c$, range--Doppler
$\rho_M$ and micro-Doppler) are the fair comparison, because they remove the
static scene on both sides. Second, the released code is single-bounce diffuse
ray casting, not the full system the RF-Genesis paper describes, and we run it
as released apart from the changes above. Third, RF-Genesis has no absolute
scale and no receiver noise, so for the columns that compare levels we multiply
its render by one constant per radar and sequence,
$g = (\sum|\mathrm{RA}_{\mathrm{real}}|^2 /
\sum|\mathrm{RA}_{\mathrm{RF\text{-}Genesis}}|^2)^{1/2}$ over the loop-0
range--azimuth magnitude on the retained range bins of every frame, the rule
that initializes 4DPS's own gain. The constant leaves every correlation and
shape column unchanged, and because no noise is simulated, RF-Genesis's $L_U$
stays far below 1. Fourth, it runs open loop: nothing is fitted to the capture,
so its row measures a forward simulator driven by the recorded motion, not a
competing fitted renderer (App.~\ref{sec:supp:limitations}).

\clearpage
\section{Diagnostic Fidelity Metrics}
\label{sec:supp:diagnostics}

The main-paper tables carry the metrics that separate methods. These are the
ones that diagnose \emph{how} a render fails, and they are reported for 4DPS,
its unfit render and, where a column is defined for it, RF-Genesis.

\textbf{Protocol details behind Sec.~\ref{sec:exp:fidelity}.} The
micro-Doppler trajectory statistics (centroid, spread, envelopes) are taken
on frames whose pose speed is at least 0.15\,m/s, because a frame with no
motion has no trajectory to score and its centroid is set by the residual
clutter rather than by the subject.

\textbf{The raw chain: ADC, complex range profile and range--azimuth--elevation.}
The same fitted scene yields the raw ADC samples, the complex range profile
and the range--azimuth--elevation magnitude with no retraining, because each
is a fixed transform of the per-path quantity Eq.~\eqref{eq:doppler_splat}
already deposits. Tables~\ref{tab:supp_fidelity_adc} and~\ref{tab:supp_fidelity_crp} score the
ADC and the CRP, and Table~\ref{tab:supp_fidelity_ra_rae} scores the RAE next to
range--angle, on the same sequences, frames and row sections as the other
representations, following the CRP and ADC protocol of
\citet{3dps}: the first loop's azimuth virtual array, near-field bins
zeroed. The magnitude columns of the CRP and the ADC compare each side
min-max normalized on its own, so they do not depend on absolute scale, and
the ADC envelope is compared after the rank-1 per-channel and per-range phase
correction, because a magnitude-only objective leaves that phase free. The
relative-phase columns need no correction: each scores the phase step between
neighboring samples (fast time $t$, range $r$, virtual channel $v$) against
the recording's, as the $|y|$-weighted mean of $\cos\Delta\phi$, and a
per-bin rotation cancels in that step.

\begin{table*}[htbp]
\centering
\caption{\textbf{Rendering fidelity on the ADC samples.} Table~\ref{tab:fidelity}
of the main text carries the seven columns its prose reads; this table and
Tables~\ref{tab:supp_fidelity_crp}--\ref{tab:supp_fidelity_rd_md} carry all of
them, grouped by representation, over the same sequences and the same renders,
with H, V and C as there. $\rho_{\mathrm{env}}$ is the envelope correlation
after the rank-1 phase correction; $\Delta\phi_t$ is the relative-phase
coherence along fast time ($1$ is the recording's phase step, $0$ random), with
its weighted RMSE in degrees. The unfit render is not
scored. Every column here is scale-free, so RF-Genesis is scored as rendered.
The ADC phase depends on the scene: on RT-Pose its median $\Delta\phi_t$ is
\rAdcPhaseClean{} on the clean recordings and \rAdcPhaseCluttered{} on the
cluttered ones. Bold: the better of the two methods in each column of a row
section.}
\label{tab:supp_fidelity_adc}
\footnotesize
\setlength{\tabcolsep}{5pt}
\begin{tabular}{lcccccc}
\toprule
\textbf{Method} & $\rho_{\mathrm{env}}\uparrow$ & PSNR$\uparrow$ & SSIM$\uparrow$ & RMSE$\downarrow$ & $\Delta\phi_t\uparrow$ & $\Delta\phi_t$ ($^\circ$)$\downarrow$ \\
\midrule
\multicolumn{7}{l}{\emph{HuPR}, horizontal single-chip radar (H)} \\
RF-Genesis~\cite{chen2023rfgenesis} & \rSuppAdcHuHRfg{} \\
\textbf{4DPS (Ours)} & \rSuppAdcHuHOurs{} \\
\midrule
\multicolumn{7}{l}{\emph{HuPR}, vertical single-chip radar (V)} \\
RF-Genesis & \rSuppAdcHuVRfg{} \\
\textbf{4DPS (Ours)} & \rSuppAdcHuVOurs{} \\
\midrule
\multicolumn{7}{l}{\emph{RT-Pose}, cascade (C), one-person sequences} \\
RF-Genesis & \rSuppAdcRTRfg{} \\
\textbf{4DPS (Ours)} & \rSuppAdcRTOurs{} \\
\bottomrule
\end{tabular}
\end{table*}

\begin{table*}[htbp]
\centering
\caption{\textbf{Rendering fidelity on the complex range profile (CRP).}
Sequences, renders and row sections as in Table~\ref{tab:supp_fidelity_adc}.
$\Delta\phi_r$ and $\Delta\phi_v$ are the relative-phase coherences along range
and virtual channel ($1$ is the recording's phase step, $0$ random), each with
its weighted RMSE in degrees. The unfit render is not
scored. Every column here is scale-free, so RF-Genesis is scored as rendered.
Its negative $\Delta\phi_r$ is a sampling convention: its range profiles step
by $\pi$ per bin against the recording's, and multiplying them by $(-1)^k$
makes the coherence positive and changes no other column; we report it as run.
Bold: the better of the two methods in each column of a row section.}
\label{tab:supp_fidelity_crp}
\footnotesize
\setlength{\tabcolsep}{5pt}
\begin{tabular}{lcccccccc}
\toprule
\textbf{Method} & $\rho\uparrow$ & PSNR$\uparrow$ & SSIM$\uparrow$ & RMSE$\downarrow$ & $\Delta\phi_r\uparrow$ & $\Delta\phi_r$ ($^\circ$)$\downarrow$ & $\Delta\phi_v\uparrow$ & $\Delta\phi_v$ ($^\circ$)$\downarrow$ \\
\midrule
\multicolumn{9}{l}{\emph{HuPR}, horizontal single-chip radar (H)} \\
RF-Genesis & \rSuppCrpHuHRfg{} \\
\textbf{4DPS (Ours)} & \rSuppCrpHuHOurs{} \\
\midrule
\multicolumn{9}{l}{\emph{HuPR}, vertical single-chip radar (V)} \\
RF-Genesis & \rSuppCrpHuVRfg{} \\
\textbf{4DPS (Ours)} & \rSuppCrpHuVOurs{} \\
\midrule
\multicolumn{9}{l}{\emph{RT-Pose}, cascade (C), one-person sequences} \\
RF-Genesis & \rSuppCrpRTRfg{} \\
\textbf{4DPS (Ours)} & \rSuppCrpRTOurs{} \\
\bottomrule
\end{tabular}
\end{table*}

\begin{table*}[htbp]
\centering
\caption{\textbf{Rendering fidelity on range--angle and
range--azimuth--elevation (RAE).} Sequences, renders and row sections as in
Table~\ref{tab:supp_fidelity_adc}. PSNR is
$20\log_{10}(\text{peak}/\text{RMSE})$, the only logarithm and applied to a
scalar ratio. The unfit render is not scored on the RAE family
(--). Restricted to the 57 sequences of the RT-Pose split the fitted row
reads \rSuppSplitRowRA{} on range--angle $\rho$ and $\rho_c$, so the fitted
and unfit sequence sets are comparable. RF-Genesis has no absolute scale, so
for SSIM, NRMSE and PSNR its render is multiplied before scoring by one
constant per radar that matches its loop-0 range--azimuth energy to the
recording's, the rule that initializes 4DPS's own gain
(App.~\ref{sec:supp:rfgenesis}); $\rho$ and $\rho_c$ are unchanged by it.
Bold: the best value in each column of a row section, by the column's arrow.}
\label{tab:supp_fidelity_ra_rae}
\footnotesize
\setlength{\tabcolsep}{4pt}
\begin{adjustbox}{max width=\textwidth}
\begin{tabular}{lccccccccc}
\toprule
& \multicolumn{5}{c}{\textbf{Range--angle}} & \multicolumn{4}{c}{\textbf{RAE}} \\
\cmidrule(lr){2-6}\cmidrule(lr){7-10}
\textbf{Method} & $\rho\uparrow$ & $\rho_c\uparrow$ & SSIM$\uparrow$ & NRMSE$\downarrow$ & PSNR$\uparrow$ & $\rho\uparrow$ & SSIM$\uparrow$ & NRMSE$\downarrow$ & PSNR$\uparrow$ \\
\midrule
\multicolumn{10}{l}{\emph{HuPR}, horizontal single-chip radar (H)} \\
RF-Genesis~\cite{chen2023rfgenesis} & \rSuppRaRaeHuHRfg{} \\
4DPS (unfit) & \rSuppRaRaeHuHUnfit{} \\
\textbf{4DPS (Ours)} & \rSuppRaRaeHuHOurs{} \\
\midrule
\multicolumn{10}{l}{\emph{HuPR}, vertical single-chip radar (V)} \\
RF-Genesis & \rSuppRaRaeHuVRfg{} \\
4DPS (unfit) & \rSuppRaRaeHuVUnfit{} \\
\textbf{4DPS (Ours)} & \rSuppRaRaeHuVOurs{} \\
\midrule
\multicolumn{10}{l}{\emph{RT-Pose}, cascade (C), one-person sequences} \\
RF-Genesis & \rSuppRaRaeRTRfg{} \\
4DPS (unfit) & \rSuppRaRaeRTUnfit{} \\
\textbf{4DPS (Ours)} & \rSuppRaRaeRTOurs{} \\
\bottomrule
\end{tabular}
\end{adjustbox}
\end{table*}

\begin{table*}[htbp]
\centering
\caption{\textbf{Rendering fidelity on range--Doppler (RD) and
micro-Doppler.} Sequences, renders and row sections as in
Table~\ref{tab:supp_fidelity_adc}. $P_M$ is the rendered-over-real power
ratio on the CFAR-occupied cells, $L_U$ and $\sigma_U$ the level and spread
ratios on the unoccupied ones, $\rho_{\mathrm{env}}$ the correlation of the
Doppler envelopes and BW their separation, and clDice scores connectivity
after independent segmentation of each signature. Restricted to the 57
sequences of the RT-Pose split the fitted row reads \rSuppSplitRowRDMD{} on
RD $\rho_M$ and micro-Doppler $\rho_S$, so the fitted and unfit sequence sets
are comparable. RF-Genesis has no absolute scale, so for $P_M$, $L_U$ and
$\sigma_U$ its render is multiplied before scoring by one constant per radar
that matches its loop-0 range--azimuth energy to the recording's, the rule
that initializes 4DPS's own gain (App.~\ref{sec:supp:rfgenesis}); every other
column is unchanged by it. Bold: the best value in each column of a row
section, by the column's arrow.}
\label{tab:supp_fidelity_rd_md}
\footnotesize
\setlength{\tabcolsep}{4pt}
\begin{adjustbox}{max width=\textwidth}
\begin{tabular}{lcccccccccc}
\toprule
& \multicolumn{2}{c}{\textbf{RD occupied}} & \multicolumn{2}{c}{\textbf{RD floor}} & \multicolumn{6}{c}{\textbf{Micro-Doppler}} \\
\cmidrule(lr){2-3}\cmidrule(lr){4-5}\cmidrule(lr){6-11}
\textbf{Method} & $\rho_M\uparrow$ & $P_M\to 1$ & $L_U\to 1$ & $\sigma_U\to 1$ & $\rho_\mu\uparrow$ & $\rho_\sigma\uparrow$ & $\rho_S\uparrow$ & $\rho_{\mathrm{env}}\uparrow$ & BW$\to 1$ & clDice$\uparrow$ \\
\midrule
\multicolumn{11}{l}{\emph{HuPR}, horizontal single-chip radar (H)} \\
RF-Genesis~\cite{chen2023rfgenesis} & \rSuppRdMdHuHRfg{} \\
4DPS (unfit) & \rSuppRdMdHuHUnfit{} \\
\textbf{4DPS (Ours)} & \rSuppRdMdHuHOurs{} \\
\midrule
\multicolumn{11}{l}{\emph{HuPR}, vertical single-chip radar (V)} \\
RF-Genesis & \rSuppRdMdHuVRfg{} \\
4DPS (unfit) & \rSuppRdMdHuVUnfit{} \\
\textbf{4DPS (Ours)} & \rSuppRdMdHuVOurs{} \\
\midrule
\multicolumn{11}{l}{\emph{RT-Pose}, cascade (C), one-person sequences} \\
RF-Genesis & \rSuppRdMdRTRfg{} \\
4DPS (unfit) & \rSuppRdMdRTUnfit{} \\
\textbf{4DPS (Ours)} & \rSuppRdMdRTOurs{} \\
\bottomrule
\end{tabular}
\end{adjustbox}
\end{table*}

\clearpage
\section{Downstream Results in Full}
\label{sec:supp:downstream}

Table~\ref{tab:supp_tstr_full} gives the full real/rendered train--test
matrix, of which Table~\ref{tab:tstr} keeps the rows tested on real data.
Tables~\ref{tab:supp_downstream}, \ref{tab:supp_downstream_joints}
and~\ref{tab:supp_downstream_groups} carry the downstream tables
behind Table~\ref{tab:tstr}. Table~\ref{tab:supp_downstream} gives the
retention and gap summary (retention is S$\to$R over R$\to$R, the inverse for MPJPE; the information
gap is the absolute difference between R$\to$R and S$\to$S, the realism
gap between S$\to$S and S$\to$R, and the reverse gap between R$\to$R and
R$\to$S, in the dataset's own unit) beside the
real-trained model on a 50/50 real/rendered test set,
Table~\ref{tab:supp_downstream_joints} the per-keypoint and per-joint
breakdowns with the PCK scores, and Table~\ref{tab:supp_downstream_groups} the
RT-Pose scores by activity, location, occlusion and scenario. Read the HuPR per-keypoint rows against the arms' AP: the arms
lose most of what the rendered arm loses (wrists \rKpWristLRR{} and
\rKpWristRRR{} real-to-real against \rKpWristLSR{} and \rKpWristRSR{}
rendered-to-real), while hips, knees and the head retain \rKpRetTorsoLo{} to
\rKpRetTorsoHi{} of their real-to-real value against the arms'
\rKpRetArmsLo{} to \rKpRetArmsHi{}.

\textbf{The RT-Pose baseline and its labels.} HRRadarPose~\cite{rtpose}
publishes no checkpoint; our reproduction on real data reaches
\rMPJPERealReleased{}\,cm against the 9.9\,cm the paper reports, and every
RT-Pose number in this paper is measured under that reproduction rather than
against the published figure. The released 3D labels stop before the end of
each sequence, carry a per-session scale and offset against the radar, and
label one sequence with another's radar, which is why we derive the 15 joints
on every radar frame from the registered body mesh that drives the render and
exclude sequence 13. HuPR trains at the released sampling ratio
for 200 epochs, and RT-Pose's remaining \rNseqRTOne{} one-person sequences are
split 44/6/13. Its \rNseqRTTwo{} two-person sequences are fitted against the
one person who carries pose annotations and are not scored downstream.

\textbf{Protocol details behind Sec.~\ref{sec:exp:tstr}.} RT-Pose's
44/6/13 split is stratified by location, scenario, activity and occlusion, so
that no split is carried by one room or one action. On rendered test data the
released decoder's confidence threshold is lowered from its released value to
$-1$, so that every joint is kept and a rendered frame is never scored on
fewer joints than its real twin; the threshold never triggers on real data,
so no real number in the paper depends on it.

\begin{table}[htbp]
\centering
\caption{\textbf{Real and rendered as train and test domains, the full matrix.} Table~\ref{tab:tstr} keeps the rows tested on real data. All six metrics
are pose-estimation metrics, and every cell is retrained from scratch for four
seeds with the standard deviation in parentheses. The first block is the
real-only reference, the second is RF-Genesis and the third the same four arms
under 4DPS. Rows tested on a renderer measure how much pose information that
render carries, not accuracy on real data. RF-Genesis's RT-Pose inputs carry
one level gain fixed before training, because its units are arbitrary
(App.~\ref{sec:supp:rfgenesis}).}
\label{tab:supp_tstr_full}
\small
\setlength{\tabcolsep}{2.8pt}
\begin{adjustbox}{max width=\linewidth}
\begin{tabular}{llcccccc}
\toprule
& & \multicolumn{3}{c}{\textbf{RT-Pose} (cm $\downarrow$)} & \multicolumn{3}{c}{\textbf{HuPR} (AP $\uparrow$)} \\
\cmidrule(lr){3-5}\cmidrule(lr){6-8}
\textbf{Train} & \textbf{Test} & MPJPE & MRPE & Abs-MPJPE & AP & AP$_{50}$ & AP$_{75}$ \\
\midrule
Real & Real & \rMPJPEReal{} & \rTstrRTMRPERR{} & \rTstrRTAbsMPJPERR{} & \rApReal{} & \rTstrHuAPfiftyRR{} & \rTstrHuAPsevfiveRR{} \\
\midrule
Real & RF-Genesis & \rTstrRfgRTMPJPERS{} & \rTstrRfgRTMRPERS{} & \rTstrRfgRTAbsMPJPERS{} & \rTstrRfgHuAPRS{} & \rTstrRfgHuAPfiftyRS{} & \rTstrRfgHuAPsevfiveRS{} \\
RF-Genesis & Real & \rTstrRfgRTMPJPESR{} & \rTstrRfgRTMRPESR{} & \rTstrRfgRTAbsMPJPESR{} & \rTstrRfgHuAPSR{} & \rTstrRfgHuAPfiftySR{} & \rTstrRfgHuAPsevfiveSR{} \\
RF-Genesis & RF-Genesis & \rTstrRfgRTMPJPESS{} & \rTstrRfgRTMRPESS{} & \rTstrRfgRTAbsMPJPESS{} & \rTstrRfgHuAPSS{} & \rTstrRfgHuAPfiftySS{} & \rTstrRfgHuAPsevfiveSS{} \\
Real $+$ RF-Genesis & Real & \rTstrRfgRTMPJPEMix{} & \rTstrRfgRTMRPEMix{} & \rTstrRfgRTAbsMPJPEMix{} & \rTstrRfgHuAPMix{} & \rTstrRfgHuAPfiftyMix{} & \rTstrRfgHuAPsevfiveMix{} \\
\midrule
Real & 4DPS & \rMPJPESub{} & \rTstrRTMRPERS{} & \rTstrRTAbsMPJPERS{} & \rApSub{} & \rTstrHuAPfiftyRS{} & \rTstrHuAPsevfiveRS{} \\
4DPS & Real & \rMPJPETSTR{} & \rTstrRTMRPESR{} & \rTstrRTAbsMPJPESR{} & \rApTSTR{} & \rTstrHuAPfiftySR{} & \rTstrHuAPsevfiveSR{} \\
4DPS & 4DPS & \rMPJPESelf{} & \rTstrRTMRPESS{} & \rTstrRTAbsMPJPESS{} & \rApSelf{} & \rTstrHuAPfiftySS{} & \rTstrHuAPsevfiveSS{} \\
Real $+$ 4DPS & Real & \rMPJPEMixed{} & \rTstrRTMRPEMix{} & \rTstrRTAbsMPJPEMix{} & \rApMixed{} & \rTstrHuAPfiftyMix{} & \rTstrHuAPsevfiveMix{} \\
\bottomrule
\end{tabular}
\end{adjustbox}
\end{table}

\begin{table*}[htbp]
\centering
\caption{\textbf{Downstream transfer summary.} Retention and the
three gaps are defined in the text and read in the dataset's own unit (AP
$\times100$ on HuPR, MPJPE in cm on RT-Pose); the last two columns score the
real-trained model on a 50/50 real/rendered test set and on the real test
set, mean (std) over four seeds.}
\label{tab:supp_downstream}
\footnotesize
\setlength{\tabcolsep}{3pt}
\begin{tabular}{lcccccc}
\toprule
 & & & & & \multicolumn{2}{c}{R$\to$R model} \\
\cmidrule(lr){6-7}
\textbf{Dataset} & Retention & Information gap & Realism gap & Reverse gap & 50/50 test & Real test \\
\midrule
HuPR (AP, pts) & 0.59 & 6.0 & 21.2 & 16.4 & 55.8 (1.1) & 65.6 (1.7) \\
RT-Pose (MPJPE, cm) & 0.76 & 1.0 & 6.1 & 3.4 & 18.93 (1.44) & 16.67 (0.42) \\
\bottomrule
\end{tabular}

\end{table*}

\begin{table*}[htbp]
\centering
\caption{\textbf{Per-keypoint and per-joint downstream accuracy}, per arm
(train$\to$test; R real, S rendered by 4DPS), same protocol as
Table~\ref{tab:supp_downstream}: HuPR AP $\times100$ per keypoint and
PCKh@0.2 (top), RT-Pose MPJPE in cm per joint and PCK at 10 and 15\,cm
(bottom). MPJPE is taken relative to the pelvis, so the pelvis reads 0; PCK on
RT-Pose is scored on real test data only (--). Mean (std) over four seeds.}
\label{tab:supp_downstream_joints}
\footnotesize
\setlength{\tabcolsep}{4pt}
\begin{tabular}{lccccc}
\toprule
\textbf{HuPR keypoint (AP)} & R$\to$R & S$\to$R & R$\to$S & S$\to$S & R+S$\to$R \\
\midrule
R\_Hip & 90.1 (0.6) & 72.1 (2.9) & 82.7 (1.7) & 91.5 (0.8) & 89.4 (0.2) \\
R\_Knee & 84.5 (0.7) & 65.3 (2.8) & 76.0 (1.3) & 87.7 (0.6) & 83.8 (0.5) \\
R\_Ankle & 73.3 (1.7) & 50.2 (2.6) & 73.9 (2.2) & 73.6 (1.3) & 71.8 (1.6) \\
L\_Hip & 90.6 (1.3) & 71.7 (3.3) & 83.0 (1.4) & 90.8 (1.3) & 88.9 (0.6) \\
L\_Knee & 83.7 (2.2) & 61.1 (3.3) & 75.7 (1.0) & 85.2 (2.6) & 82.1 (0.9) \\
L\_Ankle & 77.5 (1.3) & 63.0 (2.0) & 66.2 (1.3) & 80.8 (1.0) & 75.0 (1.0) \\
Neck & 83.1 (1.8) & 64.7 (4.1) & 73.2 (1.7) & 82.0 (1.2) & 80.4 (1.7) \\
Head & 81.4 (1.1) & 63.3 (4.0) & 70.1 (1.0) & 79.9 (1.1) & 78.9 (1.0) \\
L\_Shoulder & 74.7 (1.2) & 53.3 (4.5) & 58.0 (2.0) & 71.6 (1.2) & 72.4 (0.4) \\
L\_Elbow & 41.1 (1.4) & 16.2 (2.6) & 19.5 (1.0) & 27.6 (1.4) & 37.1 (1.3) \\
L\_Wrist & 22.2 (1.8) & 8.0 (1.1) & 7.1 (0.5) & 13.3 (0.8) & 18.1 (0.7) \\
R\_Shoulder & 76.8 (2.3) & 52.7 (3.6) & 63.7 (3.3) & 73.7 (1.7) & 74.1 (0.9) \\
R\_Elbow & 50.3 (4.1) & 16.1 (2.7) & 25.2 (0.9) & 36.8 (2.1) & 43.6 (2.4) \\
R\_Wrist & 25.1 (3.0) & 4.7 (0.8) & 9.8 (0.7) & 16.1 (2.3) & 19.3 (1.1) \\
\midrule
PCKh@0.2 & 43.6 (1.7) & 25.0 (1.8) & 27.3 (2.2) & 39.2 (2.2) & 41.7 (1.6) \\
\bottomrule
\end{tabular}

\begin{tabular}{lccccc}
\toprule
\textbf{RT-Pose joint (MPJPE, cm)} & R$\to$R & S$\to$R & R$\to$S & S$\to$S & R+S$\to$R \\
\midrule
Pelvis & 0.0 (0.0) & 0.0 (0.0) & 0.0 (0.0) & 0.0 (0.0) & 0.0 (0.0) \\
Right\_Hip & 9.5 (1.0) & 11.4 (0.9) & 11.4 (1.5) & 9.0 (0.4) & 9.2 (0.8) \\
Right\_Knee & 14.2 (1.4) & 18.6 (0.8) & 17.2 (2.4) & 20.4 (14.9) & 15.5 (1.1) \\
Right\_Ankle & 17.3 (0.9) & 23.0 (1.1) & 19.9 (1.3) & 15.2 (0.5) & 18.9 (1.0) \\
Left\_Hip & 9.5 (0.9) & 12.1 (0.9) & 11.1 (1.7) & 9.4 (0.7) & 9.0 (0.8) \\
Left\_Knee & 13.6 (0.5) & 19.7 (2.1) & 16.3 (1.4) & 13.3 (0.7) & 15.1 (1.2) \\
Left\_Ankle & 17.0 (0.6) & 22.0 (1.2) & 19.4 (1.2) & 15.8 (1.2) & 18.1 (1.2) \\
Thorax & 10.6 (0.8) & 13.8 (0.7) & 12.5 (1.2) & 10.0 (0.8) & 10.0 (1.0) \\
Head & 14.0 (0.8) & 16.4 (1.6) & 15.4 (1.4) & 13.1 (0.4) & 13.5 (0.2) \\
Left\_Shoulder & 13.2 (1.1) & 16.6 (0.8) & 15.2 (1.5) & 12.7 (0.8) & 13.1 (0.6) \\
Left\_Elbow & 23.4 (0.4) & 28.5 (1.2) & 26.6 (2.0) & 21.4 (0.5) & 23.1 (0.8) \\
Left\_Wrist & 32.6 (2.0) & 46.7 (6.5) & 37.5 (4.5) & 29.7 (2.4) & 30.7 (0.8) \\
Right\_Shoulder & 13.1 (0.9) & 16.3 (0.6) & 15.6 (0.9) & 12.6 (0.3) & 13.0 (0.9) \\
Right\_Elbow & 24.0 (0.8) & 30.0 (2.1) & 27.6 (1.8) & 21.6 (0.4) & 23.9 (1.5) \\
Right\_Wrist & 38.1 (4.2) & 52.7 (7.7) & 54.9 (18.8) & 31.6 (3.5) & 34.7 (1.3) \\
\midrule
PCK@10cm & 45.8 (2.0) & 35.9 (1.5) & -- & -- & 44.8 (2.3) \\
PCK@15cm & 64.3 (1.9) & 54.0 (1.7) & -- & -- & 62.1 (2.7) \\
\bottomrule
\end{tabular}

\end{table*}

\begin{table}[htbp]
\centering
\caption{\textbf{RT-Pose MPJPE (cm) by group}: the test sequences' activity,
location, occlusion and scenario, for the three arms scored on real test data,
same protocol as Table~\ref{tab:supp_downstream}; mean (std) over four seeds.}
\label{tab:supp_downstream_groups}
\footnotesize
\setlength{\tabcolsep}{4pt}
\begin{tabular}{llccc}
\toprule
\textbf{RT-Pose MPJPE (cm) by group} & & R$\to$R & S$\to$R & R+S$\to$R \\
\midrule
activity & Stand and Lift Leg & 9.81 (0.20) & 14.45 (1.60) & 9.77 (0.27) \\
activity & Stand and Wave Hand & 17.08 (1.08) & 22.03 (0.81) & 18.66 (1.70) \\
activity & Walk & 21.67 (2.14) & 25.69 (2.54) & 19.23 (0.43) \\
activity & Walk and Sit & 10.43 (0.51) & 25.28 (1.00) & 10.77 (1.20) \\
activity & Walk and Wave Hand & 14.58 (0.28) & 17.05 (1.31) & 14.18 (0.41) \\
location & Indoor & 19.40 (0.88) & 23.89 (1.63) & 17.92 (1.09) \\
location & Outdoor & 12.30 (0.63) & 18.60 (0.89) & 14.27 (0.67) \\
occlusion & Cardboard & 25.66 (1.99) & 24.13 (1.74) & 25.31 (0.77) \\
occlusion & Nothing & 14.62 (1.15) & 20.72 (1.24) & 13.53 (0.73) \\
occlusion & Plastic pad & 16.87 (1.86) & 24.68 (2.81) & 21.16 (1.65) \\
scenario & Clean & 16.55 (0.39) & 22.08 (1.32) & 15.30 (0.69) \\
scenario & Cluttered & 21.11 (1.29) & 24.98 (2.07) & 19.49 (1.33) \\
scenario & Normal & 12.61 (0.77) & 19.67 (1.17) & 15.17 (0.59) \\
scenario & Rain & 11.06 (0.69) & 14.36 (0.26) & 10.69 (1.14) \\
\bottomrule
\end{tabular}

\end{table}

\clearpage
\section{Representation Analysis Protocol}
\label{sec:supp:repr}

We fixed this protocol, including every measure and the $1.2\times$ margin,
before running any comparison, so the analysis is pre-registered; where a
measurement departs from it, we say so. HuPR follows RT-Pose's design and
changes only what the dataset forces. The protocol's pass rule is the point
estimate; we report a verdict only when the 95\% interval clears the margin.

\textbf{Models and comparisons.} We compare \rReprHuTrained{} checkpoints per
dataset: five real-trained seeds and the four rendered- and four
mixed-trained seeds (0--3) of Table~\ref{tab:tstr}. All arms of a seed start
from the same weights, and a seed's real and rendered arms see their batches
in the same order. Table~\ref{tab:supp_repr_cells} lists the comparisons and
what differs within each; every ratio is taken against the real--real
comparison of the same row. The different-initialization row is complete and
carries the claim. The same-initialization row has no real--real reference
of its own, because we trained no second real arm per seed, so it is read
against the different-initialization real--real mean, which bounds its ratio
from above. An untrained network is the only control.

\begin{table}[htbp]
\centering
\caption{\textbf{The cells compared}, with what differs inside each. Ratios
are taken against the real--real cell of the matched row. No real arm was
trained twice from one initialization, so the first row has no real--real
reference and its cross pairs are read against the different-initialization
real--real mean, which bounds their ratio from above.}
\label{tab:supp_repr_cells}
\small
\begin{tabular}{llccl}
\toprule
\textbf{Row} & \textbf{Cell} & \textbf{Pairs} & \textbf{Differs in} & \textbf{Reference} \\
\midrule
same init & real$_k$ vs rendered$_k$   & \rReprNXrSame{}  & data                & (none trained) \\
\midrule
diff.\ init & real$_i$ vs real$_j$     & \rReprNRr{} & init, order         & itself \\
            & real$_i$ vs rendered$_j$ & \rReprNXr{} & init, order, data   & real--real, diff.\ init \\
\midrule
beside the design & mixed vs real       & \rReprNMxSame{}, \rReprNMx{} & data (and init) & real--real, diff.\ init \\
            & rendered vs rendered      & \rReprNSs{} & init, order         & real--real, diff.\ init \\
            & untrained vs real         & \rReprNFloor{}  & training itself     & real--real, diff.\ init \\
\bottomrule
\end{tabular}
\end{table}

\textbf{Frames and layers.} On RT-Pose we use every annotated frame of the
13 real test sequences (\rReprFrames{} frames) and the same frames of the
rendered test split. The layers (taps) are the input (T0), HRNet layer~1
(T1), the parallel branches of stages 2 to 4 (T2a--b, T3a--c, T4a--d; T4a is
also the backbone output) and the head's pre-activation maps (T6). On HuPR a
frame stacks eight frames of eight chirps on two boards, so error
consistency and paired inputs use all \rReprHuFrames{} test frames, while
similarity uses every tenth (1,260 frames, every sequence represented) and
the readout is fitted on every tenth frame of the real training split. Its
layers are each board's input (H0), the chirp network's output (H1), the
three encoder levels (H2--H4; H4 feeds the decoder) and the decoder's
pre-sigmoid heatmap (H5); the two boards are never averaged. At every layer
we subsample 1,024 spatial positions (fewer where the grid is smaller),
drawn once and shared by every frame and model, and treat each position as a
row of channels.

\textbf{Measures.} Error consistency~\cite{geirhos2020error} is the
correlation of per-frame error between two models and their per-joint
agreement, where each model's per-frame error at a joint is split at that
model's own median. On HuPR a keypoint is a hit at a per-keypoint OKS of at
least 0.5, and Cohen's $\kappa$ over (frame, keypoint) pairs is the primary
statistic. Similarity uses debiased linear CKA~\cite{nguyen2021wide},
computed exactly and read as the distance $1-\text{CKA}$ in every ratio, and the angular orthogonal Procrustes
distance~\cite{williams2021shape}, because Procrustes catches functional
changes CKA misses~\cite{ding2021grounding}. The fully whitened Procrustes
stays inside the margin at every layer (\rReprWhitenLo{} to
\rReprWhitenHi{}) but barely separates an untrained network, so we do not
use it. Intervals come from 1,000 bootstrap replicates that resample frames
and model pairs.

\textbf{Checks.} Seed 0's real-trained checkpoint re-scores to
\rReprGateLocal{}\,cm against \rReprGateNode{} in the run behind
Table~\ref{tab:tstr} on RT-Pose, and to \rReprHuGateLocal{} against
\rReprHuGateNode{} AP on HuPR, under the same protocol. A model
against itself gives CKA 1 and Procrustes 0 at every layer, the untrained
network lies above every real--real band after the input, and the rendered
test split repeats the similarity picture layer for layer.

\begin{table}[htbp]
\centering
\caption{\textbf{Representational similarity by tap} (RT-Pose, real test frames). CKA is a similarity and the Procrustes distance is in
radians; \emph{ratio} is the real--rendered different-init distance over the
real--real different-init distance, with its 95\% interval, against the
pre-registered margin of 1.2. \emph{Floor} is an untrained network. T0 is the
input tensor, identical for every model.}
\label{tab:supp_repr}
\small
\setlength{\tabcolsep}{4pt}
\begin{adjustbox}{max width=\textwidth}
\begin{tabular}{lcccl|cccl}
\toprule
& \multicolumn{4}{c|}{\textbf{debiased CKA} (similarity $\uparrow$)} & \multicolumn{4}{c}{\textbf{Procrustes} (distance, rad $\downarrow$)} \\
\textbf{Tap} & real--real & real--rend. & floor & ratio & real--real & real--rend. & floor & ratio \\
\midrule
T0 input   & \rReprSimRowTZeroinput{} \\
T1 layer 1 & \rReprSimRowTOnelayerOne{} \\
T2a        & \rReprSimRowTTwoa{} \\
T2b        & \rReprSimRowTTwob{} \\
T3a        & \rReprSimRowTThreea{} \\
T3b        & \rReprSimRowTThreeb{} \\
T3c        & \rReprSimRowTThreec{} \\
T4a (= T5) & \rReprSimRowTFoura{} \\
T4b        & \rReprSimRowTFourb{} \\
T4c        & \rReprSimRowTFourc{} \\
T4d        & \rReprSimRowTFourd{} \\
T6 head    & \rReprSimRowTSixhead{} \\
\bottomrule
\end{tabular}
\end{adjustbox}
\end{table}

\textbf{Error consistency.} The two models do not fail on the same frames.
Relative to two real-trained models of different seeds, they disagree about
which joints they get right at \rReprAratio{} on RT-Pose and
\rReprHuAratio{} on HuPR, and their per-frame errors correlate less, at
\rReprRratio{} and \rReprHuRratio{} (Table~\ref{tab:supp_repr_errcons}).
Two real-trained models also disagree on frames, which is why each value is
a ratio to that disagreement. Behind the ratios, two real-trained seeds
correlate at \rReprRrr{} in per-frame MPJPE and agree on \rReprArr{} of the
per-joint indicators, against \rReprRxr{} and \rReprAxr{} across domains; on
HuPR they agree on \rReprHuArr{} of the keypoint hits against \rReprHuAxr{},
and $\kappa$ falls from \rReprHuKrr{} to \rReprHuKxr{}. On the rendered test
split the RT-Pose ratios fall to \rReprRratioS{} and \rReprAratioS{}, and
the mixed arm is indistinguishable from a real seed (ratios
\rReprMixConsLo{} to \rReprMixConsHi{}). We report this here rather than in
Sec.~\ref{sec:exp:repr} because it concerns which inputs each model gets
wrong, not what the models learned.

\begin{table}[htbp]
\centering
\caption{\textbf{Do rendered- and real-trained models fail on the same
frames?} Error consistency as the real--rendered value over the real--real
value between seeds, with its 95\% interval (1.0: as consistent as two
real-trained models). Bold: the interval lies above the $1.2\times$ margin.
The cells behind each ratio follow.}
\label{tab:supp_repr_errcons}
\small
\setlength{\tabcolsep}{6pt}
\newcommand{\rcell}[2]{\shortstack{#1\\[1pt]{\tiny #2}}}
\begin{tabular}{lcc}
\toprule
& \shortstack{Frame\\error corr.} & \shortstack{Joint\\error agr.} \\
\midrule
RT-Pose & \rcell{\rReprRratio{}}{\rReprRci{}} & \rcell{\textbf{\rReprAratio{}}}{\rReprAci{}} \\
HuPR & \rcell{\textbf{\rReprHuRratio{}}}{\rReprHuRci{}} & \rcell{\textbf{\rReprHuAratio{}}}{\rReprHuAci{}} \\
\bottomrule
\end{tabular}
\end{table}

\begin{table}[htbp]
\centering
\caption{\textbf{Representational similarity by tap} (HuPR, real test frames), as Table~\ref{tab:supp_repr} for RT-Pose. H1 is the
chirp network's output, H4 the tensor the decoder reads and H5 its
pre-sigmoid heatmap; h and v are the horizontal and vertical boards. The
level-1 and level-2 encoder taps are undefined, because they carry no signal
in some checkpoints (App.~\ref{sec:supp:repr}, dead temporal merges). The rendered probe reproduces
every computable cell to within \rReprHuProbeAgree{}.}
\label{tab:supp_repr_hupr}
\small
\setlength{\tabcolsep}{4pt}
\begin{tabular}{lcccl|c}
\toprule
& \multicolumn{4}{c|}{\textbf{debiased CKA} (similarity $\uparrow$)} & \textbf{Procrustes} \\
\textbf{Tap} & real--real & real--rend. & floor & ratio & ratio \\
\midrule
H1h chirp output    & \rReprHuCkaChirprr{} & \rReprHuCkaChirpxr{} & \rReprHuCkaChirpfloor{} & \rReprHuCkaChirpratio{} \rReprHuCkaChirpci{} & \rReprHuProChirpratio{} \\
H2h, H3h, H2v, H3v  & \multicolumn{4}{c|}{\emph{undefined: the tap is dead in some checkpoints}} & -- \\
H4h encoder level 3 & \rReprHuCkaEncrr{} & \rReprHuCkaEncxr{} & \rReprHuCkaEncfloor{} & \rReprHuCkaEncratio{} \rReprHuCkaEncci{} & \rReprHuProEncratio{} \\
H4v encoder level 3 & \rReprHuCkaVertrr{} & \rReprHuCkaVertxr{} & \rReprHuCkaVertfloor{} & \rReprHuCkaVertratio{} \rReprHuCkaVertci{} & \rReprHuProVertratio{} \\
H5 decoder logits   & \rReprHuCkaHeadrr{} & \rReprHuCkaHeadxr{} & \rReprHuCkaHeadfloor{} & \rReprHuCkaHeadratio{} \rReprHuCkaHeadci{} & \rReprHuProHeadratio{} \\
\bottomrule
\end{tabular}
\end{table}

\textbf{Mixed arm and shared initialization.} The mixed arm sits between
\rReprMixLo{} and \rReprMixHi{} and passes 1.2 only in the point estimate at
\rReprMixOutTap{} (\rReprMixOutRatio{}, interval \rReprMixOutCi{}). The
shared initialization explains little of the agreement: the
same-initialization pairs are no closer than the different-initialization
ones from stage~2 onward (\rReprSameLo{} to \rReprSameHi{}).

\textbf{Linear readout.} We freeze each backbone and fit a closed-form ridge
readout on real training data to the joint coordinates (RT-Pose: T5 pooled
to $4\times8\times16$; HuPR: H4h and H4v pooled to $4\times4$), with one
penalty chosen on held-out real frames and applied to every backbone, and
score it with the released evaluator. The registered form, a
$1\times1\times1$ convolution trained under the released losses, converged
on no backbone, the real ones included, so we use the plan's alternative. On
RT-Pose the rendered-trained models lose \rProbeHeadDrop{}\% through their
own decoders and \rProbeDrop{}\% through the readout (\rProbeRend{} against
\rProbeReal{}\,cm, ratio \rProbeRatio{}, interval \rProbeCi{}), but an
untrained backbone probes at only \rProbeRatioFloor{}, so the test barely
discriminates there. On HuPR the real, rendered, mixed and untrained
backbones reach \rReprHuProbeRealPx{}, \rReprHuProbeRendPx{},
\rReprHuProbeMixedPx{} and \rReprHuProbeFloorPx{}\,px of keypoint error, and
the untrained ratio of \rReprHuProbeFloorRatio{} is what gives this test its
power on HuPR.

\textbf{Paired inputs.} Every rendered frame has a real twin at the same
timestamp, so we measure the feature distance between the twins inside one
model, in units of the distance between neighboring real frames. To a
real-trained model the render's departure from the recording is worth
\rPairRealLo{} to \rPairRealHi{} frames of real motion, and it does not
shrink with depth. A rendered-trained model sees a smaller gap
(\rPairRendLo{} to \rPairRendHi{}), so training on renders makes a model
less sensitive to the domain difference without closing it. On HuPR we use
all \rReprHuFrames{} test frames rather than the registered tenth, so that a
neighbor is one frame away as on RT-Pose; there the twin distance at the
chirp network's output is \rReprHuPairRealChirp{} for a real-trained and
\rReprHuPairRendChirp{} for a rendered-trained model, so the two part only
after the encoder.

\textbf{Dead layers on HuPR.} MSCSA-PRGCN merges its frames with bias-free
convolutions, and training drives some of them to denormal magnitude, so the
layer after such a merge carries no signal. Of the \rReprHuTrained{}
checkpoints, H2h and H3h are dead in \rReprHuDeadHtwoh{} and
\rReprHuDeadHthreeh{}, and H2v and H3v in \rReprHuDeadHtwov{} and
\rReprHuDeadHthreev{}. A pair containing a dead layer is undefined, so we
report those layers as undefined rather than substitute another.

\clearpage
\section{Scene Editing}
\label{sec:supp:editing}

4DPS fits the static scene and the person as separate explicit components. We can therefore place a fitted person
in another fitted scene, or give a fitted scene a motion that was never recorded, and render again without
refitting. We test the first edit against real recordings (App.~\ref{sec:supp:editing:separability}). The second
edit has no real recording to compare against, so we show it without a score
(App.~\ref{sec:supp:editing:novel}).

\subsection{Separability of the scene and the person}
\label{sec:supp:editing:separability}

\textbf{The test.} We ask whether a fitted person keeps its fidelity when the scene around it changes. For two
recordings $A$ and $B$ of the same environment, we render $B$'s complete fit inside $A$'s fitted static scene.
$B$'s fit is its registered mesh, person material, per-frame velocity field and normals, and joint gains; $A$'s
scene is its points, their orientations and their materials. We then score the render against $B$'s real recording, on $B$'s frames, with
the evaluation of Table~\ref{tab:fidelity}. $B$'s receiver gain, noise floor and antenna coupling stay with $B$,
because they describe $B$'s capture and not the room. The only other change is the person's shadow on the scene,
which we cast again for $A$'s scene points with $B$'s mesh, and nothing is refitted. Rendering $B$'s fit in $B$'s
own scene through the same code reproduces $B$'s own render.

\textbf{The pairs.} Two recordings count as the same environment when the time-averaged zero-Doppler
range--azimuth maps of their real data correlate at \rEditSceneThr{} or more. We correlate the maps in dB over the retained range bins, and on
HuPR we take the lower of the two boards. We set this threshold on RT-Pose, where the largest correlation between
two recordings of different labeled scenes is \rEditCrossMax{}. Each pair comes from one recording session (one
subject in one set of clothes), with one sequence standing and the other moving. We take \rEditNpairRT{} such
pairs on RT-Pose and \rEditNpairHP{} on HuPR, evenly spaced over the candidates, whose static returns correlate
at \rEditPairRhoMinRT{} and \rEditPairRhoMinHP{} or more. Rendering every pair in both directions gives
\rEditNrenderRT{} and \rEditNrenderHP{} renders.

\begin{figure}[htbp]
  \centering
  \includegraphics[width=\linewidth]{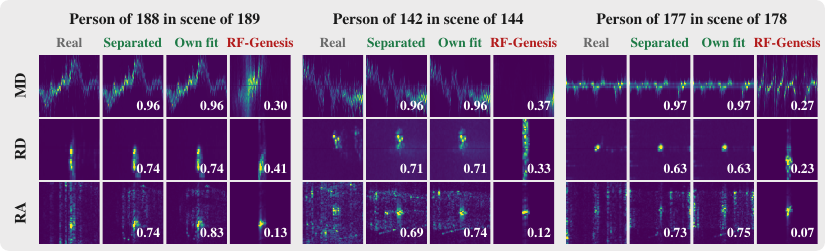}
  \\[4pt]
  \includegraphics[width=\linewidth]{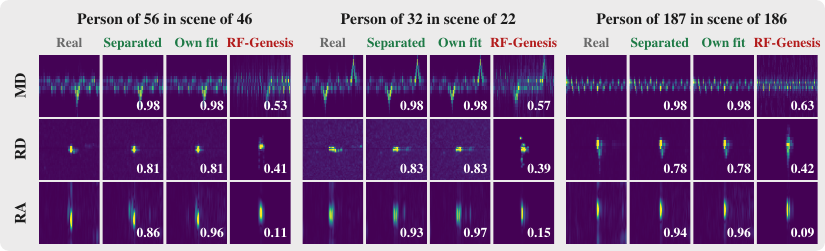}
  \caption{\textbf{Separability, RT-Pose (top) and HuPR's horizontal board (bottom).} Each block renders the person
  of one recording in the fitted scene of another recording of the same environment. Columns: the recording
  (Real), the person's complete fit rendered in the other scene (Separated), the person's own fit, and RF-Genesis
  driven by the same registered motion. Rows, frame, normalization and scores are those of
  Fig.~\ref{fig:qualitative}. We chose the pairs by a rule: the renders at the 25th, 50th and 75th percentile of
  the mean ratio of the separated render's three panel scores to the own fit's.}
  \label{fig:supp_separability}
\end{figure}

\textbf{Result.} Table~\ref{tab:supp_separability} shows that the render in the other scene keeps all of the
person's fidelity and most of the scene's. The range--Doppler and micro-Doppler columns and $L_U$ equal $B$'s own
fit: the median paired difference in micro-Doppler $\rho_S$ is \rEditDOwnMDSRT{} on RT-Pose and
\rEditDOwnMDSHP{} on HuPR. These columns are not evidence that the scene transfers, however, because they would
match even if it did not. The range--Doppler and micro-Doppler products remove the zero-Doppler return, and our
renderer is single-bounce, so the static scene reaches them only through the person's shadow; the noise floor
behind $L_U$ stays with $B$. The evidence is in the range--azimuth columns, which the scene dominates. There,
$A$'s scene explains $B$'s static return at \rEditRetRART\,\% of $B$'s own fit on $\rho$ on RT-Pose and
\rEditRetRAHP\,\% on HuPR (median paired differences \rEditDOwnRART{} and \rEditDOwnRAHP{}). On the
clutter-removed $\rho_c$, the share is \rEditRetRAcRT\,\% on RT-Pose and \rEditRetRAcHP\,\% on HuPR, and
HuPR's is the largest loss in the table. The render in the other scene still scores above $B$'s unfit render and RF-Genesis on most
renders in every column. The one exception is HuPR's range--azimuth $\rho$ against the unfit render, where it is
higher on \rEditWUnfitRAHP\,\% of renders. Fig.~\ref{fig:supp_separability} shows three pairs per dataset.

\begin{table}[htbp]
\centering
\caption{\textbf{Separability of the scene and the person.} $B$'s complete fit rendered in $A$'s fitted static
scene (``Separated''), scored against $B$'s real recording with the evaluation of Table~\ref{tab:fidelity},
beside $B$'s own fit, $B$'s unfit render and RF-Genesis driven by $B$'s registered motion, on the same renders and
frames. Medians over renders; HuPR pools its two boards, as Table~\ref{tab:fidelity} does, so $n$ counts
(render, board) products. Retention is the median of the separated render over the median of $B$'s own fit, on
the same renders. RF-Genesis's $L_U$ comes from its level-matched scoring, as in Table~\ref{tab:fidelity}.
A column undefined on a render (no frame of the evaluation's motion subset) has a smaller $n$ than the row's,
which the column shows.}
\label{tab:supp_separability}
\small
\setlength{\tabcolsep}{4pt}
\begin{tabular}{@{}lrrrrrrrr@{}}
\toprule
 & \multicolumn{2}{c}{range--azimuth} & \multicolumn{2}{c}{range--Doppler} & \multicolumn{3}{c}{micro-Doppler} & \\
\cmidrule(lr){2-3}\cmidrule(lr){4-5}\cmidrule(lr){6-8}
 & $\rho$ & $\rho_c$ & $\rho_M$ & $L_U$ & $\rho_\mu$ & $\rho_\sigma$ & $\rho_S$ & $n$ \\
\midrule
\multicolumn{9}{@{}l}{\emph{RT-Pose, \rEditNpairRT{} pairs, both directions}} \\
Separated ($A$'s scene, $B$'s fit) & \rEditSepRART & \rEditSepRAcRT & \rEditSepRDoccRT & \rEditSepLURT & \rEditSepMDcRT & \rEditSepMDsRT & \rEditSepMDSRT & \rEditSepNRT \\
$B$'s own fit & \rEditOwnRART & \rEditOwnRAcRT & \rEditOwnRDoccRT & \rEditOwnLURT & \rEditOwnMDcRT & \rEditOwnMDsRT & \rEditOwnMDSRT & \rEditOwnNRT \\
Retention (\%) & \rEditRetRART & \rEditRetRAcRT & \rEditRetRDoccRT & \rEditRetLURT & \rEditRetMDcRT & \rEditRetMDsRT & \rEditRetMDSRT & \\
$B$'s unfit render & \rEditUnfitRART & \rEditUnfitRAcRT & \rEditUnfitRDoccRT & \rEditUnfitLURT & \rEditUnfitMDcRT & \rEditUnfitMDsRT & \rEditUnfitMDSRT & \rEditUnfitNRT \\
RF-Genesis & \rEditRfgRART & \rEditRfgRAcRT & \rEditRfgRDoccRT & \rEditRfgLURT & \rEditRfgMDcRT & \rEditRfgMDsRT & \rEditRfgMDSRT & \rEditRfgNRT \\
\midrule
\multicolumn{9}{@{}l}{\emph{HuPR, \rEditNpairHP{} pairs, both directions}} \\
Separated ($A$'s scene, $B$'s fit) & \rEditSepRAHP & \rEditSepRAcHP & \rEditSepRDoccHP & \rEditSepLUHP & \rEditSepMDcHP & \rEditSepMDsHP & \rEditSepMDSHP & \rEditSepNHP \\
$B$'s own fit & \rEditOwnRAHP & \rEditOwnRAcHP & \rEditOwnRDoccHP & \rEditOwnLUHP & \rEditOwnMDcHP & \rEditOwnMDsHP & \rEditOwnMDSHP & \rEditOwnNHP \\
Retention (\%) & \rEditRetRAHP & \rEditRetRAcHP & \rEditRetRDoccHP & \rEditRetLUHP & \rEditRetMDcHP & \rEditRetMDsHP & \rEditRetMDSHP & \\
$B$'s unfit render & \rEditUnfitRAHP & \rEditUnfitRAcHP & \rEditUnfitRDoccHP & \rEditUnfitLUHP & \rEditUnfitMDcHP & \rEditUnfitMDsHP & \rEditUnfitMDSHP & \rEditUnfitNHP \\
RF-Genesis & \rEditRfgRAHP & \rEditRfgRAcHP & \rEditRfgRDoccHP & \rEditRfgLUHP & \rEditRfgMDcHP & \rEditRfgMDsHP & \rEditRfgMDSHP & \rEditRfgNHP \\
\bottomrule
\end{tabular}
\end{table}

\subsection{Novel motions}
\label{sec:supp:editing:novel}

\textbf{Source and placement.} Data augmentation for new activities and for falls needs motions that neither
dataset contains, so we render such motions in fitted scenes. We take jumping jacks, squats, kicks and one real
fall, a rug pulled from under the subject, from the CMU motion-capture set in
AMASS~\citep{mahmood2019amass}. CMU has no clean forward or backward fall, so we generate those two with
MDM~\citep{tevet2023mdm}, a text-to-motion model trained on HumanML3D~\citep{guo2022humanml3d}. MDM's samples
often end in a different posture from their prompt. We therefore drew three samples per prompt and kept the one
whose joints do what the prompt says. We judge that by the pelvis's travel relative to the radar and by where the
toes point at the end. Each clip is labeled by that measured outcome. We pose each clip with the body shape fitted to the
scene's own subject and resample it to the radar's 10\,Hz frame rate. We then place it on the scene's floor at the
subject's real position, facing the radar. No scene point more than 15\,cm above the floor may come within 5\,cm
of the body in any frame. Where the subject's own position fails that check, we move the start to the nearest
point within 1\,m that passes, which moved \rNovelNmoved{} of the \rNovelNmotion{}\,$\times$\,\rNovelNscene{}
renders by at most \rNovelMaxShift\,m. The render then follows App.~\ref{sec:supp:editing:separability}: the
scene, its materials, noise floor and gain, and the subject's person material come from the fit. The velocities
are the clip's joint kinematics scaled by the subject's fitted per-joint factors. A motion without a recording
has no per-frame velocity correction to fit, so these renders carry the motion fidelity of a render without the
velocity fit.

\textbf{What the figures show.} Figs.~\ref{fig:supp_novel_rtpose_142}--\ref{fig:supp_novel_hupr_41} render
\rNovelNmotion{} motions in each of \rNovelNscene{} fitted scenes. These motions have no real recording, so we
cannot score them. We show them as a capability for data augmentation, not as a result.

\newcommand{\novelcaption}[1]{\textbf{Novel motions in the fitted scene of #1 (not scored).} Per motion: the
posed mesh at four evenly spaced frames, seen from 45$^\circ$ between the radar's view and the side, above the
rendered micro-Doppler over the whole clip (time across, radial velocity up) and the rendered range--Doppler and
range--azimuth at the clip's frame of largest pelvis speed (range across, over the person's range $\pm$2\,m).
Panels are linear magnitude normalized to their own range, as in Fig.~\ref{fig:qualitative}. Falls: forward and
backward generated by MDM~\citep{tevet2023mdm}, the fall from a pulled rug from CMU~\citep{mahmood2019amass}; exercises from
CMU.}

\begin{figure}[tbp]
  \centering
  \includegraphics[width=\linewidth]{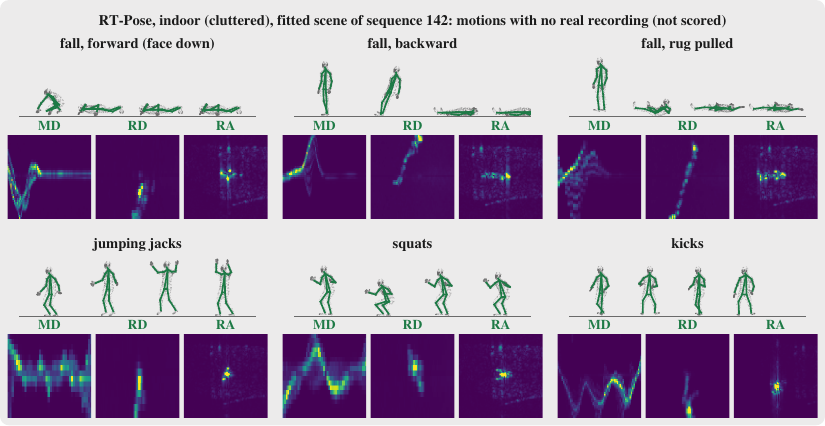}
  \caption{\novelcaption{RT-Pose sequence 142 (indoor, cluttered)}}
  \label{fig:supp_novel_rtpose_142}
\end{figure}

\begin{figure}[tbp]
  \centering
  \includegraphics[width=\linewidth]{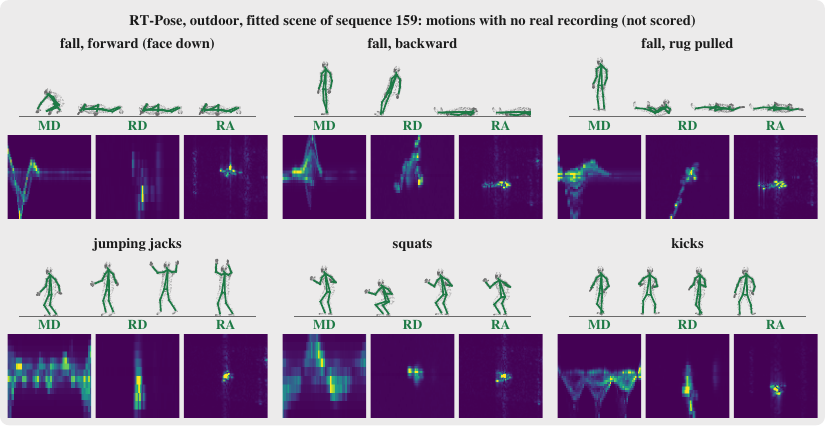}
  \caption{\novelcaption{RT-Pose sequence 159 (outdoor)}}
  \label{fig:supp_novel_rtpose_159}
\end{figure}

\begin{figure}[tbp]
  \centering
  \includegraphics[width=\linewidth]{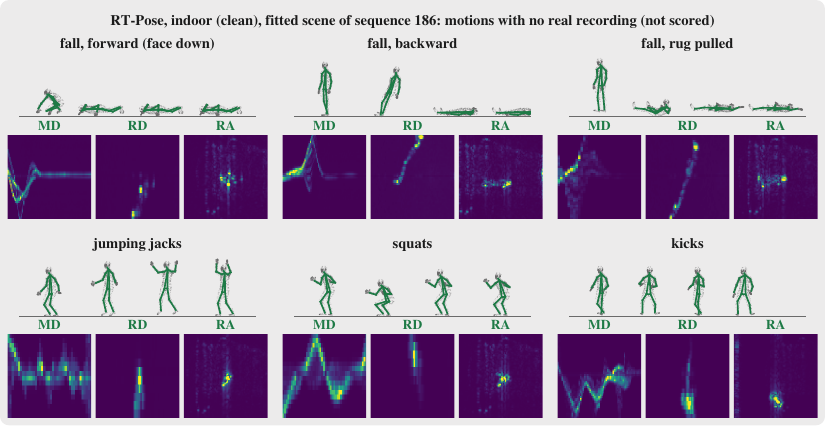}
  \caption{\novelcaption{RT-Pose sequence 186 (indoor, clean)}}
  \label{fig:supp_novel_rtpose_186}
\end{figure}

\begin{figure}[tbp]
  \centering
  \includegraphics[width=\linewidth]{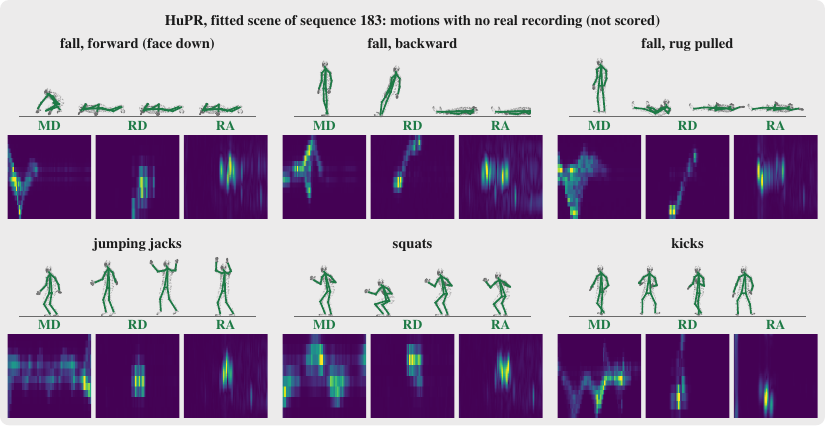}
  \caption{\novelcaption{HuPR sequence 183 (horizontal board)}}
  \label{fig:supp_novel_hupr_183}
\end{figure}

\begin{figure}[tbp]
  \centering
  \includegraphics[width=\linewidth]{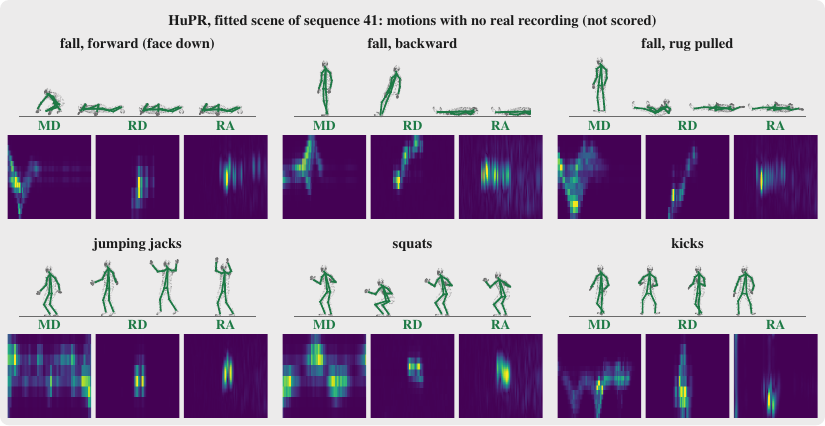}
  \caption{\novelcaption{HuPR sequence 41 (horizontal board)}}
  \label{fig:supp_novel_hupr_41}
\end{figure}
\clearpage

\clearpage
\section{Ablations, Runtime, and Peak Memory}
\label{sec:supp:ablations_runtime}

\subsection{Ablations}
\label{sec:supp:ablations}

\emph{Is every term of the objective necessary?} Four terms at unit weight
invite the question. We therefore remove each global term alone and the two windowed
terms together. The windowed pair shares one window and one scope, the
person's box, so we remove it as one unit. We expect the global loop-0 term to
matter most. The plane terms exclude DC and the windowed pair reaches only the
person's box, so the global loop-0 term is the only term that reaches the whole
static scene. Each variant multiplies the removed terms by
zero inside the loss and re-fits the renderer once per sequence, with
everything else at the default configuration (the same epochs, person inputs,
scaffolds and seed). The ``None'' row is not a re-fit. It is the
render Table~\ref{tab:fidelity} reports, and it reproduces
Table~\ref{tab:fidelity} and
Tables~\ref{tab:supp_fidelity_adc}--\ref{tab:supp_fidelity_rd_md} cell for
cell. We judge each variant twice. Reconstruction is scored over
Table~\ref{tab:fidelity}'s sequences (HuPR \rNseqHuPR{} per board, RT-Pose
\rNseqRTOne{} one-person). The gap to real, rendered-to-real minus
real-to-real paired per sequence, is read through the four real-trained seeds
that Table~\ref{tab:tstr} reports, and only on the test splits (HuPR
\rAblHuGapNNone{}, RT-Pose \rAblRTGapNNone{} sequences), because the
real-trained models are in-sample on every other sequence. We also count, per
sequence and radar, how often a variant scores worse or better than the
``None'' row by more than a threshold. The threshold is $0.01$ on bounded
scores (correlations, SSIM, clDice and the phase coherences), 2\% of the
``None'' value on the others, and 1\,cm or one AP point on the gap. We fit one
renderer seed, because sequences rather than seeds are the axis we generalize
over. Only the medians and the paired counts are therefore interpretable, never
a single sequence.

Every term carries part of the fit, but not the same part.
Table~\ref{tab:objective_ablation} summarizes the three variants, and
Tables~\ref{tab:supp_ablation_adc}--\ref{tab:supp_ablation_rd_md} report them
on every metric of
Tables~\ref{tab:supp_fidelity_adc}--\ref{tab:supp_fidelity_rd_md}. The global
loop-0 term carries the static scene, the global per-plane term carries the
noise floor and the Doppler structure, and the windowed pair carries a smaller
share on the person's cells. None of the three can be removed without a
measurable cost somewhere in that metric set.

\textbf{The global loop-0 term carries the static scene.} Without it, the
median of every magnitude column of every representation falls on every radar.
Range--azimuth $\rho$ falls from \rAblCRhoHuHNone{} to \rAblCRhoHuHNoRaeZero{}
on H and from \rAblCRhoRTNone{} to \rAblCRhoRTNoRaeZero{} on C, worse on
\rAblWRhoHuHNoRaeZero{} of \rAblNRhoHuHNoRaeZero{} and \rAblWRhoRTNoRaeZero{}
of \rAblNRhoRTNoRaeZero{} sequences. The CRP $\rho$ falls from
\rAblCCrpCorrHuHNone{} to \rAblCCrpCorrHuHNoRaeZero{} on H and from
\rAblCCrpCorrRTNone{} to \rAblCCrpCorrRTNoRaeZero{} on C, again worse on
\rAblWCrpCorrHuHNoRaeZero{} of \rAblNCrpCorrHuHNoRaeZero{} and
\rAblWCrpCorrRTNoRaeZero{} of \rAblNCrpCorrRTNoRaeZero{}. The columns scored
after static clutter removal do not move. RD $\rho_M$ reads
\rAblCRhoMRTNoRaeZero{} against \rAblCRhoMRTNone{} on C and micro-Doppler
$\rho_S$ \rAblCMdSRTNoRaeZero{} against \rAblCMdSRTNone{}, and the floor
ratios match the ``None'' row.
The two gaps differ. RT-Pose's gap grows from \rAblRTGapNone{} to
\rAblRTGapNoRaeZero{}\,cm, worse on \rAblRTGapWNoRaeZero{} of
\rAblRTGapNNoRaeZero{} test sequences, while HuPR's reads
\rAblHuGapNoRaeZero{} against \rAblHuGapNone{} AP (\rAblHuGapWNoRaeZero{}
worse, \rAblHuGapBNoRaeZero{} better). HuPR's released preprocessing removes
static clutter over loops before its model reads the heatmaps, so the static
scene this term fits never reaches the HuPR model, while HRRadarPose reads it.

\textbf{The global per-plane term carries the noise floor and the Doppler
structure.} Without it, the floor level $L_U$ falls on every radar, from
\rAblCLUHuHNone{} to \rAblCLUHuHNoRaePl{} on H (worse on \rAblWLUHuHNoRaePl{}
of \rAblNLUHuHNoRaePl{}), from \rAblCLUHuVNone{} to \rAblCLUHuVNoRaePl{} on V
and from \rAblCLURTNone{} to \rAblCLURTNoRaePl{} on C. The floor spread
$\sigma_U$ on C moves away from 1, from \rAblCSigmaURTNone{} to
\rAblCSigmaURTNoRaePl{}.
RD $\rho_M$ falls on every radar, worse on \rAblWRhoMHuHNoRaePl{} of
\rAblNRhoMHuHNoRaePl{} sequences on H and \rAblWRhoMRTNoRaePl{} of
\rAblNRhoMRTNoRaePl{} on C, and RT-Pose's micro-Doppler $\rho_S$ falls from
\rAblCMdSRTNone{} to \rAblCMdSRTNoRaePl{} (worse on \rAblWMdSRTNoRaePl{} of
\rAblNMdSRTNoRaePl{}). SSIM on the CRP and range--azimuth maps also falls on
both HuPR boards, worse on \rAblWCrpSsimHuHNoRaePl{} and
\rAblWSsimHuHNoRaePl{} of \rAblNSsimHuHNoRaePl{} sequences on H. The geometry
does not move. Range--azimuth $\rho$ reads \rAblCRhoRTNoRaePl{} on C, and
neither gap separates from the ``None'' row (HuPR \rAblHuGapWNoRaePl{} worse
and \rAblHuGapBNoRaePl{} better, RT-Pose \rAblRTGapWNoRaePl{} and
\rAblRTGapBNoRaePl{}). The term therefore fits the floor and the Doppler
planes, not the geometry.

\textbf{The windowed pair has the smallest effect, on the person's cells in
the scene.} Without it, RT-Pose's clutter-removed range--azimuth $\rho_c$ falls
from \rAblCRhoCRTNone{} to \rAblCRhoCRTNoWin{}, worse on \rAblWRhoCRTNoWin{}
of \rAblNRhoCRTNoWin{} sequences, and its RAE $\rho$ from
\rAblCRaeRhoRTNone{} to \rAblCRaeRhoRTNoWin{}. HuPR's gap widens from
\rAblHuGapNone{} to \rAblHuGapNoWin{} AP (\rAblHuGapWNoWin{} test sequences
worse, \rAblHuGapBNoWin{} better), while its reconstruction stays at the
``None'' row, with range--azimuth $\rho$ at \rAblCRhoHuHNoWin{} on H.
RT-Pose's gap does not separate (\rAblRTGapWNoWin{} worse,
\rAblRTGapBNoWin{} better). The effect is the smallest of the three, and it
rests on paired counts from one renderer seed.

The phase-coherence columns ($\Delta\phi_t$, $\Delta\phi_r$, $\Delta\phi_v$)
do not separate the variants in a consistent direction, so we do not read them
as evidence for or against any term.

\begin{table}[htbp]
\centering
\caption{\textbf{Objective ablation, summary.} Each row removes the named
terms of Eq.~\eqref{eq:objective} and re-fits the renderer, with everything
else at the default configuration (setup in App.~\ref{sec:supp:ablations}). ``None''
is the render Table~\ref{tab:fidelity} reports, not a re-fit. $\rho$ is the
median range--azimuth correlation and $\rho_S$ the median micro-Doppler
spectrogram correlation, over Table~\ref{tab:fidelity}'s sequences (HuPR both
boards pooled). $\Delta$ is the median per-sequence gap to real on the test
splits, read through the real-trained seeds of Table~\ref{tab:tstr}. On HuPR
it is real-to-real minus rendered-to-real AP, and on RT-Pose rendered-to-real
minus real-to-real MPJPE and MRPE in cm, so a larger $\Delta$ is a worse
render. W/B
counts the test sequences whose $\Delta$ is worse / better than the ``None''
row's by more than one AP point or 1\,cm (MPJPE). One renderer seed.
Tables~\ref{tab:supp_ablation_adc}--\ref{tab:supp_ablation_rd_md} give every
metric per radar.}
\label{tab:objective_ablation}
\small
\setlength{\tabcolsep}{3.5pt}
\begin{adjustbox}{max width=\textwidth}
\begin{tabular}{lcccccccccc}
\toprule
& \multicolumn{4}{c}{\textbf{HuPR}} & \multicolumn{5}{c}{\textbf{RT-Pose}} \\
\cmidrule(lr){2-5}\cmidrule(lr){6-10}
\textbf{Term removed} & $\rho\uparrow$ & $\rho_S\uparrow$ & $\Delta$ (AP)$\downarrow$ & W/B
                      & $\rho\uparrow$ & $\rho_S\uparrow$ & $\Delta$ MPJPE$\downarrow$ & $\Delta$ MRPE$\downarrow$ & W/B \\
\midrule
None (full objective) & \rAblHuRhoNone{} & \rAblHuRhoSNone{} & \rAblHuGapNone{} & -- & \rAblRTRhoNone{} & \rAblRTRhoSNone{} & \rAblRTGapNone{} & \rAblRTGapMRPENone{} & -- \\
\midrule
$-\mathcal{L}_{\mathrm{RAE},0}$ (global, loop 0) & \rAblHuRhoNoRaeZero{} & \rAblHuRhoSNoRaeZero{} & \rAblHuGapNoRaeZero{} & \rAblHuGapWNoRaeZero{}/\rAblHuGapBNoRaeZero{} & \rAblRTRhoNoRaeZero{} & \rAblRTRhoSNoRaeZero{} & \rAblRTGapNoRaeZero{} & \rAblRTGapMRPENoRaeZero{} & \rAblRTGapWNoRaeZero{}/\rAblRTGapBNoRaeZero{} \\
$-\mathcal{L}_{\mathrm{RAE},\mathrm{pl}}$ (global, per plane) & \rAblHuRhoNoRaePl{} & \rAblHuRhoSNoRaePl{} & \rAblHuGapNoRaePl{} & \rAblHuGapWNoRaePl{}/\rAblHuGapBNoRaePl{} & \rAblRTRhoNoRaePl{} & \rAblRTRhoSNoRaePl{} & \rAblRTGapNoRaePl{} & \rAblRTGapMRPENoRaePl{} & \rAblRTGapWNoRaePl{}/\rAblRTGapBNoRaePl{} \\
$-(\mathcal{L}^{\mathrm{win}}_{\mathrm{RAE},0}+\mathcal{L}^{\mathrm{win}}_{\mathrm{RAE},\mathrm{pl}})$ (windowed) & \rAblHuRhoNoWin{} & \rAblHuRhoSNoWin{} & \rAblHuGapNoWin{} & \rAblHuGapWNoWin{}/\rAblHuGapBNoWin{} & \rAblRTRhoNoWin{} & \rAblRTRhoSNoWin{} & \rAblRTGapNoWin{} & \rAblRTGapMRPENoWin{} & \rAblRTGapWNoWin{}/\rAblRTGapBNoWin{} \\
\bottomrule
\end{tabular}
\end{adjustbox}
\end{table}

\begin{table*}[htbp]
\centering
\caption{\textbf{Objective ablation on the ADC samples.} Columns as
Table~\ref{tab:supp_fidelity_adc} ($\rho_{\mathrm{env}}$ the envelope
correlation after the rank-1 phase correction, $\Delta\phi_t$ the fast-time
relative-phase coherence with its weighted RMSE in degrees); rows as
Table~\ref{tab:objective_ablation}, over the same sequences (HuPR
\rNseqHuPR{} per board, RT-Pose \rNseqRTOne{} one-person). Bold: the best value
in each column of a row section, by the column's arrow.}
\label{tab:supp_ablation_adc}
\footnotesize
\setlength{\tabcolsep}{5pt}
\begin{tabular}{lcccccc}
\toprule
\textbf{Term removed} & $\rho_{\mathrm{env}}\uparrow$ & PSNR$\uparrow$ & SSIM$\uparrow$ & RMSE$\downarrow$ & $\Delta\phi_t\uparrow$ & $\Delta\phi_t$ ($^\circ$)$\downarrow$ \\
\midrule
\multicolumn{7}{l}{\emph{HuPR}, horizontal single-chip radar (H)} \\
None (full objective) & \rAblAdcHuHNone{} \\
$-\mathcal{L}_{\mathrm{RAE},0}$ & \rAblAdcHuHNoRaeZero{} \\
$-\mathcal{L}_{\mathrm{RAE},\mathrm{pl}}$ & \rAblAdcHuHNoRaePl{} \\
$-(\mathcal{L}^{\mathrm{win}}_{\mathrm{RAE},0}+\mathcal{L}^{\mathrm{win}}_{\mathrm{RAE},\mathrm{pl}})$ & \rAblAdcHuHNoWin{} \\
\midrule
\multicolumn{7}{l}{\emph{HuPR}, vertical single-chip radar (V)} \\
None (full objective) & \rAblAdcHuVNone{} \\
$-\mathcal{L}_{\mathrm{RAE},0}$ & \rAblAdcHuVNoRaeZero{} \\
$-\mathcal{L}_{\mathrm{RAE},\mathrm{pl}}$ & \rAblAdcHuVNoRaePl{} \\
$-(\mathcal{L}^{\mathrm{win}}_{\mathrm{RAE},0}+\mathcal{L}^{\mathrm{win}}_{\mathrm{RAE},\mathrm{pl}})$ & \rAblAdcHuVNoWin{} \\
\midrule
\multicolumn{7}{l}{\emph{RT-Pose}, cascade (C), one-person sequences} \\
None (full objective) & \rAblAdcRTNone{} \\
$-\mathcal{L}_{\mathrm{RAE},0}$ & \rAblAdcRTNoRaeZero{} \\
$-\mathcal{L}_{\mathrm{RAE},\mathrm{pl}}$ & \rAblAdcRTNoRaePl{} \\
$-(\mathcal{L}^{\mathrm{win}}_{\mathrm{RAE},0}+\mathcal{L}^{\mathrm{win}}_{\mathrm{RAE},\mathrm{pl}})$ & \rAblAdcRTNoWin{} \\
\bottomrule
\end{tabular}
\end{table*}

\begin{table*}[htbp]
\centering
\caption{\textbf{Objective ablation on the complex range profile (CRP).}
Columns as Table~\ref{tab:supp_fidelity_crp} ($\Delta\phi_r$ and
$\Delta\phi_v$ the relative-phase coherences along range and virtual channel,
each with its weighted RMSE in degrees); rows, sequences and bold as
Table~\ref{tab:supp_ablation_adc}.}
\label{tab:supp_ablation_crp}
\footnotesize
\setlength{\tabcolsep}{5pt}
\begin{tabular}{lcccccccc}
\toprule
\textbf{Term removed} & $\rho\uparrow$ & PSNR$\uparrow$ & SSIM$\uparrow$ & RMSE$\downarrow$ & $\Delta\phi_r\uparrow$ & $\Delta\phi_r$ ($^\circ$)$\downarrow$ & $\Delta\phi_v\uparrow$ & $\Delta\phi_v$ ($^\circ$)$\downarrow$ \\
\midrule
\multicolumn{9}{l}{\emph{HuPR}, horizontal single-chip radar (H)} \\
None (full objective) & \rAblCrpHuHNone{} \\
$-\mathcal{L}_{\mathrm{RAE},0}$ & \rAblCrpHuHNoRaeZero{} \\
$-\mathcal{L}_{\mathrm{RAE},\mathrm{pl}}$ & \rAblCrpHuHNoRaePl{} \\
$-(\mathcal{L}^{\mathrm{win}}_{\mathrm{RAE},0}+\mathcal{L}^{\mathrm{win}}_{\mathrm{RAE},\mathrm{pl}})$ & \rAblCrpHuHNoWin{} \\
\midrule
\multicolumn{9}{l}{\emph{HuPR}, vertical single-chip radar (V)} \\
None (full objective) & \rAblCrpHuVNone{} \\
$-\mathcal{L}_{\mathrm{RAE},0}$ & \rAblCrpHuVNoRaeZero{} \\
$-\mathcal{L}_{\mathrm{RAE},\mathrm{pl}}$ & \rAblCrpHuVNoRaePl{} \\
$-(\mathcal{L}^{\mathrm{win}}_{\mathrm{RAE},0}+\mathcal{L}^{\mathrm{win}}_{\mathrm{RAE},\mathrm{pl}})$ & \rAblCrpHuVNoWin{} \\
\midrule
\multicolumn{9}{l}{\emph{RT-Pose}, cascade (C), one-person sequences} \\
None (full objective) & \rAblCrpRTNone{} \\
$-\mathcal{L}_{\mathrm{RAE},0}$ & \rAblCrpRTNoRaeZero{} \\
$-\mathcal{L}_{\mathrm{RAE},\mathrm{pl}}$ & \rAblCrpRTNoRaePl{} \\
$-(\mathcal{L}^{\mathrm{win}}_{\mathrm{RAE},0}+\mathcal{L}^{\mathrm{win}}_{\mathrm{RAE},\mathrm{pl}})$ & \rAblCrpRTNoWin{} \\
\bottomrule
\end{tabular}
\end{table*}

\begin{table*}[htbp]
\centering
\caption{\textbf{Objective ablation on range--angle and
range--azimuth--elevation (RAE).} Columns as
Table~\ref{tab:supp_fidelity_ra_rae}; rows, sequences and bold as
Table~\ref{tab:supp_ablation_adc}. Every row is a 4DPS render, which carries
an absolute scale, so SSIM, NRMSE and PSNR are scored directly, without the
level matching that RF-Genesis needs.}
\label{tab:supp_ablation_ra_rae}
\footnotesize
\setlength{\tabcolsep}{4pt}
\begin{adjustbox}{max width=\textwidth}
\begin{tabular}{lccccccccc}
\toprule
& \multicolumn{5}{c}{\textbf{Range--angle}} & \multicolumn{4}{c}{\textbf{RAE}} \\
\cmidrule(lr){2-6}\cmidrule(lr){7-10}
\textbf{Term removed} & $\rho\uparrow$ & $\rho_c\uparrow$ & SSIM$\uparrow$ & NRMSE$\downarrow$ & PSNR$\uparrow$ & $\rho\uparrow$ & SSIM$\uparrow$ & NRMSE$\downarrow$ & PSNR$\uparrow$ \\
\midrule
\multicolumn{10}{l}{\emph{HuPR}, horizontal single-chip radar (H)} \\
None (full objective) & \rAblRaRaeHuHNone{} \\
$-\mathcal{L}_{\mathrm{RAE},0}$ & \rAblRaRaeHuHNoRaeZero{} \\
$-\mathcal{L}_{\mathrm{RAE},\mathrm{pl}}$ & \rAblRaRaeHuHNoRaePl{} \\
$-(\mathcal{L}^{\mathrm{win}}_{\mathrm{RAE},0}+\mathcal{L}^{\mathrm{win}}_{\mathrm{RAE},\mathrm{pl}})$ & \rAblRaRaeHuHNoWin{} \\
\midrule
\multicolumn{10}{l}{\emph{HuPR}, vertical single-chip radar (V)} \\
None (full objective) & \rAblRaRaeHuVNone{} \\
$-\mathcal{L}_{\mathrm{RAE},0}$ & \rAblRaRaeHuVNoRaeZero{} \\
$-\mathcal{L}_{\mathrm{RAE},\mathrm{pl}}$ & \rAblRaRaeHuVNoRaePl{} \\
$-(\mathcal{L}^{\mathrm{win}}_{\mathrm{RAE},0}+\mathcal{L}^{\mathrm{win}}_{\mathrm{RAE},\mathrm{pl}})$ & \rAblRaRaeHuVNoWin{} \\
\midrule
\multicolumn{10}{l}{\emph{RT-Pose}, cascade (C), one-person sequences} \\
None (full objective) & \rAblRaRaeRTNone{} \\
$-\mathcal{L}_{\mathrm{RAE},0}$ & \rAblRaRaeRTNoRaeZero{} \\
$-\mathcal{L}_{\mathrm{RAE},\mathrm{pl}}$ & \rAblRaRaeRTNoRaePl{} \\
$-(\mathcal{L}^{\mathrm{win}}_{\mathrm{RAE},0}+\mathcal{L}^{\mathrm{win}}_{\mathrm{RAE},\mathrm{pl}})$ & \rAblRaRaeRTNoWin{} \\
\bottomrule
\end{tabular}
\end{adjustbox}
\end{table*}

\begin{table*}[htbp]
\centering
\caption{\textbf{Objective ablation on range--Doppler (RD) and
micro-Doppler.} Columns as Table~\ref{tab:supp_fidelity_rd_md} ($P_M$ the
rendered-over-real power ratio on the CFAR-occupied cells, $L_U$ and
$\sigma_U$ the level and spread ratios on the unoccupied ones,
$\rho_{\mathrm{env}}$ the Doppler-envelope correlation, BW the envelope
separation, clDice the connectivity of the segmented signatures); rows,
sequences and bold as Table~\ref{tab:supp_ablation_adc}. Every row is a 4DPS
render, which carries an absolute scale, so $P_M$, $L_U$ and $\sigma_U$ are
scored directly, without the level matching that RF-Genesis needs.}
\label{tab:supp_ablation_rd_md}
\footnotesize
\setlength{\tabcolsep}{4pt}
\begin{adjustbox}{max width=\textwidth}
\begin{tabular}{lcccccccccc}
\toprule
& \multicolumn{2}{c}{\textbf{RD occupied}} & \multicolumn{2}{c}{\textbf{RD floor}} & \multicolumn{6}{c}{\textbf{Micro-Doppler}} \\
\cmidrule(lr){2-3}\cmidrule(lr){4-5}\cmidrule(lr){6-11}
\textbf{Term removed} & $\rho_M\uparrow$ & $P_M\to 1$ & $L_U\to 1$ & $\sigma_U\to 1$ & $\rho_\mu\uparrow$ & $\rho_\sigma\uparrow$ & $\rho_S\uparrow$ & $\rho_{\mathrm{env}}\uparrow$ & BW$\to 1$ & clDice$\uparrow$ \\
\midrule
\multicolumn{11}{l}{\emph{HuPR}, horizontal single-chip radar (H)} \\
None (full objective) & \rAblRdMdHuHNone{} \\
$-\mathcal{L}_{\mathrm{RAE},0}$ & \rAblRdMdHuHNoRaeZero{} \\
$-\mathcal{L}_{\mathrm{RAE},\mathrm{pl}}$ & \rAblRdMdHuHNoRaePl{} \\
$-(\mathcal{L}^{\mathrm{win}}_{\mathrm{RAE},0}+\mathcal{L}^{\mathrm{win}}_{\mathrm{RAE},\mathrm{pl}})$ & \rAblRdMdHuHNoWin{} \\
\midrule
\multicolumn{11}{l}{\emph{HuPR}, vertical single-chip radar (V)} \\
None (full objective) & \rAblRdMdHuVNone{} \\
$-\mathcal{L}_{\mathrm{RAE},0}$ & \rAblRdMdHuVNoRaeZero{} \\
$-\mathcal{L}_{\mathrm{RAE},\mathrm{pl}}$ & \rAblRdMdHuVNoRaePl{} \\
$-(\mathcal{L}^{\mathrm{win}}_{\mathrm{RAE},0}+\mathcal{L}^{\mathrm{win}}_{\mathrm{RAE},\mathrm{pl}})$ & \rAblRdMdHuVNoWin{} \\
\midrule
\multicolumn{11}{l}{\emph{RT-Pose}, cascade (C), one-person sequences} \\
None (full objective) & \rAblRdMdRTNone{} \\
$-\mathcal{L}_{\mathrm{RAE},0}$ & \rAblRdMdRTNoRaeZero{} \\
$-\mathcal{L}_{\mathrm{RAE},\mathrm{pl}}$ & \rAblRdMdRTNoRaePl{} \\
$-(\mathcal{L}^{\mathrm{win}}_{\mathrm{RAE},0}+\mathcal{L}^{\mathrm{win}}_{\mathrm{RAE},\mathrm{pl}})$ & \rAblRdMdRTNoWin{} \\
\bottomrule
\end{tabular}
\end{adjustbox}
\end{table*}

\subsection{Runtime and Peak Memory}
\label{sec:supp:runtime}

This subsection reports what the two experiments cost in time and in GPU
memory: reconstructing every sequence (Sec.~\ref{sec:exp:fidelity}) with
4DPS, its unfit render and RF-Genesis, and training the downstream models
(Sec.~\ref{sec:exp:tstr}).
Every timing in Tables~\ref{tab:runtime}
and~\ref{tab:train} is a wall time from the run that produced the paper's
numbers, one sequence or one training per GPU, for 4DPS and for RF-Genesis
alike. The 4DPS fits, the unfit renders and every
downstream training, including those on RF-Genesis's renders, ran on NVIDIA
H100 GPUs.
A 4DPS fit's time runs from the person stage through the 400-epoch velocity
and material fit to the render and its evaluation, and excludes the scene
preprocessing it inherits (App.~\ref{sec:supp:preproc}). The unfit render
runs the same stages with no fit. RF-Genesis's time covers its mesh export,
its ray-cast simulation, the representations computed from the simulated
cube and their scoring against the recording.

\begin{table}[htbp]
\centering
\caption{\textbf{Reconstruction cost} (Sec.~\ref{sec:exp:fidelity}): the
median wall time per sequence and the total over the corpus. RT-Pose's unfit
render covers the \rRtUnfitNRT{} split sequences it is scored on. The 4DPS
step is the median over sequences of the training step's median.
RF-Genesis ran on RTX~4090s and 4DPS on H100s, so the two methods' times
are not a like-for-like speed comparison.}
\label{tab:runtime}
\small
\setlength{\tabcolsep}{4pt}
\begin{tabular}{llccccc}
\toprule
\textbf{Data} & \textbf{Method} & GPU & Sequences & Per sequence (min) & Total (GPU-h) & Step (ms) \\
\midrule
HuPR    & 4DPS          & H100     & \rRtFitNHu{}   & \rRtFitMinHu{}   & \rRtFitGPUhHu{}   & \rRtStepMsHu{} \\
        & 4DPS (unfit)  & H100     & \rRtUnfitNHu{} & \rRtUnfitMinHu{} & \rRtUnfitGPUhHu{} & -- \\
        & RF-Genesis    & RTX 4090 & \rRtRfgNHu{}   & \rRtRfgMinHu{}   & \rRtRfgGPUhHu{}   & -- \\
\midrule
RT-Pose & 4DPS          & H100     & \rRtFitNRT{}   & \rRtFitMinRT{}   & \rRtFitGPUhRT{}   & \rRtStepMsRT{} \\
        & 4DPS (unfit)  & H100     & \rRtUnfitNRT{} & \rRtUnfitMinRT{} & \rRtUnfitGPUhRT{} & -- \\
        & RF-Genesis    & RTX 4090 & \rRtRfgNRT{}   & \rRtRfgMinRT{}   & \rRtRfgGPUhRT{}   & -- \\
\bottomrule
\end{tabular}
\end{table}

\textbf{Downstream training} (Sec.~\ref{sec:exp:tstr}).
Table~\ref{tab:train} gives the cost of one training per H100 at the
protocol's fixed schedule, for every arm of Table~\ref{tab:tstr}. The real
arm is shared: the RF-Genesis arms were compared against the same four real
trainings, which were not retrained. A training's time is the sum over
epochs of its training time, plus the per-epoch validation on HuPR.

\begin{table}[htbp]
\centering
\caption{\textbf{Downstream training cost} (Sec.~\ref{sec:exp:tstr}): hours
per training, mean (std) over the four seeds, and the total over the four
seeds. Every training ran on one H100. Evaluating Table~\ref{tab:tstr}'s
five cells over the four seeds adds \rRtEvalGPUhHu{} (4DPS) and
\rRtEvalRfgGPUhHu{} (RF-Genesis) GPU-h on HuPR, and \rRtEvalGPUhRT{} and
\rRtEvalRfgGPUhRT{} on RT-Pose.}
\label{tab:train}
\small
\setlength{\tabcolsep}{4pt}
\begin{tabular}{llccccccc}
\toprule
& & & \multicolumn{3}{c}{\textbf{HuPR}} & \multicolumn{3}{c}{\textbf{RT-Pose}} \\
\cmidrule(lr){4-6}\cmidrule(lr){7-9}
\textbf{Arm} & \textbf{Renderer} & GPU & h / training & Epochs & GPU-h & h / training & Epochs & GPU-h \\
\midrule
Real     & none (real)  & H100 & \rRtTrainRealHu{}        & \rRtTrainEpochsHu{} & \rRtTrainGPUhRealHu{}        & \rRtTrainRealRT{}        & \rRtTrainEpochsRT{} & \rRtTrainGPUhRealRT{} \\
Rendered & 4DPS         & H100 & \rRtTrainRenderedHu{}    & \rRtTrainEpochsHu{} & \rRtTrainGPUhRenderedHu{}    & \rRtTrainRenderedRT{}    & \rRtTrainEpochsRT{} & \rRtTrainGPUhRenderedRT{} \\
Mixed    & 4DPS         & H100 & \rRtTrainMixedHu{}       & \rRtTrainEpochsHu{} & \rRtTrainGPUhMixedHu{}       & \rRtTrainMixedRT{}       & \rRtTrainEpochsRT{} & \rRtTrainGPUhMixedRT{} \\
Rendered & RF-Genesis   & H100 & \rRtTrainRfgRenderedHu{} & \rRtTrainEpochsHu{} & \rRtTrainGPUhRfgRenderedHu{} & \rRtTrainRfgRenderedRT{} & \rRtTrainEpochsRT{} & \rRtTrainGPUhRfgRenderedRT{} \\
Mixed    & RF-Genesis   & H100 & \rRtTrainRfgMixedHu{}    & \rRtTrainEpochsHu{} & \rRtTrainGPUhRfgMixedHu{}    & \rRtTrainRfgMixedRT{}    & \rRtTrainEpochsRT{} & \rRtTrainGPUhRfgMixedRT{} \\
\bottomrule
\end{tabular}
\end{table}

\textbf{Peak memory.} Table~\ref{tab:memory} gives the peak GPU memory of
each stage, and its caption says which counter each cell reads. The largest
4DPS fit allocates \rRtPeakFitMaxRT{}\,GiB at its peak.

\begin{table}[htbp]
\centering
\caption{\textbf{Peak GPU memory} (GiB): the median and the maximum over
sequences for reconstruction, and the maximum over the four seeds for
training. Every 4DPS cell and every training cell is PyTorch's allocator peak
(\texttt{max\_memory\_allocated}), as the 4DPS runs record it per sequence and
RT-Pose's trainer logs it; the reserved pool is larger. The RF-Genesis
reconstruction cells are instead the card-level peak from nvidia-smi, because
its ray tracer allocates memory outside PyTorch (PyTorch alone records
\rRtPeakRfgTorchHu{} and \rRtPeakRfgTorchRT{}\,GiB), so they are not the same
quantity as the 4DPS cells. The cells marked $^\dagger$ were measured
separately from the runs that produced the paper's numbers, with the same
code, protocol and data: RF-Genesis as the maximum over the shortest, median
and longest sequence of each dataset on an RTX~4090, and HuPR training over
\rRtPeakTrainStepsHu{} steps and one validation pass per arm on an H100; the
\rRtPeakTrainArmsHu{} HuPR arms are identical (\rRtPeakTrainSmiHu{}\,GiB at
card level).}
\label{tab:memory}
\small
\setlength{\tabcolsep}{4pt}
\begin{tabular}{lllcccc}
\toprule
& & & \multicolumn{2}{c}{\textbf{HuPR}} & \multicolumn{2}{c}{\textbf{RT-Pose}} \\
\cmidrule(lr){4-5}\cmidrule(lr){6-7}
\textbf{Stage} & \textbf{Method / arm} & GPU & median & max & median & max \\
\midrule
Reconstruction & 4DPS         & H100 & \rRtPeakFitMedHu{}   & \rRtPeakFitMaxHu{}   & \rRtPeakFitMedRT{}   & \rRtPeakFitMaxRT{} \\
(Sec.~\ref{sec:exp:fidelity}) & RF-Genesis   & RTX 4090 & -- & \rRtPeakRfgMaxHu{}$^\dagger$ & -- & \rRtPeakRfgMaxRT{}$^\dagger$ \\
\midrule
Training       & real                & H100 & -- & \rRtPeakTrainHu{}$^\dagger$ & -- & \rRtPeakTrainRealRT{} \\
(Sec.~\ref{sec:exp:tstr}) & 4DPS rendered & H100 & -- & \rRtPeakTrainHu{}$^\dagger$ & -- & \rRtPeakTrainRenderedRT{} \\
               & 4DPS mixed          & H100 & -- & \rRtPeakTrainHu{}$^\dagger$ & -- & \rRtPeakTrainMixedRT{} \\
               & RF-Genesis rendered & H100 & -- & \rRtPeakTrainHu{}$^\dagger$ & -- & \rRtPeakTrainRfgRenderedRT{} \\
               & RF-Genesis mixed    & H100 & -- & \rRtPeakTrainHu{}$^\dagger$ & -- & \rRtPeakTrainRfgMixedRT{} \\
\bottomrule
\end{tabular}
\end{table}

\textbf{The training step.}
\label{sec:supp:runtime_gates}
Three decisions make a corpus-scale fit affordable. First, the step has no
data-dependent shape and no host synchronization: the path count is fixed per
step and the loss chain and frames are batched, so the whole training step is
captured once as a CUDA graph and replayed. Second, the forward model runs in
fused CUDA kernels with hand-written backward passes, and the Doppler-plane
amplitudes are evaluated in closed form (Eq.~\eqref{eq:doppler_closed_form}). Third, everything in Eq.~\eqref{eq:doppler_splat} that does not
read a normal or a material is cached per frame, for the scene and for the
person's paths, and the person's paths are sorted once per frame by their
target cell, so each accumulator cell is one contiguous run summed without
atomics.

\clearpage
\section{Limitations}
\label{sec:supp:limitations}

\textbf{Single-bounce transport, and one mover.} As in prior radar
splatting~\cite{3dps, kung2025radarsplat}, we render the single-bounce path
from each transmitter to a point and back, so the returns that reach a receiver
only after reflecting off a second surface are absent from every product. Only
the subject moves: the scene is fitted once per sequence and held static, and
one person is modeled, because RT-Pose annotates the pose of only one person
in its multi-person sequences. Its \rNseqRTTwo{} two-person sequences are
therefore fitted against the annotated person alone, and the paper reports
results on the one-person sequences.

\textbf{Motion and labels come from the video's body model.} Each point's
velocity stays within a bounded residual of the kinematics the monocular body
model supplies, and the skeleton comes from that model, never from a radar
label. The fit therefore corrects the video's motion rather than recovering
it independently: a limb the body model misplaces beyond the bound is a limb
the radar term cannot invent, and the bound is what keeps the Doppler fit from
absorbing scaffold error as spurious velocity. RT-Pose's labels are derived on
every radar frame from the same registered mesh that drives the render, with
the released-label result reported beside them (App.~\ref{sec:supp:downstream}).

\textbf{Elevation is the weakest axis.} The scaffold reconstructs elevation
least well of the three spatial axes, because it is built from a monocular
depth or LiDAR scene and a body model rather than measured by the radar. The
products that resolve elevation are therefore the harder ones, which is why
HuPR's vertical board and the RT-Pose cascade score below HuPR's horizontal
board throughout Sec.~\ref{sec:exp:fidelity}. Two of HuPR's encoder taps are
also undefined in the representation analysis, because the temporal merges
feeding them are denormal in some trained checkpoints
(App.~\ref{sec:supp:repr}).

\textbf{No fitted baseline exists to compare against.} No public radar
simulator takes a capture as input, so RF-Genesis is driven forward from the
registered motion rather than fitted to the recording we score against. The
comparison that carries Sec.~\ref{sec:exp:fidelity} is therefore against our
own unfit render, which isolates what the radar fit adds over the camera but
cannot show what a competing fitted renderer would achieve. The survey in
App.~\ref{sec:supp:baselines} records that this is a property of the field
rather than of our selection.

\textbf{What the equivalence measurements cannot settle.} An equivalence claim
needs a control that detects a difference of a known size, and the only one we
trained is the untrained network. It separates cleanly at the shallowest taps
and barely at all at the backbone output and the head, so the pre-registered
margin means less the deeper the tap, and we read the two deepest taps as
degenerate rather than as passes. We also did not train a second real arm per
seed from a shared initialization, so the same-initialization row is read
against the different-initialization real--real mean rather than against a
reference of its own (App.~\ref{sec:supp:repr}).

\textbf{Rendered features transfer better on one dataset than the other.}
Freezing each backbone and fitting a fresh linear head on real data recovers
most of the rendered arm's deficit on RT-Pose and about half of it on HuPR
(Sec.~\ref{sec:exp:repr}). We report the asymmetry rather than explain it:
the two datasets differ in radar geometry, in downstream architecture and in
label provenance at once, so this evaluation cannot attribute the difference
to any one of them.

\textbf{Most edits are not scored.} Fitting an explicit scene is what makes
translation, retiming and recomposition possible. One edit, rendering a fitted
person in another fitted scene of the same environment, is evaluated in
App.~\ref{sec:supp:editing:separability}; translation, retiming and new
motions are not scored, because no real recording exists to compare them
against. A corpus of edited sequences is the next step.

\section{Additional Qualitative Results}
\label{sec:supp:qualitative}

\begin{figure}[p]
  \centering
  \includegraphics[width=\linewidth]{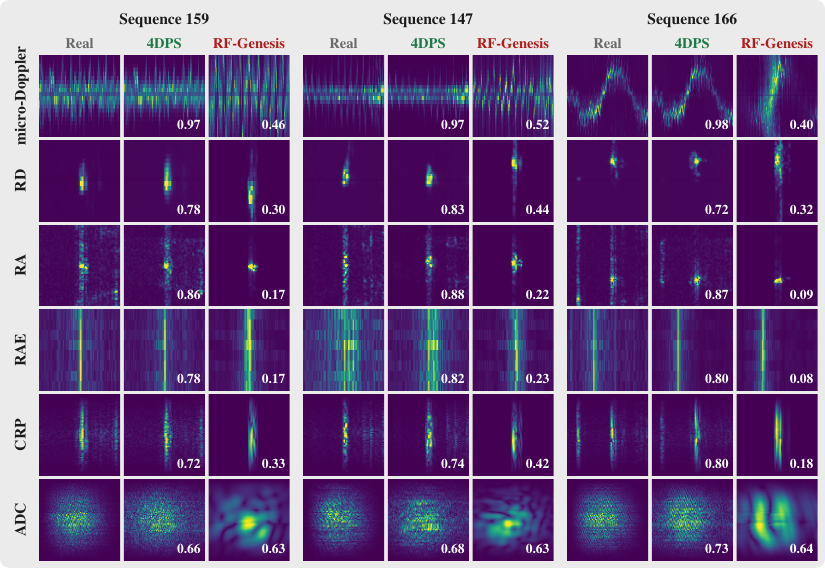}
  \\[4pt]
  \includegraphics[width=\linewidth]{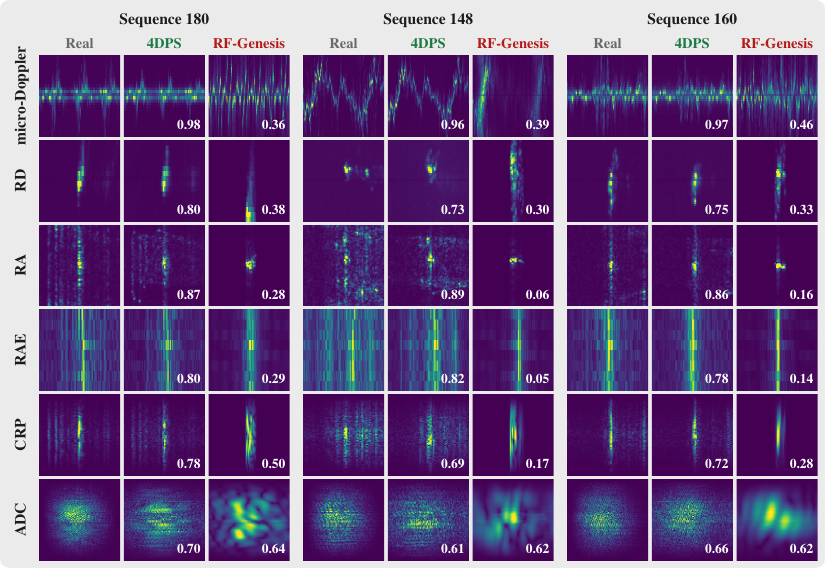}
  \caption{\textbf{Qualitative gallery, RT-Pose (1 of 3): sequences 159, 147 and 166 (top) and 180, 148 and 160 (bottom)} (4DPS-fidelity ranks 8, 1, 16, 4, 9 and 13 of 63). The top three are the sequences of Fig.~\ref{fig:qualitative}. Selection rule, rows, columns, frames and scores are given at the start of App.~\ref{sec:supp:qualitative}.}
  \label{fig:supp_gallery_rtpose_1}
\end{figure}

\begin{figure}[p]
  \centering
  \includegraphics[width=\linewidth]{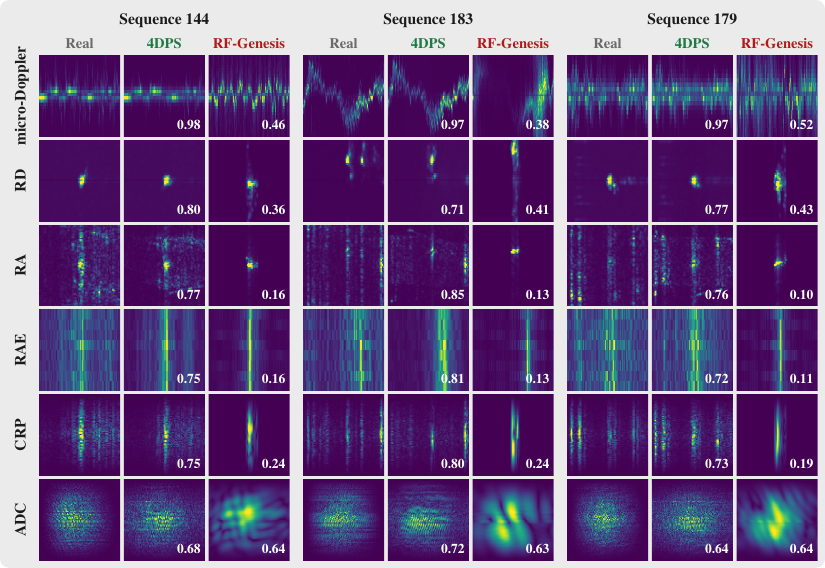}
  \\[4pt]
  \includegraphics[width=\linewidth]{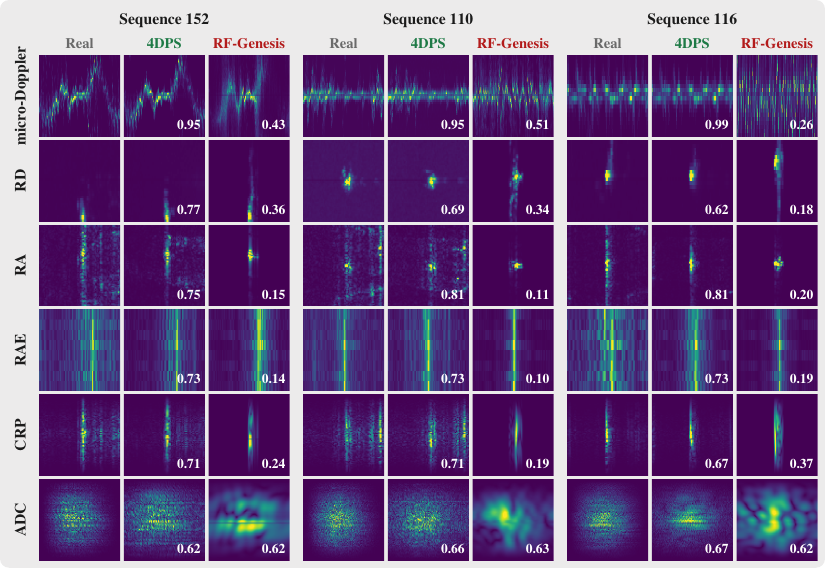}
  \caption{\textbf{Qualitative gallery, RT-Pose (2 of 3): sequences 144, 183 and 179 (top) and 152, 110 and 116 (bottom)} (4DPS-fidelity ranks 18, 22, 26, 30, 34 and 38 of 63). As Fig.~\ref{fig:supp_gallery_rtpose_1}.}
  \label{fig:supp_gallery_rtpose_2}
\end{figure}

\begin{figure}[p]
  \centering
  \includegraphics[width=\linewidth]{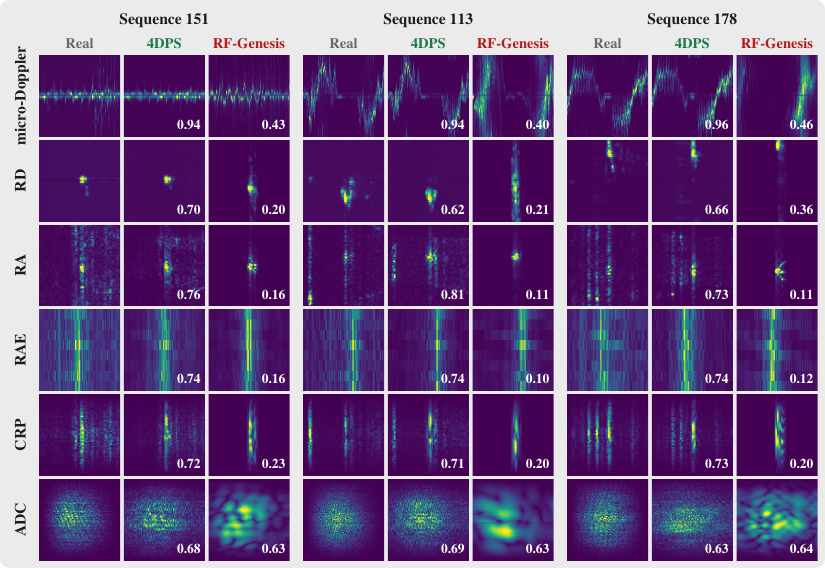}
  \\[4pt]
  \includegraphics[width=\linewidth]{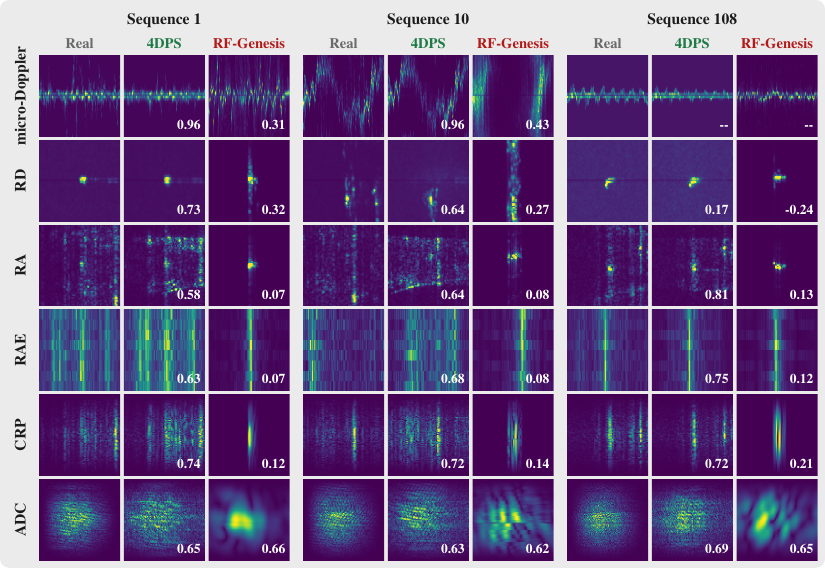}
  \caption{\textbf{Qualitative gallery, RT-Pose (3 of 3): sequences 151, 113 and 178 (top) and 1, 10 and 108 (bottom)} (4DPS-fidelity ranks 42, 46, 50, 54, 58 and 62 of 63). As Fig.~\ref{fig:supp_gallery_rtpose_1}.}
  \label{fig:supp_gallery_rtpose_3}
\end{figure}

\begin{figure}[p]
  \centering
  \includegraphics[width=\linewidth]{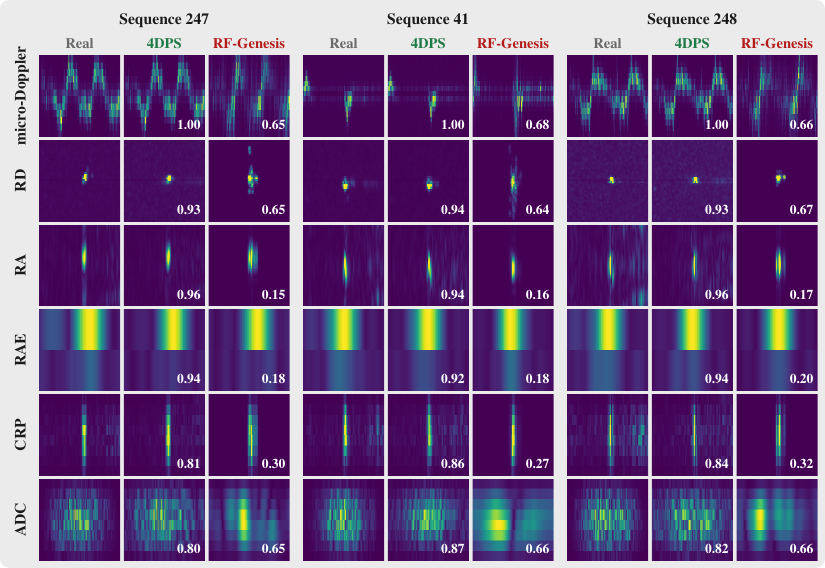}
  \\[4pt]
  \includegraphics[width=\linewidth]{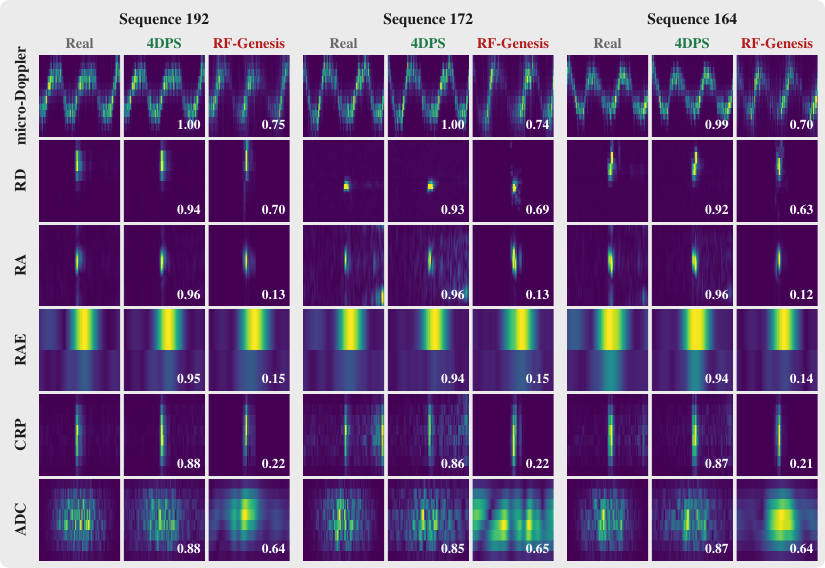}
  \caption{\textbf{Qualitative gallery, HuPR, horizontal board (1 of 3): sequences 247, 41 and 248 (top) and 192, 172 and 164 (bottom)} (4DPS-fidelity ranks 11, 57, 13, 8, 26 and 41 of 235). The top three are the HuPR test sequences ranked like those of Fig.~\ref{fig:qualitative} (4DPS fidelity plus rendered-to-real OKS). A single board resolves elevation poorly, so the range--azimuth--elevation rows are coarse along that axis.}
  \label{fig:supp_gallery_hupr_1}
\end{figure}

\begin{figure}[p]
  \centering
  \includegraphics[width=\linewidth]{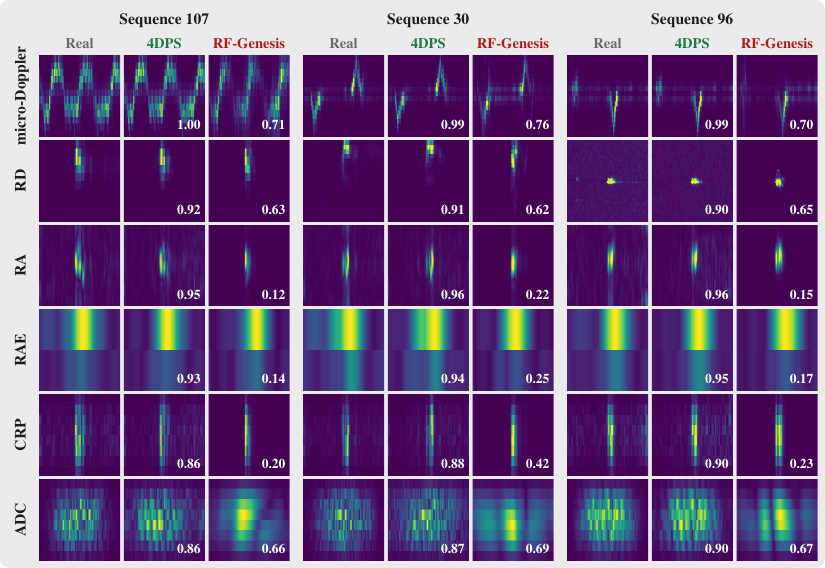}
  \\[4pt]
  \includegraphics[width=\linewidth]{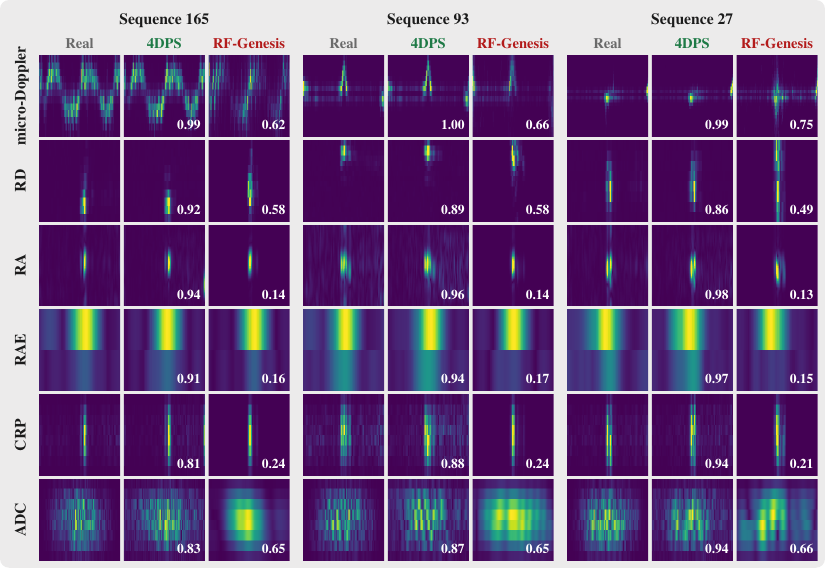}
  \caption{\textbf{Qualitative gallery, HuPR, horizontal board (2 of 3): sequences 107, 30 and 96 (top) and 165, 93 and 27 (bottom)} (4DPS-fidelity ranks 58, 73, 89, 104, 120 and 135 of 235). As Fig.~\ref{fig:supp_gallery_hupr_1}.}
  \label{fig:supp_gallery_hupr_2}
\end{figure}

\begin{figure}[p]
  \centering
  \includegraphics[width=\linewidth]{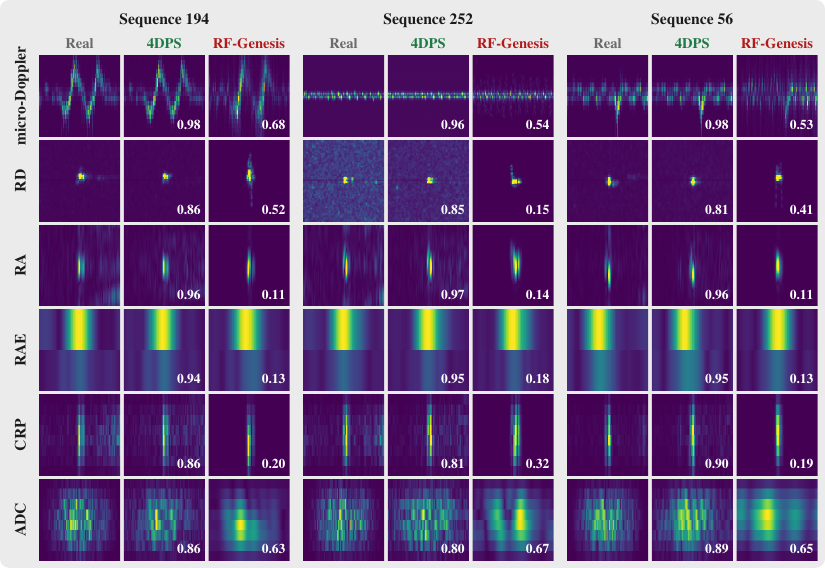}
  \\[4pt]
  \includegraphics[width=\linewidth]{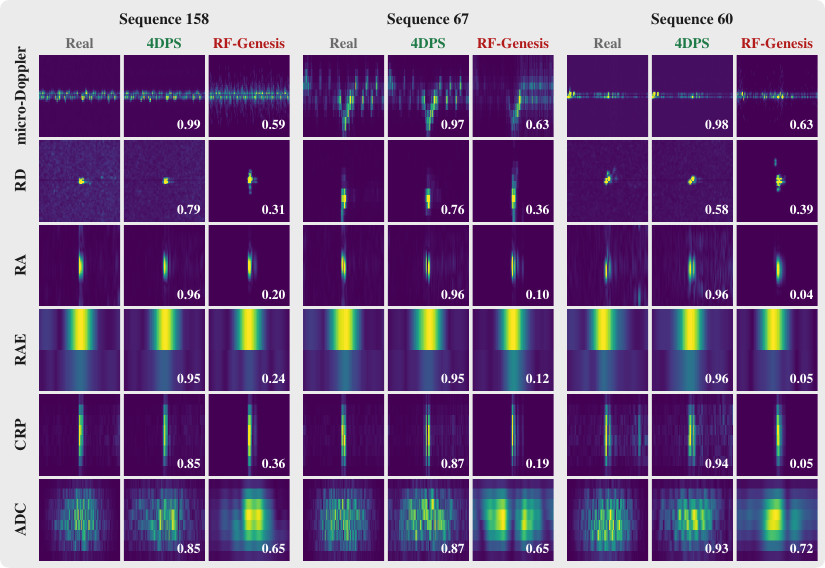}
  \caption{\textbf{Qualitative gallery, HuPR, horizontal board (3 of 3): sequences 194, 252 and 56 (top) and 158, 67 and 60 (bottom)} (4DPS-fidelity ranks 150, 166, 181, 197, 212 and 228 of 235). As Fig.~\ref{fig:supp_gallery_hupr_1}.}
  \label{fig:supp_gallery_hupr_3}
\end{figure}

Figs.~\ref{fig:supp_gallery_rtpose_1}--\ref{fig:supp_gallery_hupr_3} show 18 sequences per dataset, six per figure, chosen by a rule fixed before rendering so that they are typical rather than the best: we rank the corpus of Table~\ref{tab:fidelity} (HuPR on its horizontal board) by 4DPS fidelity (the mean range--azimuth, range--Doppler and micro-Doppler correlation), keep Fig.~\ref{fig:qualitative}'s three RT-Pose sequences and the three HuPR test sequences ranked the same way, and add 15 at evenly spaced quantiles of the rest. Columns, frames, normalization, scores and the first three rows follow Fig.~\ref{fig:qualitative}; the others are range--azimuth--elevation (azimuth $\times$ elevation, maximum over the person's range $\pm$10 bins), the complex range profile (range $\times$ virtual channel) and ADC samples (fast-time sample $\times$ channel). An undefined score prints as --.

\clearpage

\end{document}